\pdfoutput=1
\documentclass[11pt]{article}

\usepackage[final]{acl}

\usepackage{times}
\usepackage{latexsym}

\usepackage[T1]{fontenc}
\usepackage[utf8]{inputenc}

\usepackage{microtype}

\usepackage{inconsolata}

\usepackage{graphicx}
\usepackage{subcaption} % For subfigures
\usepackage{multirow}
\usepackage{amsmath,amsfonts,amssymb}
\usepackage{subcaption}
\usepackage{array}
\usepackage{makecell}
\usepackage{multicol}
\usepackage{booktabs}

\newcommand{\myvec}[1]{\mathbf{#1}}
\newcommand{\mymat}[1]{\mathbf{#1}}

\newenvironment{itemizesquish}[2]{\begin{list}{\labelitemi}{\setlength{\itemsep}{#1}\setlength{\labelwidth}{#2}\setlength{\leftmargin}{\labelwidth}\addtolength{\leftmargin}{\labelsep}}}{\end{list}}

\title{Layer by Layer: Uncovering Where Multi-Task Learning Happens in Instruction-Tuned Large Language Models}

\author{Zheng Zhao\textsuperscript{1} \quad Yftah Ziser\textsuperscript{2} \quad Shay B. Cohen\textsuperscript{1} \\
  \textsuperscript{1}Institute for Language, Cognition and Computation, University of Edinburgh \\
  \textsuperscript{2}Nvidia Research \\
  \texttt{zheng.zhao@ed.ac.uk} ,
  \texttt{yziser@nvidia.com} ,
  \texttt{scohen@inf.ed.ac.uk}}

\begin{document}
\maketitle
\begin{abstract}

Fine-tuning pre-trained large language models (LLMs) on a diverse array of tasks has become a common approach for building models that can solve various natural language processing (NLP) tasks. However, where and to what extent these models retain task-specific knowledge remains largely unexplored. This study investigates the task-specific information encoded in pre-trained LLMs and the effects of instruction tuning on their representations across a diverse set of over 60 NLP tasks. We use a set of matrix analysis tools to examine the differences between the way pre-trained and instruction-tuned LLMs store task-specific information. Our findings reveal that while some tasks are already encoded within the pre-trained LLMs, others greatly benefit from instruction tuning. Additionally, we pinpointed the layers in which the model transitions from high-level general representations to more task-oriented representations. This finding extends our understanding of the governing mechanisms of LLMs and facilitates future research in the fields of parameter-efficient transfer learning and multi-task learning.\footnote{Our code is available at: \url{https://github.com/zsquaredz/layer_by_layer/}} 

\end{abstract}

\section{Introduction}

While pre-trained LLMs exhibit impressive performance across diverse tasks and demonstrate remarkable generalization capabilities \cite{brown-etal-2020-language,wei-etal-2022-emergent,touvron2023llama,chowdhery-etal-2023-palm,openai2024gpt4}, the representations they learn and the task-specific information encoded during pre-training remain largely opaque and unexplored.

Recent research has investigated fine-tuning strategies to adapt LLMs to specific tasks, including supervised fine-tuning on task-specific datasets and instruction tuning \cite{mishra-etal-2022-cross,chung2022scaling,sanh-etal-2022-multitask}. While these approaches have shown promising results in tailoring LLMs for improved task performance, a comprehensive understanding of their impact on the learned representations is still lacking.

\begin{figure}[t]
    \centering
    \includegraphics[width=\linewidth]{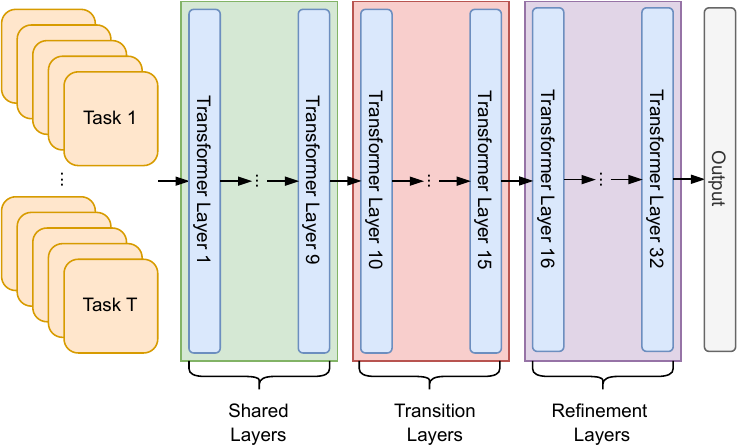}       
\caption{An illustration of our findings using the Llama 2 7B model \cite{touvron2023llama} as an example. We show that when instruction tuning on $T$ different tasks, the layers are divided into three functional sections: the shared layers (layers 1 to 9) form general representations shared among all tasks, the transition layers (layers 10 to 15) transition the representations into task-specific information, and the refinement layers (layers 16 to 32) continue to refine the representations toward specific tasks.}
\label{fig:diagram}
\end{figure}

In this study, we perform a set of analyses to investigate task-specific information encoded in pre-trained LLMs and the effects of instruction tuning on their representations. The analysis leverages a sub-population analysis technique called Model-Oriented Sub-population and Spectral Analysis (MOSSA; \citealt{zhao-etal-2022-understanding}), which provides an alternative to traditional probing methods for analyzing model representations within specific sub-populations of the training data. MOSSA involves comparing two models: a \textit{control} model trained on the data relevant to the sub-population of interest (e.g., a particular task), and an \textit{experimental} model that is identical to the control model but is also trained on additional data from different sources (e.g., multiple tasks). By analyzing the representational differences between these models, we can isolate the task-specific information encoded within the control model for the sub-population of interest.

To compare the representations learned by different LLM variants, we leverage the Center Kernel Alignment (CKA; \citealp[]{kornblith-etal-2019-similarity}) metric. CKA measures the alignment between representations in a kernel space, providing a robust measure of similarity that is insensitive to scaling and centering. By using MOSSA and CKA, we investigate the following research questions:
\begin{enumerate}
    \item To what extent are different NLP tasks already encoded in pre-trained LLMs?
    
    \item In what ways does instruction tuning modify the representational landscape of LLMs?

    \item Do the representational effects of instruction tuning generalize to unseen tasks?
\end{enumerate}

Through an extensive analysis spanning over 60 diverse NLP tasks following the Flan framework \cite{longpre-etal-2023-flan}, we shed light on the underlying mechanisms that govern the encoding and adaptation of task-specific information within LLMs under instruction tuning. A key finding of our work is the identification of three functional groups of layers: a) shared layers, in which more general information is learned and shared across tasks; b) transition layers, in which task-specific information is intensified; c) refinement layers, in which the LLMs continue to refine their representations towards task-specific predictions. Our findings contribute to a deeper understanding of the inner workings of LLMs and hold promising implications for future research in parameter-efficient fine-tuning (PEFT), multi-task learning (MTL), and model compression, benefiting a wide range of NLP applications.

We structure this study as follows: \S\ref{sec:methogology} describes our methodology for our analysis, while \S\ref{sec:exp_setup} outlines the experimental setup and tools used to train and analyze our LLMs. \S\ref{sec:results} then attempts to answer each of the research questions outlined above by presenting and analyzing our results. Finally, in \S\ref{sec:discussion}, we summarize our key findings and discuss their potential implications.

\section{Methodology}
\label{sec:methogology}

We use the MOSSA framework introduced by \citet{zhao-etal-2022-understanding}. Unlike standard probing methods \cite{belinkov-etal-2017-neural,belinkov-etal-2017-evaluating,giulianelli-etal-2018-hood}, which build a model to predict a downstream task for quantifying encoded information, MOSSA compares representations from two models: a control model trained on data of interest and an experimental model trained on additional data from different sources. Here, the data of interest refers to tasks. Probing methods, while useful, can be limited because they rely on different metrics to evaluate performance across various tasks, making it challenging to directly compare the amount of information stored about tasks as diverse as sentiment analysis and translation. MOSSA, on the other hand, circumvents this issue by comparing the latent representations of models rather than their downstream performance metrics. MOSSA calculates the similarity between the representations of the control and experimental models, thus representing the information captured from the relevant sub-population of data through their latent representations. By comparing different models to each other, we can learn what information is captured when a subset of the data is used versus the whole dataset.

We use matrix analysis to compare representation similarity between the experimental model, such as pre-trained, instruction-tuned, and corresponding single-task control models trained on individual tasks. Intuitively, a high similarity between the experimental and control models indicates the experimental model stores task-specific information learned by the control model, which was fine-tuned solely on data from that task. 
The similarity is measured using the CKA metric, which quantifies the similarity between two representations in a kernel space.

Formally, let $[T]$ be an index set of tasks, and let $\mathbf{E}$ be the experimental model and $\mathbf{C}_{t}$ be the control model for task $t \in [T]$. We assume a set of inputs $\mathcal{X} = \bigcup_{t=1}^{T} \mathcal{X}_{t}$, where each $\mathcal{X}_{t} = \{ \myvec{x}_{t,1}, \ldots, \myvec{x}_{t,n} \}$ represents a set of input instructions for task $t$. For simplicity, we assume that all sets have the same size $n$, although this is not a strict requirement.\footnote{In our actual experimental setup for this work, we use different dataset sizes for each task, which reflects real-world scenarios. For more details, please refer to \S\ref{sec:exp_setup}.}

For each $t \in [T]$ and $i \in [n]$, we apply the experimental model $\mathbf{E}$ and the control model $\mathbf{C}_{t}$ to the input instruction $\myvec{x}_{t,i}$ to obtain two corresponding representations $\myvec{y}_{t,i} \in \mathbb{R}^{d}$ and $\myvec{z}_{t,i} \in \mathbb{R}^{d_{t}}$, respectively. Here, $d$ is the dimension of the experimental model representations, and $d_{t}$ is the dimension of the control model representations for task $t$. To obtain the representations $\myvec{y}_{t,i}$ and $\myvec{z}_{t,i}$, we use the last token representation following previous work \cite{qiu2024spectral,wang2024improving}, as LLMs are decoder-only and the last token captures all input information. These representations can be extracted from any layers of the respective models.

By stacking these vectors into two matrices for each task $t$, we obtain the paired matrices $\mymat{Y}_{t} \in \mathbb{R}^{n \times d}$ and $\mymat{Z}_{t} \in \mathbb{R}^{n \times d_{t}}$.
We calculate the CKA value between $\mymat{Y}_{t}$ and $\mymat{Z}_{t}$ following the procedure:

\begin{itemizesquish}{-0.1em}{0.5em}
\item Computing the kernel matrices ${K}_{\mymat{Y}_{t}} \in \mathbb{R}^{n \times n}$ and ${K}_{\mymat{Z}_{t}} \in \mathbb{R}^{n \times n}$ for $\mymat{Y}_{t}$ and $\mymat{Z}_{t}$, respectively, using the same kernel function (e.g., linear, Gaussian, or polynomial).\footnote{For linear kernel, which is what we use in our experiment, ${K}_{\mymat{Y}_{t}} = \mymat{Y}_{t}\mymat{Y}_{t}^{\top}$, and $ {K}_{\mymat{Z}_{t}} = \mymat{Z}_{t}\mymat{Z}_{t}^{\top}$.}

\item Centering the kernel matrices by \( K_{\mymat{Y}_t} = K_{\mymat{Y}_t} - \frac{1}{n} \mathbf{1} K_{\mymat{Y}_t} - \frac{1}{n} K_{\mymat{Y}_t} \mathbf{1} + \frac{1}{n^2} \mathbf{1} K_{\mymat{Y}_t} \mathbf{1} \), similarly for ${K}_{\mymat{Z}_{t}}$, where \(\mathbf{1}\) is a matrix of ones.

\item Computing the CKA value by first compute the Frobenius inner product of the centered Gram matrices: \(\text{HSIC}(K_{\mymat{Y}_t}, K_{\mymat{Z}_t}) = \text{Tr}(K_{\mymat{Y}_t}^\top K_{\mymat{Z}_t})\), where Tr denotes the trace of a matrix. Then normalize the CKA value: 

\end{itemizesquish}
\vspace{-3ex}

\begin{equation}
\scriptstyle
    \text{CKA}(\mymat{Y}_t, \mymat{Z}_t) = \frac{\text{HSIC}(K_{\mymat{Y}_t}, K_{\mymat{Z}_t})}{\sqrt{\text{HSIC}(K_{\mymat{Y}_t}, K_{\mymat{Y}_t}) \cdot \text{HSIC}(K_{\mymat{Z}_t}, K_{\mymat{Z}_t})}}.
\end{equation}

While other similarity metrics like SVCCA \cite{raghu-etal-2017-svcca} exist, they have a limitation due to the constraint of being invariant to invertible linear transformations, which requires the number of data points to be greater than the number of representation dimensions. We use CKA as it has shown robust results when the data sample is smaller \cite{kornblith-etal-2019-similarity}, as is sometimes the case for datasets used in our work.

Our method provides an approach to quantify the task-specific information encoded in the representations of LLMs. By comparing the experimental model's representations with those of single-task control models, we can gain insights into the extent to which the experimental model captures task-specific knowledge and how this knowledge is distributed across its representations.

\begin{figure*}[ht]
    \centering
    \includegraphics[width=\linewidth]{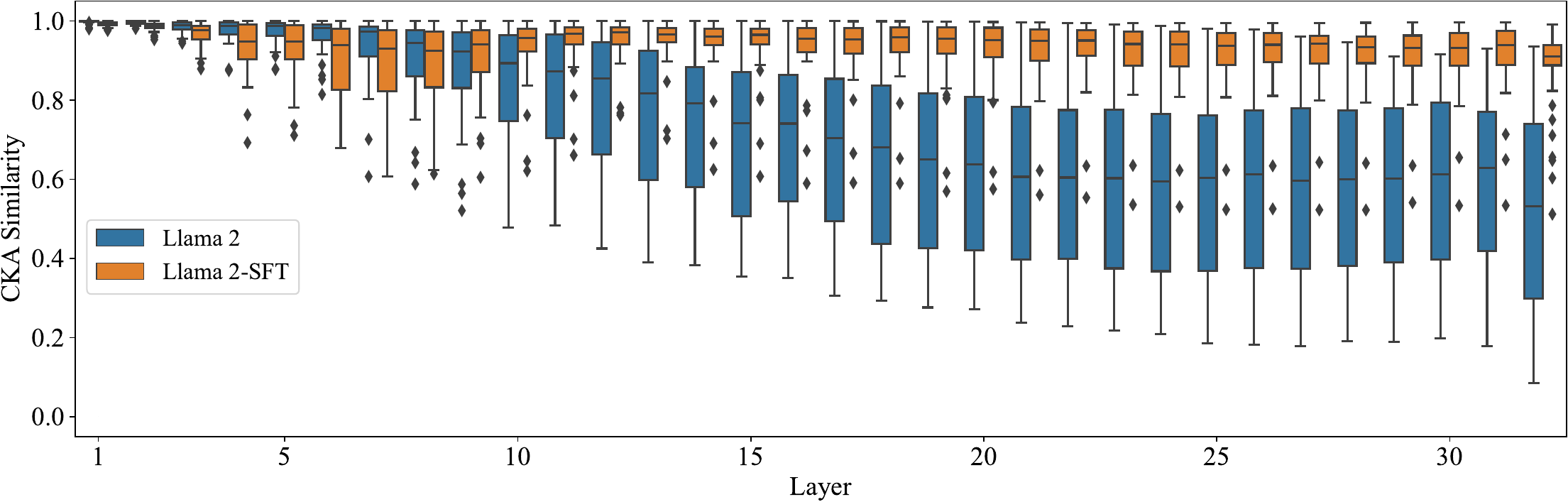}       
\caption{Distribution of CKA similarities across all layers for the pre-trained Llama 2 model and the instruction-tuned Llama 2-SFT model. The boxplots illustrate the spread and variation of CKA similarities between each model and the control models across different tasks. The comparison between the two models highlights the impact of instruction tuning on shaping task-specific representations in different layers.}
\label{fig:cka_all_layers}
\end{figure*}

\section{Experimental Setup}
\label{sec:exp_setup}
\paragraph{Data} We use the Flan 2021 dataset \cite{wei-etal-2022-finetuned} to fine-tune our LLMs. The Flan dataset is a comprehensive collection of more than 60 NLP datasets, including both language understanding and generation tasks. These datasets are organized into twelve task clusters, where datasets within a given cluster belong to the same task type.
To enhance instruction diversity, we follow the approach of \citet{wei-etal-2022-finetuned} and use ten unique natural language instruction templates for each dataset. These templates provide varying descriptions of the task to be performed. Our instruction tuning pipeline combines all datasets and randomly samples from each dataset during training.
To mitigate the impact of dataset size imbalances, we limit the number of training examples per task cluster to 50k and use the examples-proportional mixing scheme \cite{raffel-etal-2020-exploring} with a mixing rate maximum of 3,000 per task. This means that no task receives additional sampling weight for examples in excess of 3,000. We provide further details about the dataset in Appendix~\ref{app:dataset}.

\paragraph{Models} We have two types of models: the experimental model $\mathbf{E}$, fine-tuned using all $T$ available tasks, and the single-task model $\mathbf{C}_t$ for $t \in [T]$, fine-tuned only on the $t$-th task. In some experiments, the model $\mathbf{E}$ can also be the pre-trained model. We use the Llama 2 models \cite{touvron2023llama} as the starting training checkpoint for both $\mathbf{E}$ and $\mathbf{C}_t$. Specifically, we use the 7B variant, which consists of 32 layers and 4096 hidden dimensions. This model allows us to conduct a more comprehensive set of experiments while maintaining control over experimental conditions. Since we have over 60 control models, exploring larger models or different families would have been computationally infeasible due to resource constraints. Given these limitations, we choose to fully explore a realistic multi-task scenario, involving more than 60 different tasks, with the aim of extracting significant findings that we expect to generalize to other models.

\begin{figure}[h]
    \centering
    \begin{subfigure}{\linewidth}
        \includegraphics[width=\textwidth]{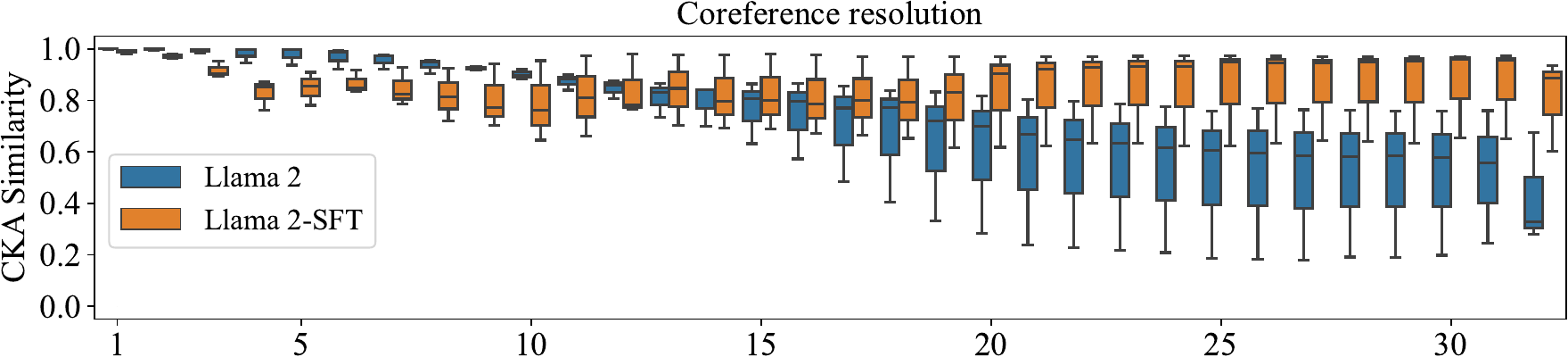}
        \vspace{-2ex}
        \label{fig:cka_task_coreference}  
    \end{subfigure}
    \begin{subfigure}{\linewidth}
        \includegraphics[width=\textwidth]{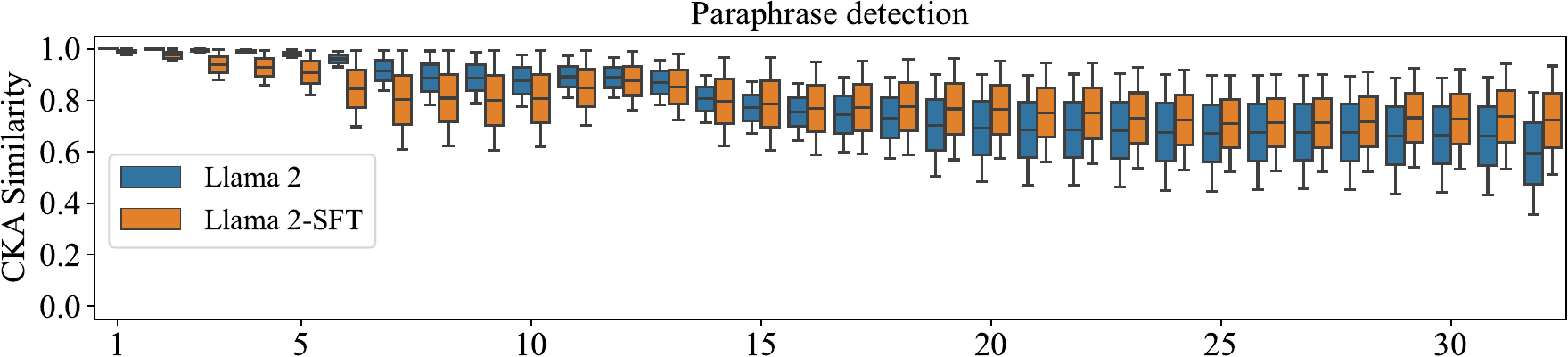}
        \vspace{-2ex}
        \label{fig:cka_task_paraphrase}
    \end{subfigure}
    \begin{subfigure}{\linewidth}
        \includegraphics[width=\textwidth]{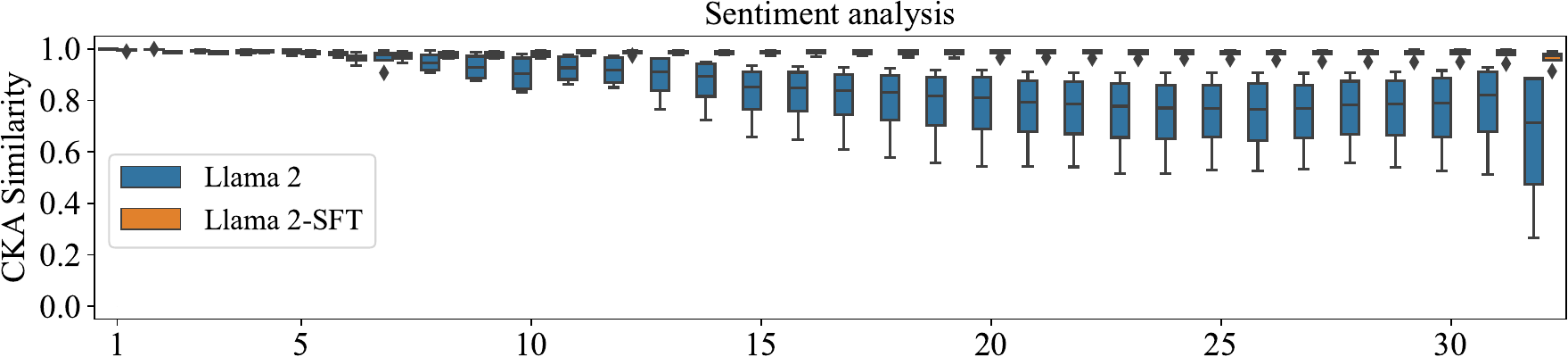}
        \vspace{-2ex}
        \label{fig:cka_task_sentiment}
    \end{subfigure}
    \begin{subfigure}{\linewidth}
        \includegraphics[width=\textwidth]{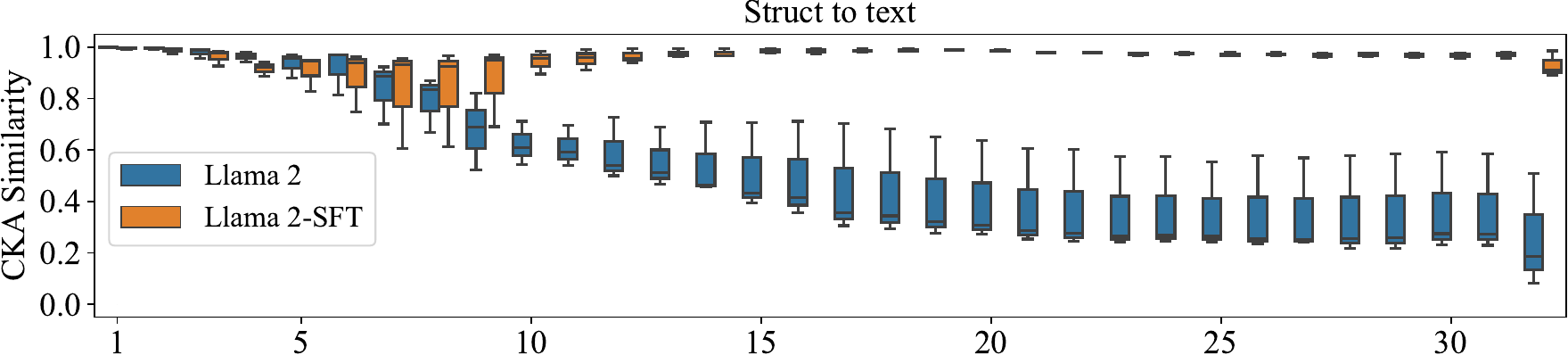}
        \vspace{-2ex}
        \label{fig:cka_task_s2t}
    \end{subfigure}
    \begin{subfigure}{\linewidth}
        \includegraphics[width=\textwidth]{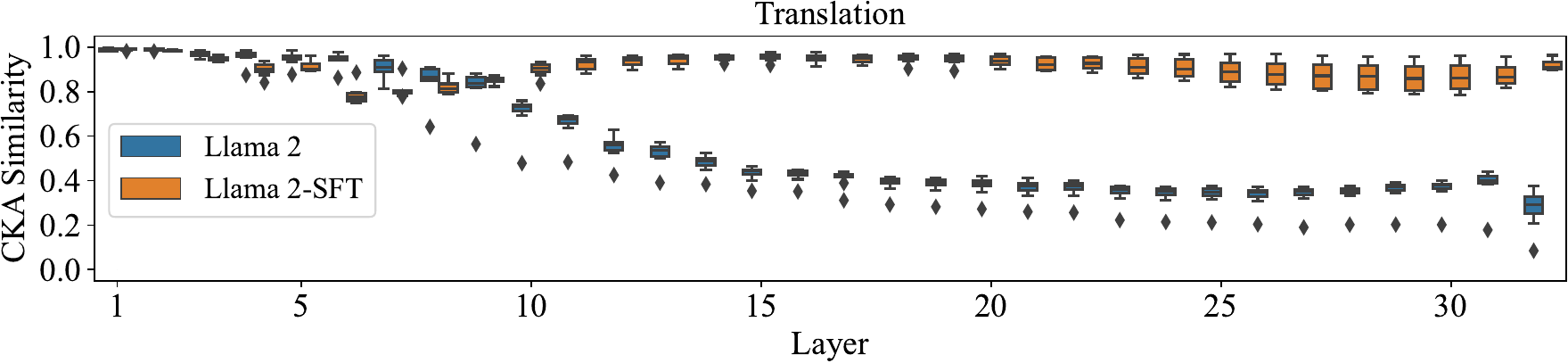}
        \vspace{-3ex}
        \label{fig:cka_task_translation}
    \end{subfigure}
\caption{Distribution of CKA similarities across all layers for the pre-trained Llama 2 model and the instruction-tuned Llama 2-SFT model, grouped by different task clusters.}
\label{fig:cka_all_layers_all_clusters_example}
\end{figure} 

\paragraph{Training} We use LoRA \cite{hu-etal-2022-lora} for fine-tuning our LLMs, with the rank $r$ set to 8. We use the AdamW optimizer \citep{loshchilov-hutter-2019-decoupled} with a learning rate of $5 \times 10^{-5}$ for fine-tuning the instruction dataset. We use the same vocabulary, tokenizer, and learning rate scheduler for Llama 2-7B as in \citet{touvron2023llama}.
We train the multi-task model $\mathbf{E}$ (which we refer to as Llama 2-SFT in our experiment) for a maximum of 100K steps and the single-task models $\mathbf{C}_t$ for a maximum of 10K steps, using validation set cross-entropy loss for early stopping.
Our multi-task models are trained on four NVIDIA A100 GPUs with a batch size of 16 per GPU, while single-task models are trained on one NVIDIA A100 GPU with a batch size of 16.We use PyTorch \citep{NEURIPS2019_9015}, the HuggingFace library \citep{wolf2019huggingface}, and the LLaMA-Factory library \citep{zheng2024llamafactory} for all model implementations and LoRA fine-tuning.

\begin{figure*}[th]
     \centering
     \begin{subfigure}[b]{\textwidth}
         \centering
         \includegraphics[width=\textwidth]{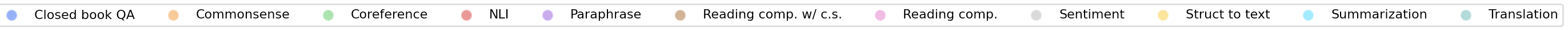}
         \label{fig:tsne_llama2_legend}
         \vspace{-2ex}
     \end{subfigure}
     \begin{subfigure}[b]{0.19\textwidth}
         \centering
         \includegraphics[width=\textwidth]{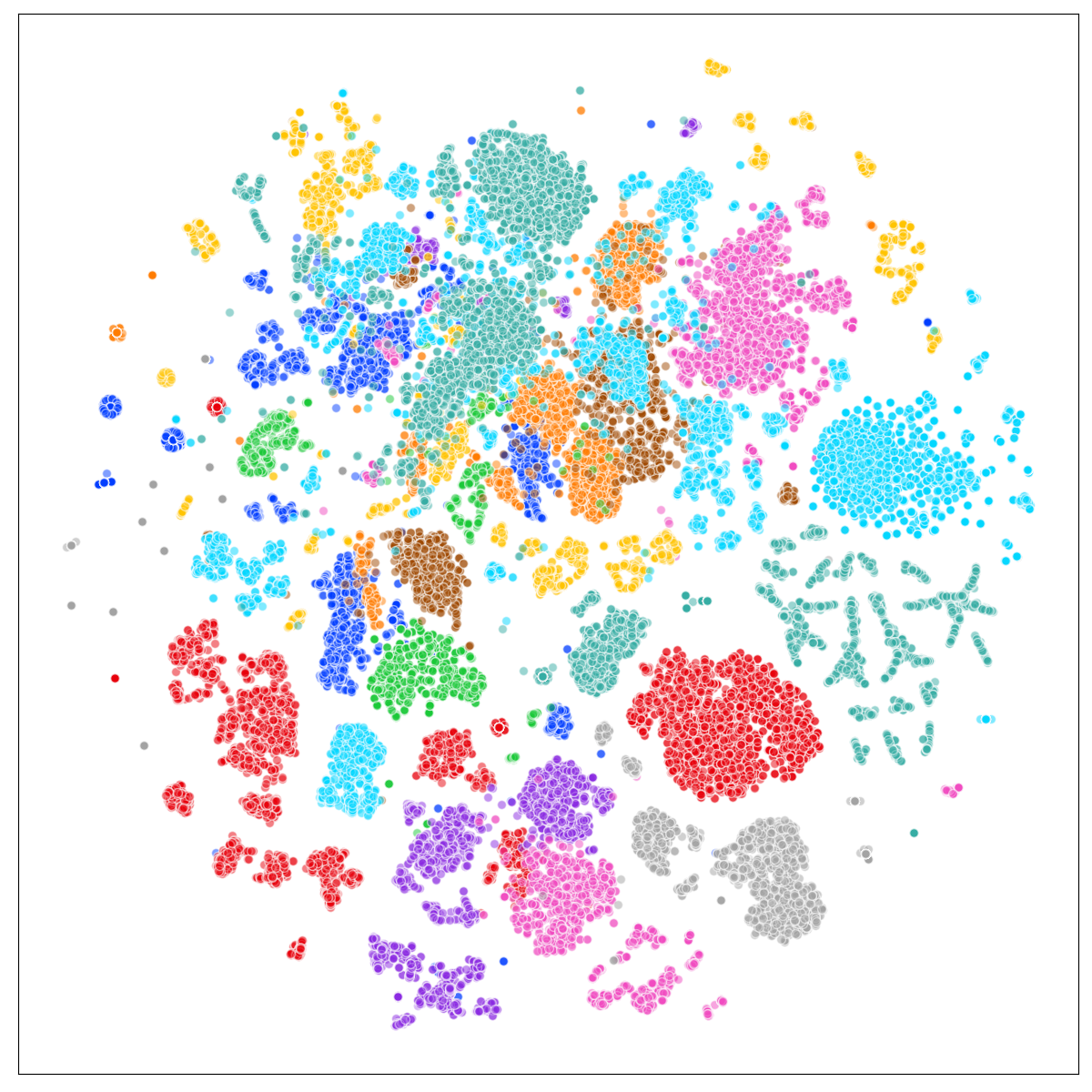}
         \caption{Llama 2 L1}
         \label{fig:tsne_llama2_l1}
     \end{subfigure}
    \hfill
    \begin{subfigure}[b]{0.19\textwidth}
         \centering
         \includegraphics[width=\textwidth]{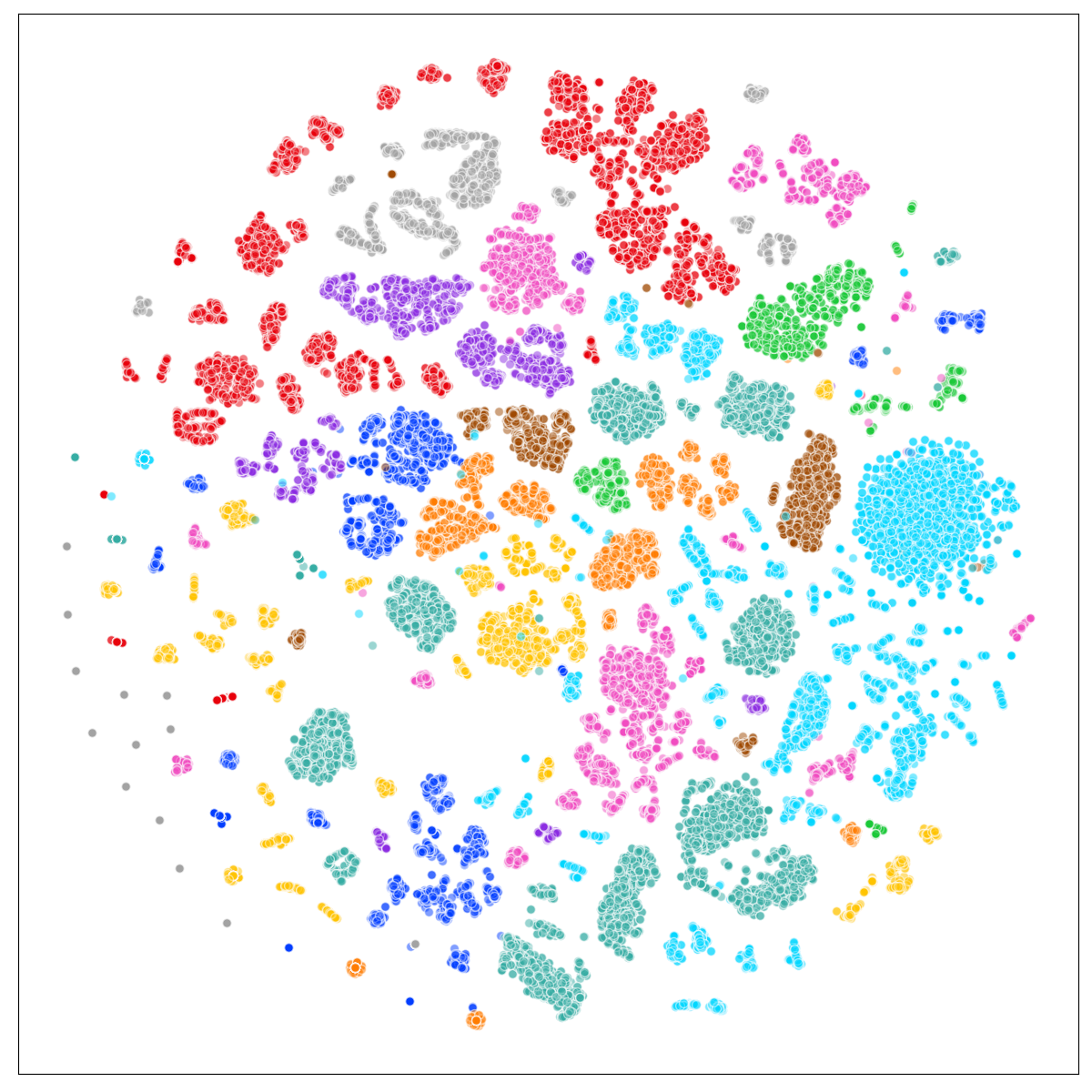}
         \caption{Llama 2  L10}
         \label{fig:tsne_llama2_l10}
     \end{subfigure}
     \hfill
    \begin{subfigure}[b]{0.19\textwidth}
         \centering
         \includegraphics[width=\textwidth]{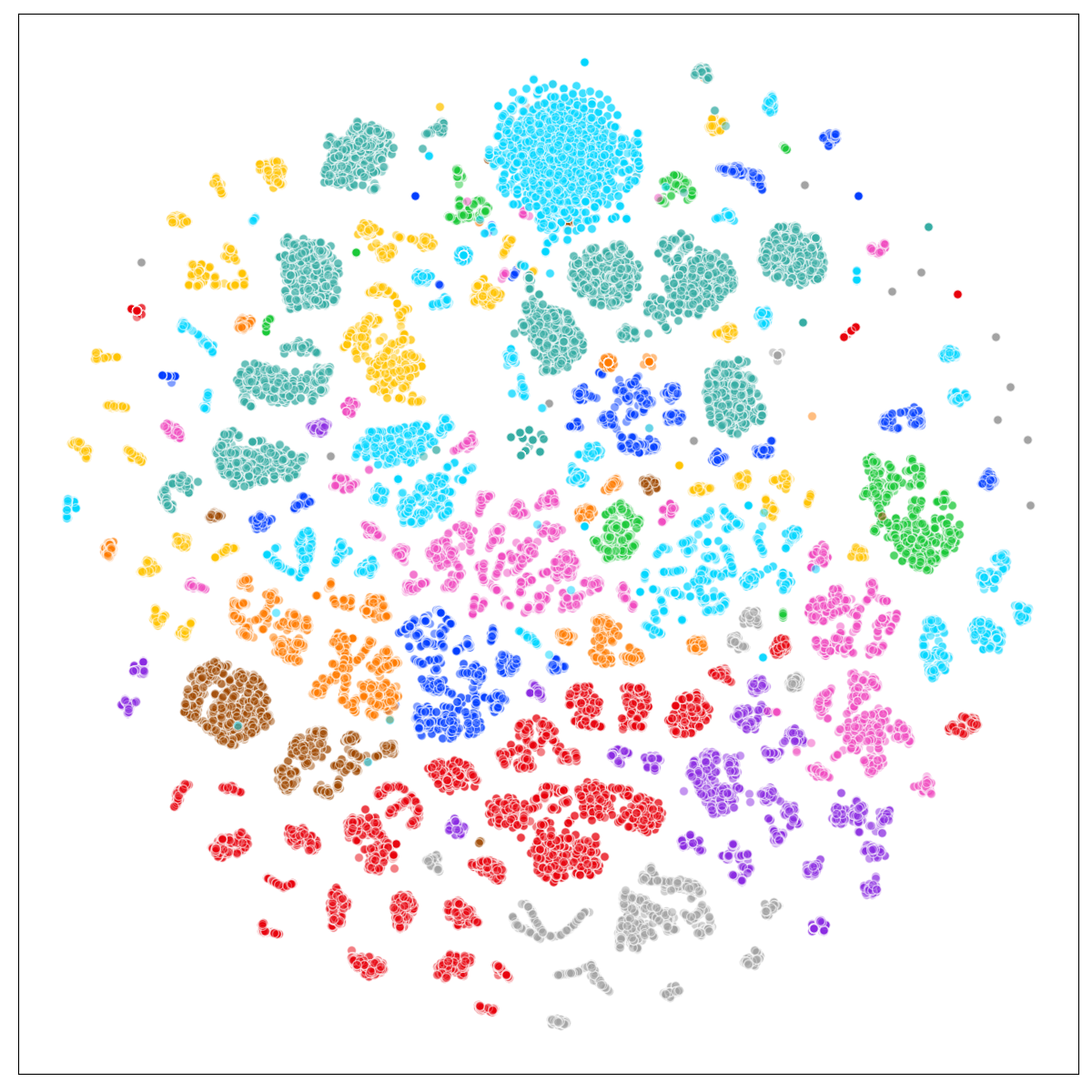}
         \caption{Llama 2 L15}
         \label{fig:tsne_llama2_l15}
     \end{subfigure}
     \hfill
    \begin{subfigure}[b]{0.19\textwidth}
         \centering
         \includegraphics[width=\textwidth]{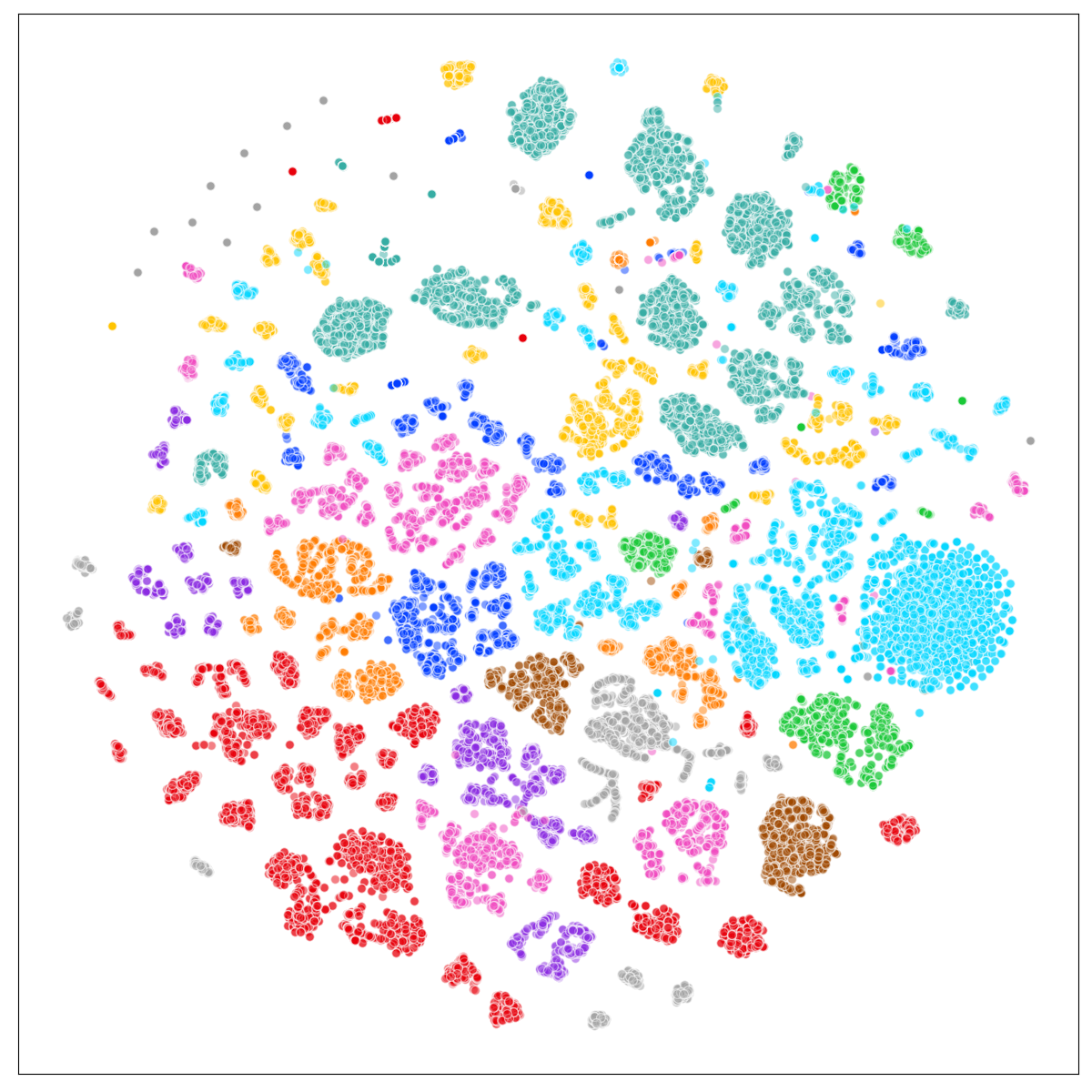}
         \caption{Llama 2 L20}
         \label{fig:tsne_llama2_l20}
     \end{subfigure}
     \hfill
     \begin{subfigure}[b]{0.19\textwidth}
         \centering
         \includegraphics[width=\textwidth]{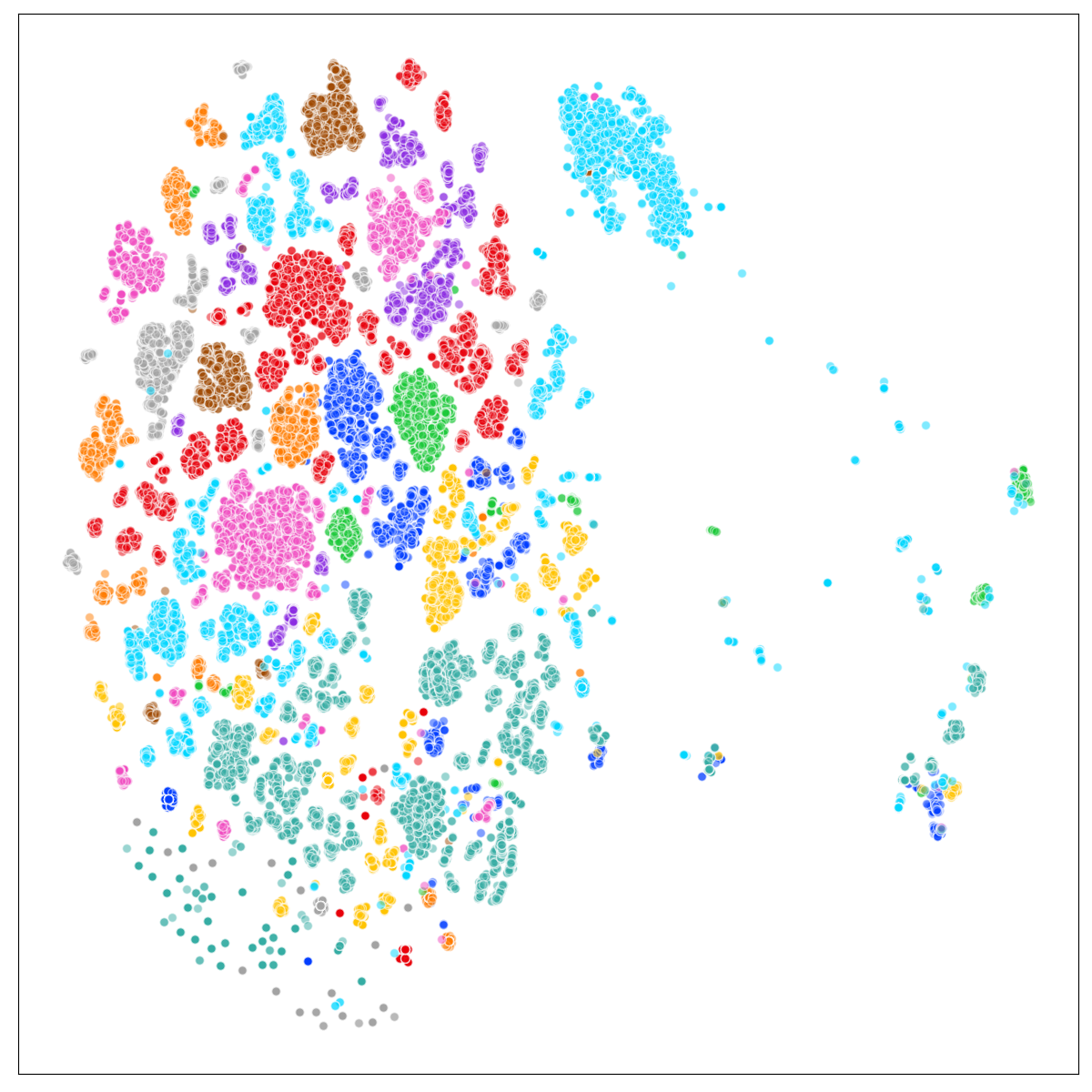}
         \caption{Llama 2 L32}
         \label{fig:tsne_llama2_l32}
     \end{subfigure}
     \begin{subfigure}[b]{0.19\textwidth}
         \centering
         \includegraphics[width=\textwidth]{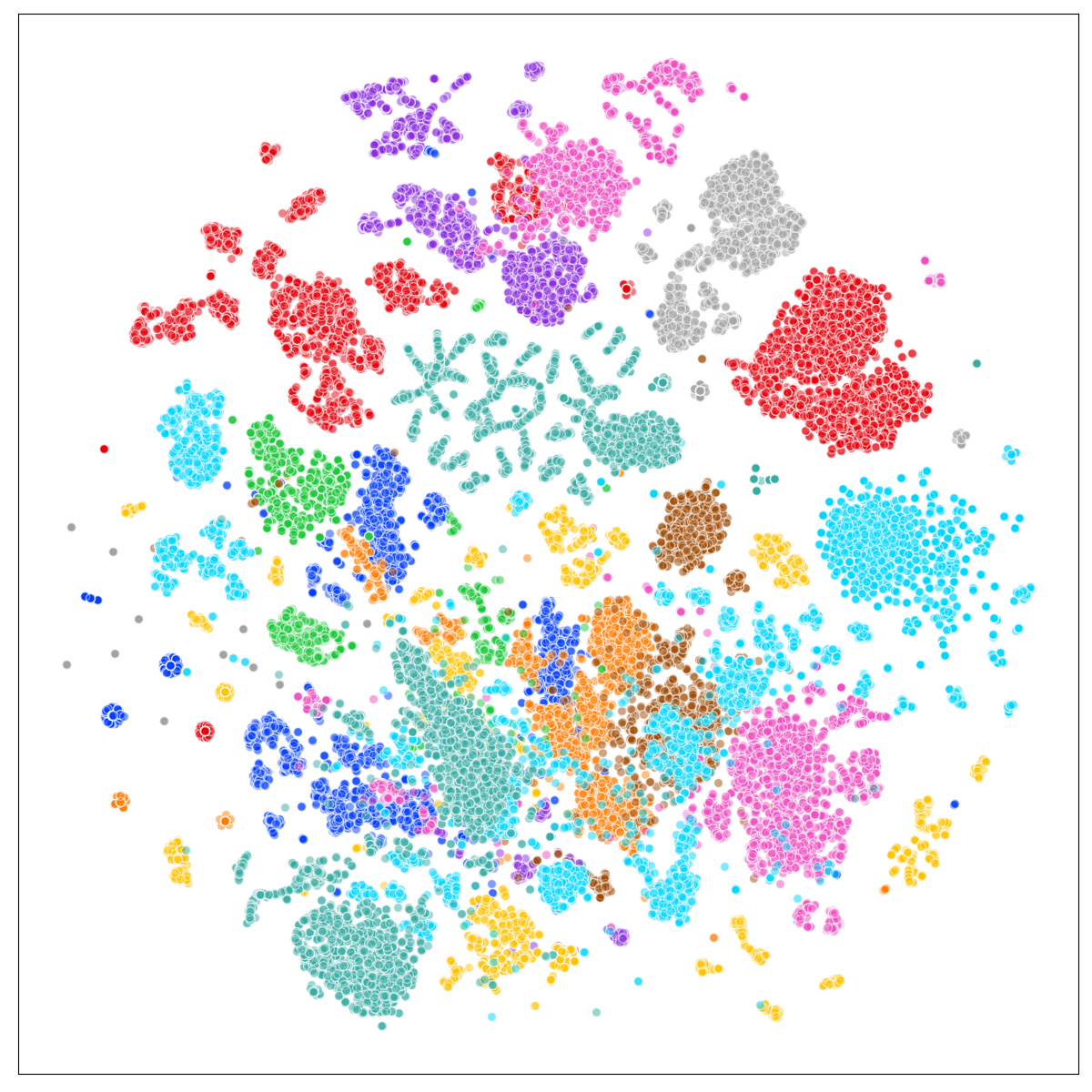}
         \caption{Llama 2-SFT L1}
         \label{fig:tsne_llama2_sft_l1}
     \end{subfigure}
    \hfill
    \begin{subfigure}[b]{0.19\textwidth}
         \centering
         \includegraphics[width=\textwidth]{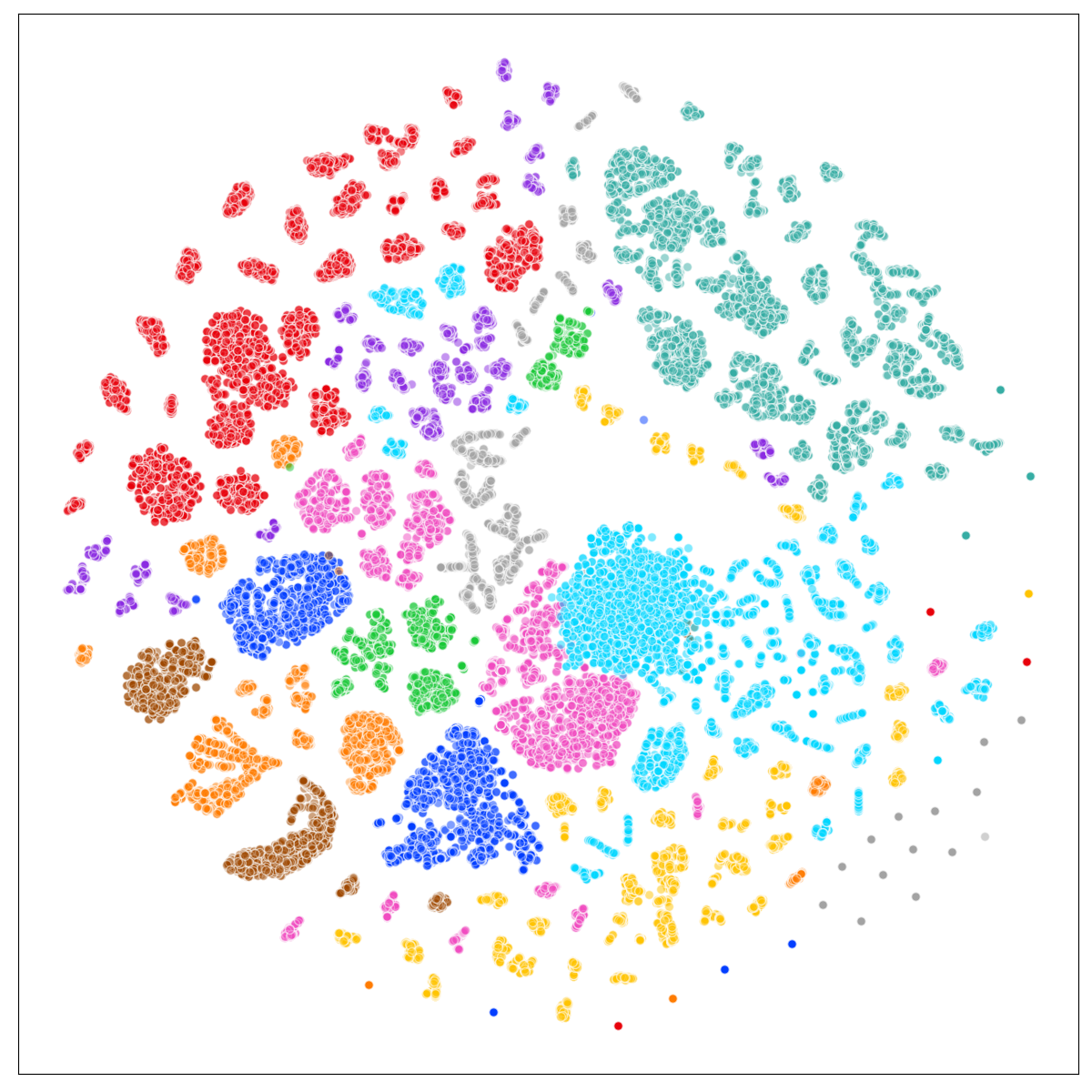}
         \caption{Llama 2-SFT L10}
         \label{fig:tsne_llama2_sft_l10}
     \end{subfigure}
    \hfill
    \begin{subfigure}[b]{0.19\textwidth}
         \centering
         \includegraphics[width=\textwidth]{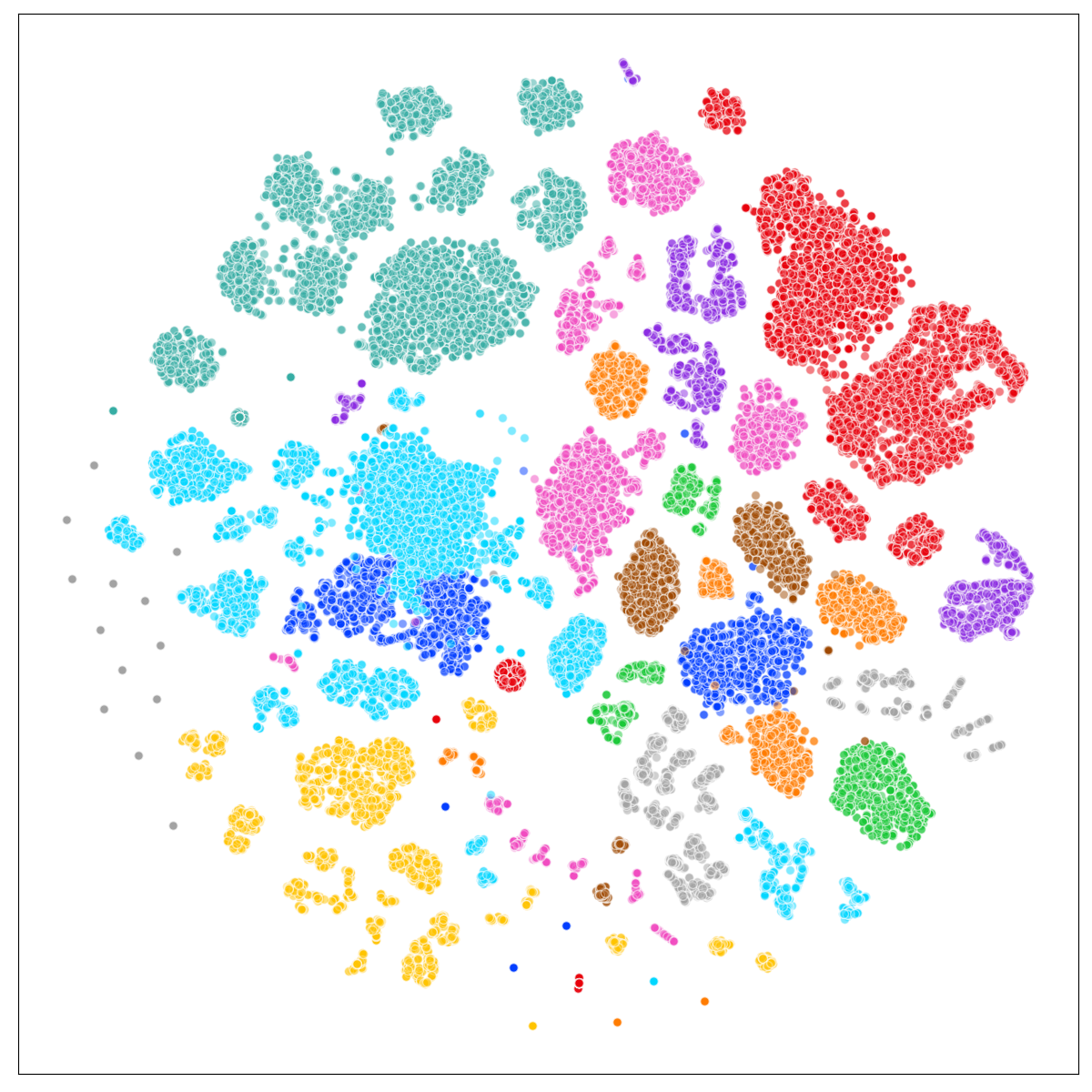}
         \caption{Llama 2-SFT L15}
         \label{fig:tsne_llama2_sft_l15}
     \end{subfigure}
    \hfill
     \begin{subfigure}[b]{0.19\textwidth}
         \centering
         \includegraphics[width=\textwidth]{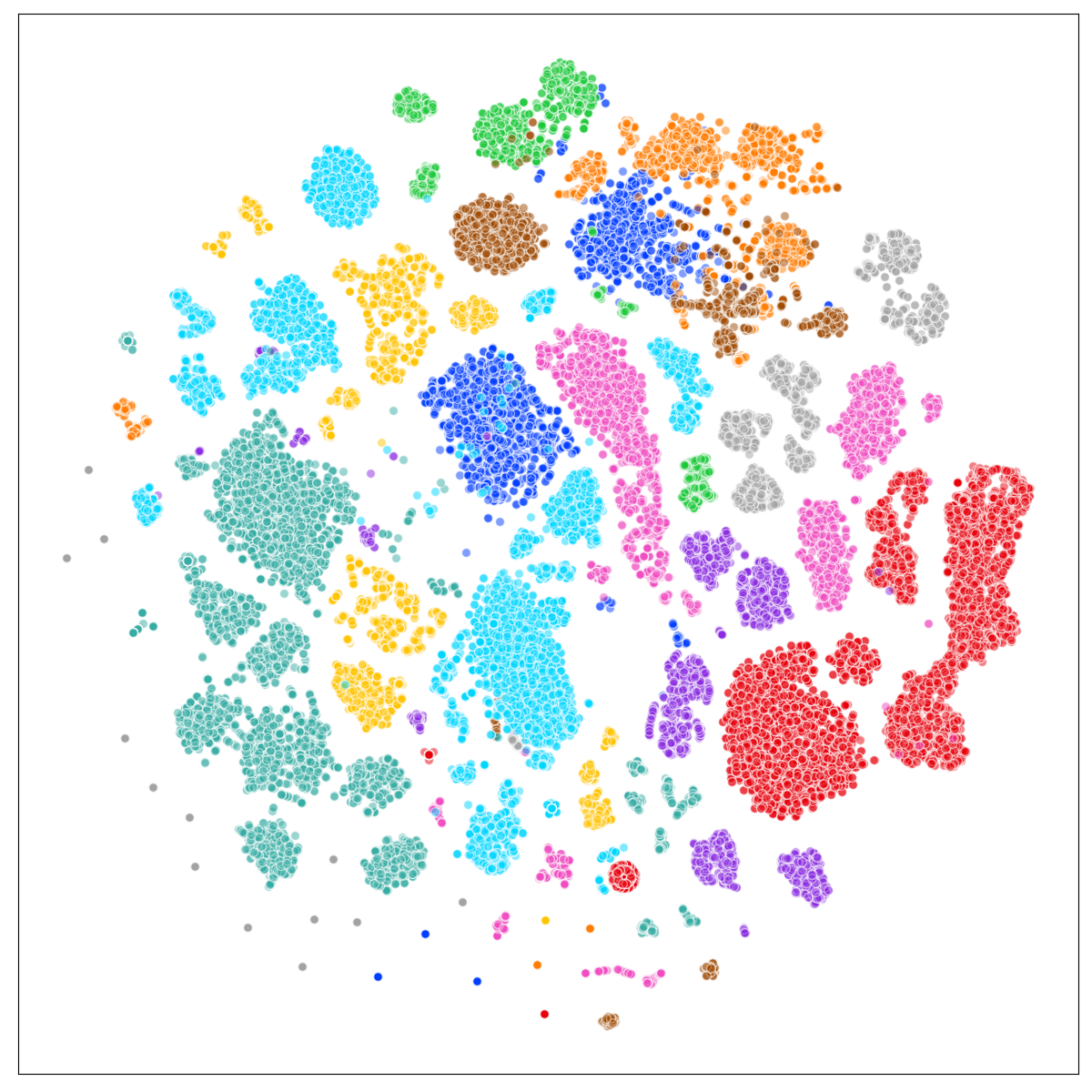}
         \caption{Llama 2-SFT L20}
         \label{fig:tsne_llama2_sft_l20}
     \end{subfigure}
    \hfill
     \begin{subfigure}[b]{0.19\textwidth}
         \centering
         \includegraphics[width=\textwidth]{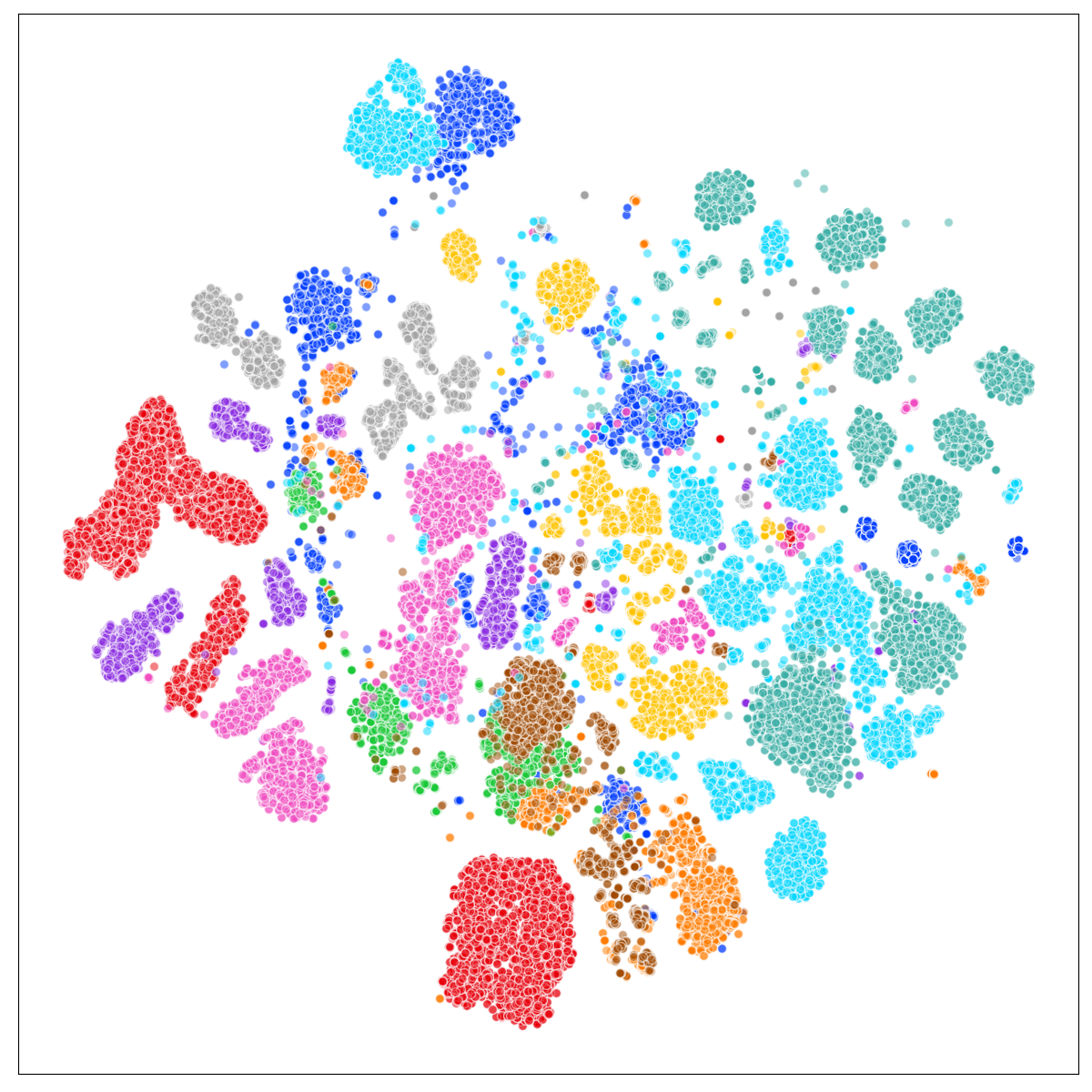}
         \caption{Llama 2-SFT L32}
         \label{fig:tsne_llama2_sft_l32}
     \end{subfigure}
    \caption{t-SNE visualizations of the representations for each task cluster in different layers of the pre-trained Llama 2 model and the instruction-tuned Llama 2-SFT model. Each subplot presents the t-SNE projection of the representations, color-coded by task cluster, for a specific layer of the respective model. ``Reading comp.'' denotes reading comprehension tasks, and ``reading comp. w/ c.s.'' denotes reading comprehension tasks with commonsense reasoning.}
    \label{fig:tsne_selection}
\end{figure*}

\section{Experiments and Results}
\label{sec:results}

To shed light on the underlying mechanisms of MTL \cite{caruana-etal-1997-multitask} in LLMs, we start by examining what NLP tasks are encoded in the pre-trained LLM representations, establishing a baseline for comparison with the instruction-tuned model (\S\ref{sec:tasks_in_llm}). Then, using matrix analysis methods, we contrast the representational properties of the pre-trained and instruction-tuned LLMs to understand the effects of instruction tuning (\S\ref{sec:impact_of_instruct}, \ref{sec:tsne_and_pca}, and \ref{sec:task_property_analysis}). Finally, we evaluate the generalization of our findings to unseen tasks (\S\ref{sec:eval_on_unseen_tasks}).

\subsection{Task Information in Pre-trained LLMs}
\label{sec:tasks_in_llm}

To identify task-relevant information in pre-trained LLMs, we compared representations from the pre-trained Llama 2 model with task-specific fine-tuned models ($\{ \mathbf{C}_t \}_t$). Figure~\ref{fig:cka_all_layers} shows the distribution of CKA similarities across all tasks and layers for the Llama 2 model. The CKA similarities between pre-trained Llama 2 and control models generally decrease through higher layers.

Llama 2 maintains high CKA similarities in earlier layers, and since CKA compares against control models fine-tuned on individual tasks, this suggests that representational changes in the earlier layers are minimal across tasks. However, we observe widespread variance in CKA values across different tasks in the middle and higher layers, suggesting that some tasks are better captured in the Llama 2 model representations than others.

To gain a more fine-grained understanding, we analyzed the CKA results at the task cluster level, where each cluster consists of a group of similar tasks. The Flan dataset organizes tasks into 12 different clusters, detailed in Appendix~\ref{app:dataset}. We present CKA results for a selection of representative clusters in Figures~\ref{fig:cka_all_layers_all_clusters_example}, with the full results provided in Appendix~\ref{app:additional_results_cka}.

For clusters like closed-book QA, commonsense reasoning, paraphrase detection, and sentiment analysis, which heavily rely on general linguistic and semantic understanding, the CKA similarity for Llama 2 is high. This indicates that pre-trained models already encode these tasks well in their representations. Conversely, for clusters like coreference resolution, reading comprehension, structured data to text generation, summarization, and translation, which require specialized, structured, or domain-specific knowledge involving complex transformations or extended context management, the CKA similarities are low, suggesting that next token prediction at pre-training is insufficient for encoding these tasks.

\subsection{Impact of Instruction Tuning}
\label{sec:impact_of_instruct}
\paragraph{Mapping Layers to Their Functionality}

To investigate how instruction tuning affects the representations learned by LLMs, we compared the instruction-tuned model (Llama 2-SFT) with task-specific fine-tuned control models. As illustrated in Figure~\ref{fig:cka_all_layers}, {the CKA similarities between Llama 2-SFT and the control models do not decrease as significantly as those for the pre-trained model (Llama 2) across layers}. In the early layers (1 to 9), we observe that for many tasks, the CKA scores are lower for Llama 2-SFT compared to Llama 2, indicating that Llama 2-SFT representations diverge from those of the control models, which were fine-tuned on individual tasks (thus specializing in them). This suggests that, unlike the Llama 2 model, training Llama 2-SFT on a high number of tasks encourages it diverge from the control models' representations and learn more general representations in the lower layers, a characteristic typical of MTL models. We denote layers 1-9 as ``shared layers'', as our findings suggest their representations are shared across tasks, similar to more studied MTL models. 

In the middle layers (10-15), there is a significant transition, with the Llama 2-SFT model exhibiting high similarity to \emph{all control models}. This indicates that these layers encode a high degree of task-specific information, as their representations are almost identical to those of the specialized control models. We denote layers 10-15 as ``transitional layers'', as our findings suggest the transition to task-specific representations occurs within these layers. This trend continues, albeit to a lesser extent, up to the final layers (16-32), which we denote as ``refinement layers'', as they keep refining the representations up to the final prediction. Based on our findings, {we can map each layer in the Llama 2-SFT model to its corresponding function with respect to MTL} (see Figure \ref{fig:diagram}). While previous work \cite{wei-etal-2022-finetuned,chung2022scaling} has empirically demonstrated the effectiveness of instruction tuning for improving performance on a variety of NLP tasks, to the best of our knowledge, we are the first to propose such a mapping. In the following sections, we provide additional analyses to further validate our mapping.

\begin{figure}[t]   
    \centering
     \includegraphics[width=1.0\linewidth]{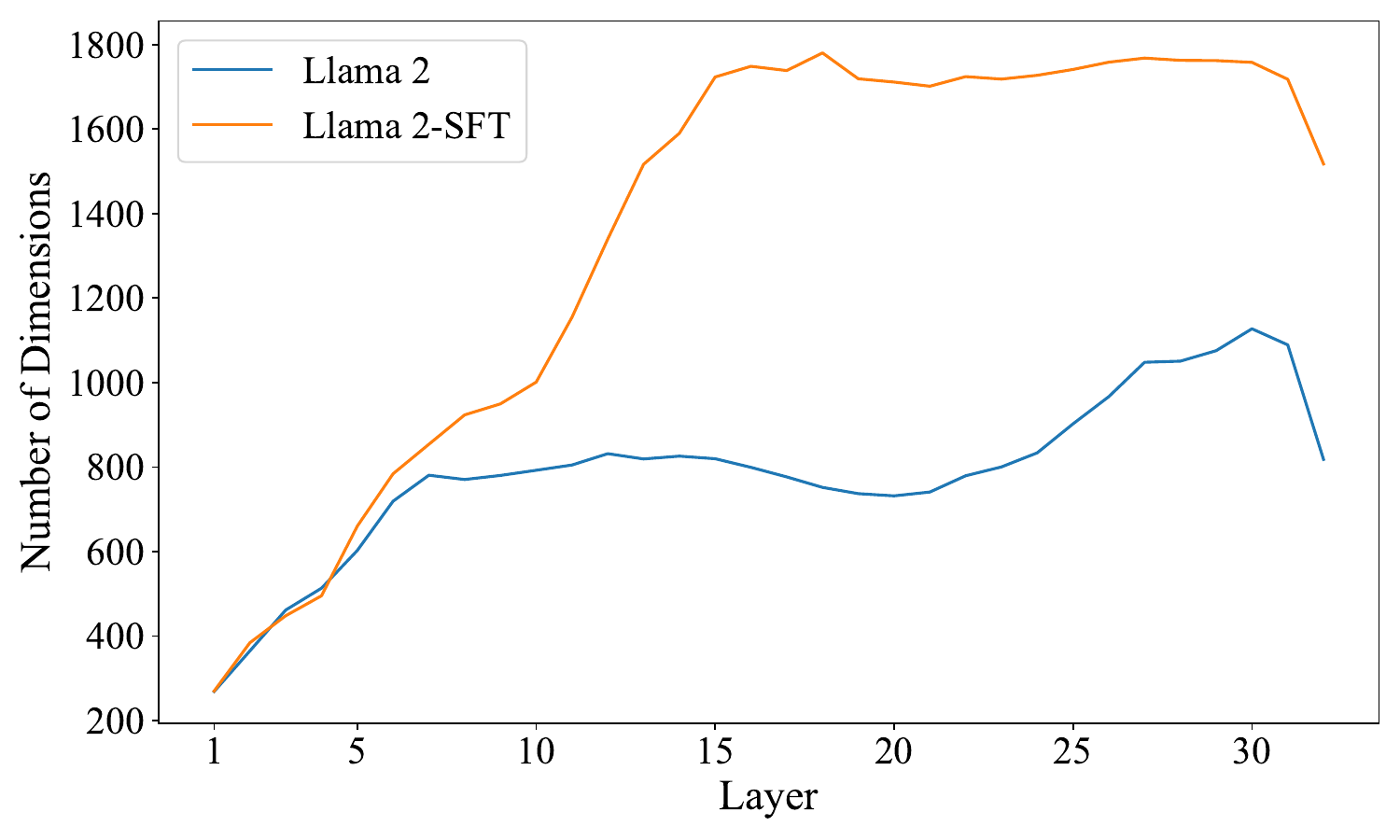}    
    \caption{Average number of dimensions required to explain 99\% of the representational variance across all tasks, as a function of the layer number.}
    \label{fig:variance_explained}
\end{figure}

\begin{figure*}[th]
     \centering
     \begin{subfigure}[b]{\linewidth}
         \centering
         \includegraphics[width=\textwidth]{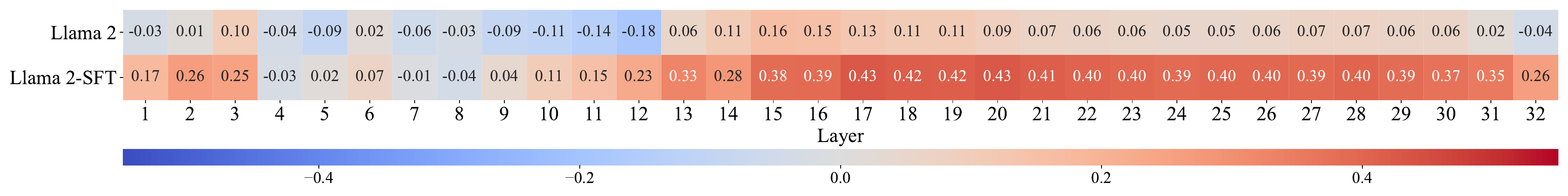}
         \vspace{-4ex}
         \caption{Flesch–Kincaid grade level}
         \label{fig:corr_cka_fk}
     \end{subfigure}
    \hfill
    \vspace{0.2ex}
     \begin{subfigure}[b]{\linewidth}
         \centering
         \includegraphics[width=\textwidth]{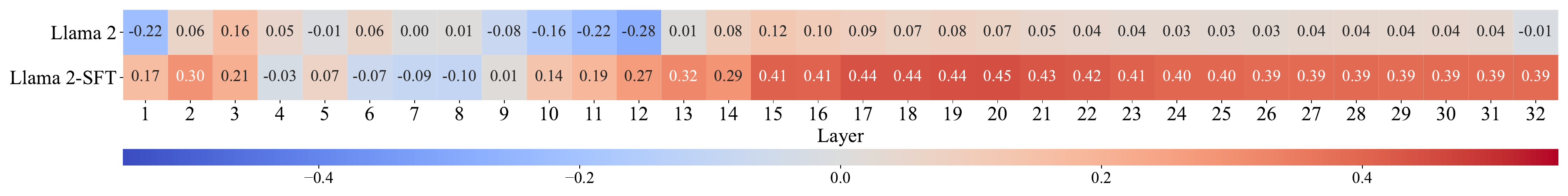}
         \vspace{-4ex}
         \caption{Coleman-Liau index}
         \label{fig:corr_cka_coleman}
     \end{subfigure}
     \hfill
     \vspace{0.2ex}
     \begin{subfigure}[b]{\linewidth}
         \centering
         \includegraphics[width=\textwidth]{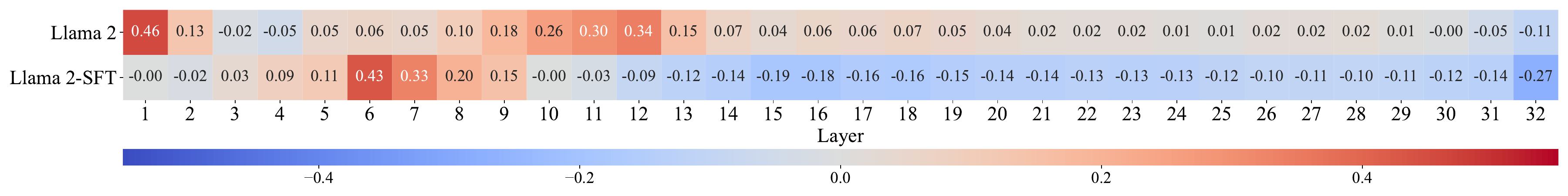}
         \vspace{-4ex}
         \caption{Data size}
         \label{fig:corr_test_size_cka}
     \end{subfigure}
    \caption{Pearson correlation results between the CKA similarities for all tasks, their reading difficulty, and data size across all layers. Higher values in reading difficulty measures correspond to greater reading difficulty.}
    \label{fig:corr_cka_fk_diff_perf}
\end{figure*}

\paragraph{Examining individual task clusters} Figures~\ref{fig:cka_all_layers_all_clusters_example} demonstrates that for tasks that are not well encoded in the pre-trained Llama 2 (e.g., structured data to text generation, translation), the CKA similarities from the instruction-tuned Llama 2-SFT remained high throughout all transition and refinement layers (10-32). Instruction tuning for these tasks induced significant representational shifts, adapting the model's internal structure to meet their specific demands. This aligns with prior work \cite{aghajanyan-etal-2021-muppet} showing that tasks requiring more sophisticated reasoning and modeling benefit greatly from task-specific tuning of pre-trained language models.

\subsection{Representation Clustering and Variance Analysis}
\label{sec:tsne_and_pca}

To further investigate representational differences, we used t-SNE \cite{van2008visualizing} to visualize task clusters across layers. Figure~\ref{fig:tsne_selection} presents a representative selection of layers, including a shared layer (layer 1), transition layers (layers 10 and 15), and refinement layers (layers 20 and 32). The full results for all layers are provided in Appendix~\ref{app:additional_results_cka}. In the first layer, both Llama 2 and Llama 2-SFT exhibit similar clustering. However, as we move to the transition layers, from layers 10 to 15, the Llama 2-SFT model forms more distinct task clusters compared to the Llama 2 model. This is further evidence that instruction tuning transforms the representations towards task-specificity in the transition layers. This clustering becomes increasingly pronounced in refinement layers, highlighting the effectiveness of instruction tuning in differentiating task-specific information and enhancing the ability to specialize representations for different tasks.

To quantify these differences, we performed variance analysis on the representations. We sought to determine if the model's ability to retain a large amount of task-specific information for many tasks affects its representation complexity. We analyzed the number of principal components required to explain 99\% of the variance in representation matrices across layers. The average number of components over all tasks is presented in Figure~\ref{fig:variance_explained}. In the shared layers, both Llama 2 and Llama 2-SFT models require a similar number of dimensions. Then, in the transition layers, Llama 2-SFT model begins to require more dimensions, suggesting it captures more complex task-specific information. This further demonstrates that the transition layers are indeed the layers where the transition to the task-specific representations occurs.

\subsection{Assessing Task Specific Information via Readability}
\label{sec:task_property_analysis}

In the preceding sections, we observed that the Llama 2 model exhibited a high variance in the amount of task-specific information stored across different tasks. In contrast, the Llama 2-SFT model demonstrated a low variance, storing a high level of task-specific information in its transition and refinement layers. While the Llama 2-SFT model exhibited low variance, we aimed to investigate the task priorities within the representation and identify features that could predict it. Previous research by \citet{zhao-etal-2022-understanding} has shown that when masked language models, such as BERT \cite{devlin-etal-2019-bert}, are trained on data from multiple domains, they tend to allocate their parameters to store domain-specific information. Unlike our approach, which examines instruction-level representations using the last token of an instruction, their study used the MOSSA method to analyze contextualized word embeddings, allowing them to focus on domain-specific words. We followed a similar analysis to examine task-specific information, which is strongly related to domain-specific information (as tasks can be viewed as domains). We used readability as a proxy for domain-specific information, relying on the finding by \citet{pitler-nenkova-2008-revisiting} that texts with more domain-specific and less commonly used words tend to have lower readability, resulting in higher reading difficulty scores.

We used two highly popular reading difficulty measures: the Flesch-Kincaid grade level score \cite{kincaid1975derivation} and the Coleman-Liau Index \cite{coleman1975computer}. The Flesch-Kincaid score assesses text readability based on factors like average sentence length and syllables per word, with lower scores indicating easier reading. Similarly, the Coleman-Liau Index estimates the required reading grade level based on characters, words, and sentences, with higher values corresponding to greater difficulty. We performed Pearson correlation analyses between CKA similarity and reading difficulty measures for all tasks across all layers. Specifically, we first calculated the readability measure for each input instruction, then obtained CKA similarities for representations from each layer. Finally, we computed the Pearson correlation coefficients between each input’s readability measure and the corresponding CKA similarities from each layer.

As illustrated in Figure~\ref{fig:corr_cka_fk}, we found a positive correlation between CKA similarity and the Flesch-Kincaid score for Llama 2-SFT. This correlation rapidly increases between layer 10 and layer 15 (the transition layers) and then saturates. These transitional layers are where task specialization transformations occur, as discussed earlier. This correlation is much weaker for the Llama 2 model. A similar pattern is observed with the Coleman-Liau Index, as shown in Figure~\ref{fig:corr_cka_coleman}. These findings suggest that instruction-tuned models encode more information for tasks with more task-specific vocabulary, as measured by their texts' readability indices. These findings thus suggest that instruction-tuned models encode and preserve task-specific information in the transition layers and retain it through the refinement layers, complementing our earlier findings.
Moreover, we previously noted that one of the advantages of CKA, compared to other similarity metrics, is its minimal requirement for a large number of data points in the analysis. To verify this, we conducted a correlation analysis between data size and CKA similarity, with the results presented in Figure~\ref{fig:corr_test_size_cka}. The analysis revealed no clear correlation between data size and CKA similarities, indicating that the number of data points used for CKA per task does not impact the CKA similarity. 

\begin{figure*}[ht]
    \centering
    \includegraphics[width=\linewidth]{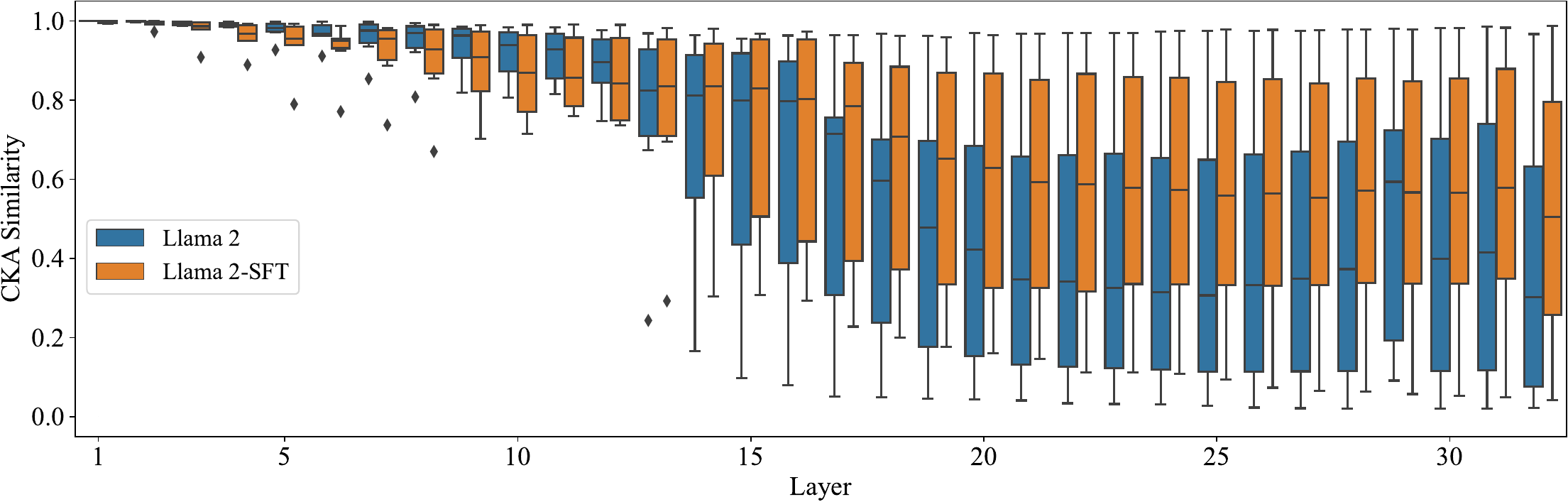}       
\caption{Distribution of CKA similarities across all layers for the pre-trained Llama 2 model and the instruction-tuned Llama 2-SFT model on unseen tasks.}
\label{fig:cka_all_layers_unseen}
\end{figure*}

\subsection{Evaluating Representations on Unseen Tasks}
\label{sec:eval_on_unseen_tasks}

While our previous analyses focused on evaluating representations against models trained on the same task data, it is crucial to examine how well our findings generalize to unseen tasks. To investigate this, we held out a set of seven tasks, including conversational question answering, question classification, math problems, linguistic acceptability, and word sense disambiguation (details in Appendix~\ref{app:dataset}). Our instruction-tuned models had no exposure to any of these seven tasks during training.

The CKA similarity results in Figure~\ref{fig:cka_all_layers_unseen} reveal an interesting pattern. For the lower layers (up to layer 12), the Llama 2 model exhibited slightly higher CKA similarities than Llama 2-SFT for several tasks, similar to what we find in \S\ref{sec:impact_of_instruct}. This indicates that while the Llama 2-SFT model was not trained using these tasks, it produced more divergent representations in lower layers and thus more general than the ones produced by Llama 2  (we refer the reader to shared layers discussion in \S\ref{sec:impact_of_instruct} for more details). However, as we move to the middle and higher layers responsible for encoding more specialized, task-specific knowledge, the Llama 2-SFT model began matching and ultimately surpassing the CKA similarities of the Llama 2 model. We can also see high variances between task similarities for both models, showing that we can not identify transition layers for Llama 2-SFT in this setup, just shared and refinement layers. These findings suggest that in addition to being trained on instructions, instruction-tuned models benefit from more general and thus better feature representations in their lower layers, which boost their performance for unseen instruction-based tasks compared to pre-trained LLMs.

\section{Discussion}
\label{sec:discussion}

Our study offers comprehensive insights into the impact of instruction tuning on the representations learned by LLMs. Previous work has discussed the benefits of instruction tuning \cite{wei-etal-2022-emergent,chung2022scaling,longpre-etal-2023-flan}, but ours is the first to analyze their effects from a representational perspective.

Our analysis revealed that LLMs instruction-tuned on multiple tasks learned different representations in the lower layers compared to LLMs tuned on individual tasks. Similar to MTL, such representations can be shared and used across tasks \cite{maurer-etal-2016-benefit}. Our analysis uncovered a key novel finding – we observed clear differences between pre-trained and instruction-tuned models, with the most significant representational transformations occurring in the middle transitional layers. This finding highlights the critical role of middle layers in encoding the specialized task knowledge induced by instruction tuning. Similarly, previous studies in multilingual settings have also identified language-neutral transformations in the middle layers of the network \cite{muller-etal-2021-first,zhao-etal-2023-joint}. Furthermore, our analysis suggests that in the refinement layers, instruction-tuned models continue to shape representations toward specific tasks but without substantial representational changes with respect to task-specific information. Overall, our finding about functionality for different layers in LLMs generally aligns with previous findings on BERT, which have shown that lower layers are more general, while upper layers are known to be more task-specific \cite{rogers-etal-2020-primer,merchant-etal-2020-happens}.

Our correlation analysis also revealed insights into the relationship between representations and task complexity. Instruction-tuned models exhibited a positive correlation with reading complexity measures in the transition and refinement layers, suggesting better encoding of task-specific information for tasks with more specific vocabulary – a capability not observed in pre-trained models. Notably, instruction tuning enabled models to preserve and enhance task-specific information across a broader range of layers, as evidenced by higher CKA similarities compared to control models. Our evaluation of unseen tasks further underscored the benefits of instruction tuning for improving generalization, with instruction-tuned models outperforming their pre-trained counterparts in deeper layers responsible for encoding complex task knowledge. This aligns with empirical evidence from \citet{wei-etal-2022-finetuned} but also highlights how representational changes facilitated by instruction tuning strengthen cross-task transfer capabilities.

\section{Conclusion}

Our study used several analyses to investigate how instruction tuning shapes representations in LLMs. These analyses revealed that unlike the pre-trained LLM (Llama 2), the instruction-tuned model (Llama 2-SFT) retained a high amount of task-specific information for all tasks from the middle layers onward. Moreover, we were able to map the layers of Llama 2-SFT into three groups based on their functionality: shared layers (layers 1-9), transition layers (10-15), and refinement layers (16-32). In addition to expanding our understanding of LLMs, such mapping can greatly benefit future research in the fields of PEFT, MTL, and model compression. We also demonstrated that our mapping does not generalize to unseen tasks, revealing that a potential additional reason for the strong generalization capabilities of instruction-tuned models to unseen tasks can be related to their multi-task nature of producing more general representations.

\section*{Limitations}

While our study provides valuable insights into the impact of instruction tuning on the representations learned by LLMs, there are several limitations that should be considered. 

Firstly, the instruction tuning in our experiments was implemented using LoRA instead of full fine-tuning. While LoRA is computationally efficient and effective in many scenarios, it may not capture the full range of representational changes that full fine-tuning can achieve. This limitation might have influenced the depth of insights into how instruction tuning affects the model representations.

Secondly, our study exclusively used the Llama 2 model due to limited computational resources available. Although Llama 2 is a powerful and widely used LLM, relying on a single model limits the generalizability of our findings. Different models may exhibit varied representational dynamics and responses to instruction tuning. Expanding our analysis to include multiple models from different architectures would provide a more comprehensive understanding of these effects.

Additionally, we conducted our experiments on the 7B parameter version of Llama 2. While this model size is substantial, it is not the largest available. Larger models, with their greater capacity and potentially different representational capabilities, might show different patterns in response to fine-tuning. Investigating multiple model sizes would help ascertain whether the observed trends hold across different scales.

Moreover, our experiments focused solely on NLP tasks and did not explore fine-tuning on code or other specialized domains. Coding tasks often involve unique representational challenges and might reveal different insights into the impact of fine-tuning. Including such tasks in future work would broaden the scope and applicability of our findings.

\section*{Acknowledgments}
This work was supported by the UKRI Centre for Doctoral Training (CDT) in Natural Language Processing through UKRI grant EP/S022481/1. We would like to thank Bonnie Webber, Ivan Titov and the anonymous reviewers for their helpful feedback. We appreciate the use of computing resources through the CSD3 cluster at the University of Cambridge and the Baskerville cluster at the University of Birmingham.

% Bibliography entries for the entire Anthology, followed by custom entries
\bibliography{anthology,custom}
% Custom bibliography entries only
% \bibliography{custom}

\clearpage
\appendix
\section{Dataset Details}
\label{app:dataset}

This appendix provides a detailed overview of the datasets used in this study. We followed \citet{wei-etal-2022-finetuned} and organized all tasks into the following task clusters:

\begin{itemize}
    
    \item \textbf{Closed-book Question Answering (QA)} requires models to answer questions about the world without direct access to the answer-containing information.    

    \item \textbf{Commonsense Reasoning} tests the capacity for physical or scientific reasoning infused with common sense.    
    
    \item \textbf{Coreference Resolution} identifies expressions referring to the same entity within a given text.

    \item \textbf{Natural Language Inference (NLI)} focuses on the relationship between two sentences, typically evaluating if the second sentence is true, false, or possibly true based on the first sentence.
    
    \item \textbf{Paraphrase Detection} involves evaluating if two sentences have the same meaning. While it can be considered a form of bidirectional entailment, it remains distinct from NLI in academic contexts.
    
    \item \textbf{Reading Comprehension} assesses the ability to answer questions based on a given passage containing the necessary information.

    \item \textbf{Reading Comprehension with Commonsense} merges the tasks of reading comprehension and commonsense reasoning.

    \item \textbf{Sentiment Analysis} is a traditional NLP task that determines whether a text expresses a positive or negative sentiment.
     
    \item \textbf{Struct-to-Text} involves generating natural language descriptions from structured data.
    
    \item \textbf{Translation} is the task of translating text from one language to another.

    \item \textbf{Summarization} involves creating concise summaries from longer texts.
    
    \item \textbf{Unseen} clusters uses the original miscellaneous task cluster from \citet{wei-etal-2022-finetuned} which includes:
    \begin{enumerate}
        \item Conversational question-answering;
        \item Evaluating context-sentence word meanings;
        \item Linguistic acceptability;
        \item Math questions;
        \item Question classification.
    \end{enumerate}
\end{itemize}

We provide tasks contained in each cluster in Table~\ref{tab:dataset_details}.

\begin{table*}[ht]
    \small
    \centering
    \begin{tabular}{ll|ll}
        \toprule
        \textbf{Task Cluster} & \textbf{Dataset} & \textbf{Task Cluster} & \textbf{Dataset} \\
        \midrule
        \multirow{7}{*}{\shortstack[l]{Natural language\\inference}} & ANLI  & \multirow{6}{*}{\shortstack[l]{Reading\\comprehension}} & BoolQ  \\
         & CB & & DROP  \\
         & MNLI  & & MultiRC  \\
         & QNLI  & & OBQA  \\
         & SNLI  & & SQuADv1 \\
         & WNLI  & & SQuADv2 \\
         & RTE  & & \\
        \midrule
        \multirow{4}{*}{\shortstack[l]{Commonsense\\reasoning}} & COPA  & \multirow{4}{*}{\shortstack[l]{Sentiment\\analysis}} & IMDB \\
         & HellaSwag  & & Sentiment140  \\
         & PiQA  & & SST-2  \\
         & StoryCloze  & & Yelp  \\
        \midrule
        \multirow{4}{*}{\shortstack[l]{Closed-book\\QA}} & ARC  & \multirow{3}{*}{\shortstack[l]{Paraphrase\\detection}} & MRPC \\
         & NQ  & & QQP  \\
         & TriviaQA  & & Paws Wiki \\
         & & & STS-B \\
        \midrule
        \multirow{3}{*}{\shortstack[l]{Coreference\\resolution}} & DPR  & \multirow{2}{*}{\shortstack[l]{Reading\\comprehension\\with commonsense}} & CosmosQA  \\
         & Winogrande  & & ReCoRD  \\
         & WSC273 & & \\
        \midrule
        \multirow{4}{*}{\shortstack[l]{Struct to text}} & CommonGen  & \multirow{3}{*}{\shortstack[l]{Translation}} & En--Fr from WMT'14  \\
         & DART  & &  WMT'16  \\
         & E2ENLG & & En--Es from Paracrawl  \\
         & WebNLG  & & \\
        \midrule
        \multirow{11}{*}{\shortstack[l]{Summarization}} & AESLC  & \multirow{6}{*}{\shortstack[l]{Unseen}} & CoQA  \\
         & CNN-DM  & & QuAC  \\
         & Gigaword  & & WiC  \\
         & MultiNews  & & TREC  \\
         & Newsroom  & & CoLA  \\
         & Samsum  & & Math questions \\
         & XSum & & \\
         & AG News & & \\
         & Opinion Abstracts - Rotten Tomatoes & & \\
         & Opinion Abstracts - iDebate & & \\
         & Wikilingua English & & \\
        \bottomrule
    \end{tabular}
    \caption{Dataset details grouped by task clusters. For WMT'16, we include En--De, En--Tr, En--Cs, En--Fi, En--Ro, and En--Ru translation pairs. For all details about each dataset including the dataset size, please refer to \citet{wei-etal-2022-finetuned}.}
    \label{tab:dataset_details}
\end{table*}

\section{Additional Results}

\subsection{Results on Model Evaluation}
\label{app:additional_results_model_eval}
We provide the results on all control models and instruction-tuned Llama 2-SFT in Table~\ref{tab:dataset_details_results_nlu} (for natural language understanding tasks) and Table~\ref{tab:dataset_details_results_nlg} (for natural language generation tasks). To further evaluate the validness of our instruction tuning, we also benchmark our models on two popular benchmark datasets: MMLU \cite{hendrycks-etal-2020-mmlu} and BBH \cite{suzgun2022challenging}. We provide results in Table~\ref{tab:mmlu_bbh}. We can see that Llama 2-SFT outperforms Llama 2 on both of these benchmarks. 

\begin{table}[h]
\centering
\begin{tabular}{lll}
\toprule
\textbf{}         & \textbf{MMLU}     & \textbf{BBH} \\ \toprule
Llama 2           &    41.25        & 32.82    \\
Llama 2-SFT     & 47.81          & 37.49     \\ \midrule
\end{tabular}
\caption{Results for Llama 2 and Llama 2-SFT on MMLU and BBH. We use a 0-shot evaluation for MMLU to assess our models. For BBH, we follow the default evaluation protocol and use a 3-shot evaluation.}
\label{tab:mmlu_bbh}
\end{table}

\begin{table*}[ht]
    \small
    \centering
    \begin{tabular}{llcc}
        \toprule
        \multirow{2}{*}{\textbf{Dataset}} & \multirow{2}{*}{\textbf{Metric}} & \multicolumn{2}{c}{\textbf{Result}} \\
        & & Llama 2-SFT & Control Model \\
        \midrule
        \multicolumn{1}{l}{\textbf{\underline{Natural Language Inference}}} & & \\
        ANLI (r1) & Accuracy & 51.87 & 54.45 \\
        ANLI (r2) & Accuracy & 49.45 & 55.85 \\
        ANLI (r3) & Accuracy & 47.48 & 54.14 \\
        CB & Accuracy & 49.59 & 83.17 \\
        MNLI (matched) & Accuracy & 87.25 &  88.64 \\
        MNLI (mismatched) & Accuracy & 87.72 & 89.41 \\
        QNLI & Accuracy & 83.00 & 86.46 \\
        SNLI & Accuracy & 82.96 & 84.06 \\
        WNLI & Accuracy & 71.22 & 69.64 \\
        RTE & Accuracy & 81.52 & 81.21 \\
        \midrule
        \multicolumn{3}{l}{\textbf{\underline{Reading Comprehension}}} \\
        BoolQ & Accuracy & 83.53 &  88.18 \\
        DROP & F1 & 44.42 & 52.05 \\
        MultiRC & F1 & 72.19 &  73.92 \\
        OBQA & Accuracy & 64.92 & 65.37 \\
        SQuADv1 & F1 & 73.91 & 74.24 \\
        SQuADv2 & F1 & 22.75 & 23.55 \\
        \midrule
        \multicolumn{3}{l}{\textbf{\underline{Commonsense Reasoning}}} \\
        COPA & Accuracy & 83.56 & 76.97 \\
        HellaSwag & Accuracy & 71.43 & 73.49 \\
        PiQA & Accuracy & 78.21 & 78.43 \\
        StoryCloze & Accuracy & 85.81 & 84.82 \\
        \midrule
        \multicolumn{3}{l}{\textbf{\underline{Sentiment Analysis}}} \\
        IMDB & Accuracy & 72.06 & 74.54 \\
        Sentiment140 & Accuracy & 45.52 & 44.53 \\
        SST-2 & Accuracy & 79.14 & 79.03 \\
        Yelp & Accuracy & 74.35 & 74.40 \\
        \midrule
        \multicolumn{3}{l}{\textbf{\underline{Closed-book QA}}} \\
        ARC (Challenge) & Accuracy & 59.09 & 52.83 \\
        ARC (Easy) & Accuracy & 67.18 & 65.72 \\
        TriviaQA & Accuracy & 59.00 & 59.26 \\
        NQ & Accuracy & 28.79 & 31.18 \\
        \midrule
        \multicolumn{3}{l}{\textbf{\underline{Paraphrase Detection}}} \\
        MRPC & Accuracy & 78.35 & 84.73 \\
        QQP & Accuracy & 84.91 & 87.37 \\
        PAWS Wiki & Accuracy & 91.77 & 94.15 \\
        STS-B & Accuracy & 47.46 & 51.20 \\
        \midrule
        \multicolumn{3}{l}{\textbf{\underline{Coreference Resolution}}} \\
        DPR & Accuracy & 85.12 & 72.53 \\
        Winogrande & Accuracy & 69.68 & 69.93 \\
        WSC273 & Accuracy & 55.78 & 47.24 \\
        \midrule
        \multicolumn{1}{l}{\textbf{\underline{Read. Comp. w/ Commonsense}}} & & \\
        CosmosQA & Accuracy & 66.60 & 69.36 \\
        ReCoRD & Accuracy & 85.13 & 85.78 \\
        \midrule
        \multicolumn{3}{l}{\textbf{\underline{Unseen}}} \\
        CoQA & Accuracy & 66.60 & 73.93 \\
        QuAC & Accuracy & 18.29 & 33.99 \\
        WiC & Accuracy & 56.47 & 70.77 \\
        TREC & Accuracy & 57.05 & 80.25 \\
        CoLA & Accuracy & 34.85 & 70.91 \\
        Math Questions & Accuracy & 4.43 & 35.50 \\
        \bottomrule
    \end{tabular}
    \caption{Performance metrics grouped by natural language understanding task clusters for Llama 2-SFT and control models (Llama 2 model individually fine-tuned on each task). ``Read. Comp. w/ Commonsense'' denotes reading comprehension with commonsense.}
    \label{tab:dataset_details_results_nlu}
\end{table*}

\begin{table*}[ht]
    \small
    \centering
    \begin{tabular}{llcc}
        \toprule
        \multirow{2}{*}{\textbf{Dataset}} & \multirow{2}{*}{\textbf{Metric}} & \multicolumn{2}{c}{\textbf{Result}} \\
        & & Llama 2-SFT & Control Model \\
        \midrule
        \multicolumn{3}{l}{\textbf{\underline{Struct-to-Text}}} \\
        CommonGen & ROUGE-L & 45.92 & 46.52 \\
        DART & ROUGE-L & 55.46 & 57.28 \\
        E2ENLG & ROUGE-L & 50.17 & 50.96 \\
        WebNLG & ROUGE-L & 62.92 & 65.22 \\
        \midrule
        \multicolumn{3}{l}{\textbf{\underline{Translation}}} \\
        WMT'14 En--Fr & BLEU & 59.30 & 59.29 \\
        WMT'16 En--De & BLEU & 56.84 & 57.45 \\
        WMT'16 En--Tr & BLEU & 39.41 & 43.58 \\
        WMT'16 En--Cs & BLEU & 46.92 & 47.21 \\
        WMT'16 En--Fi & BLEU & 48.57 & 50.28 \\
        WMT'16 En--Ro & BLEU & 56.03 & 57.70 \\
        WMT'16 En--Ru & BLEU & 51.41 & 52.12 \\
        ParaCrawl En--Es & BLEU & 54.76 & 56.39 \\
        \midrule
        \multicolumn{3}{l}{\textbf{\underline{Summarization}}} \\
        AESLC & ROUGE-L & 29.98 & 31.68 \\
        CNN-DM & ROUGE-L & 17.38 & 19.59 \\
        Gigaword & ROUGE-L & 28.69 & 30.22 \\
        MultiNews & ROUGE-L & 15.17 & 16.61 \\
        Newsroom & ROUGE-L & 18.95 & 22.43 \\
        Samsum & ROUGE-L & 36.36 & 37.72 \\
        XSum & ROUGE-L & 25.51 & 29.57 \\
        AG News & ROUGE-L & 77.26 & 80.99 \\
        Opinion Abstracts - Rotten Tomatoes & ROUGE-L & 19.36 & 21.70 \\
        Opinion Abstracts - iDebate & ROUGE-L & 18.90 & 23.14 \\
        Wikilingua English & ROUGE-L & 30.22 & 32.18 \\
        \bottomrule
    \end{tabular}
    \caption{Performance metrics grouped by natural language generation task clusters for Llama 2-SFT and control models (Llama 2 model individually fine-tuned on each task).}
    \label{tab:dataset_details_results_nlg}
\end{table*}

\subsection{Results on Analysis}
\label{app:additional_results_cka}
\begin{figure*}[ht]
    \centering
    \begin{subfigure}{\textwidth}
        \includegraphics[width=\textwidth]{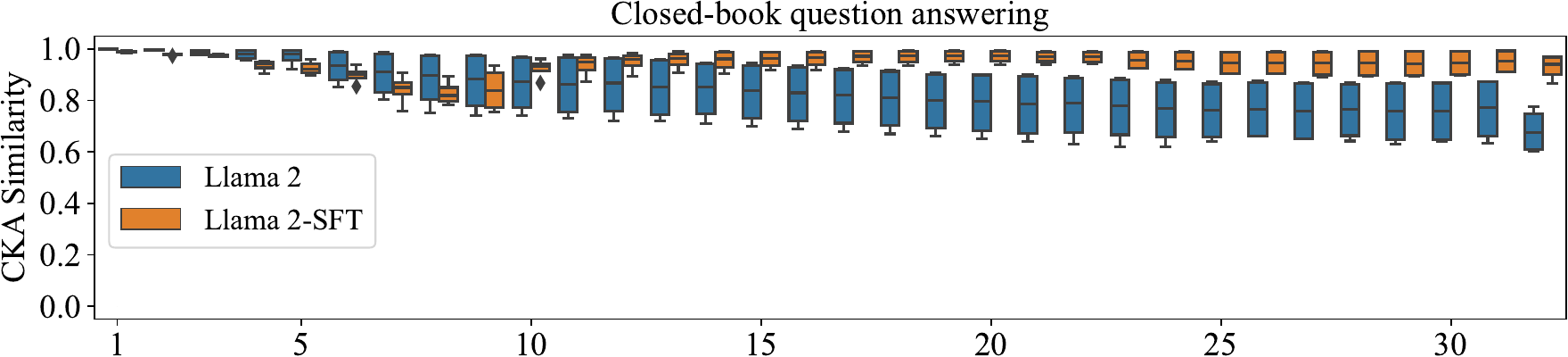}
        \vspace{-2ex}
    \end{subfigure}
    \begin{subfigure}{\textwidth}
        \includegraphics[width=\textwidth]{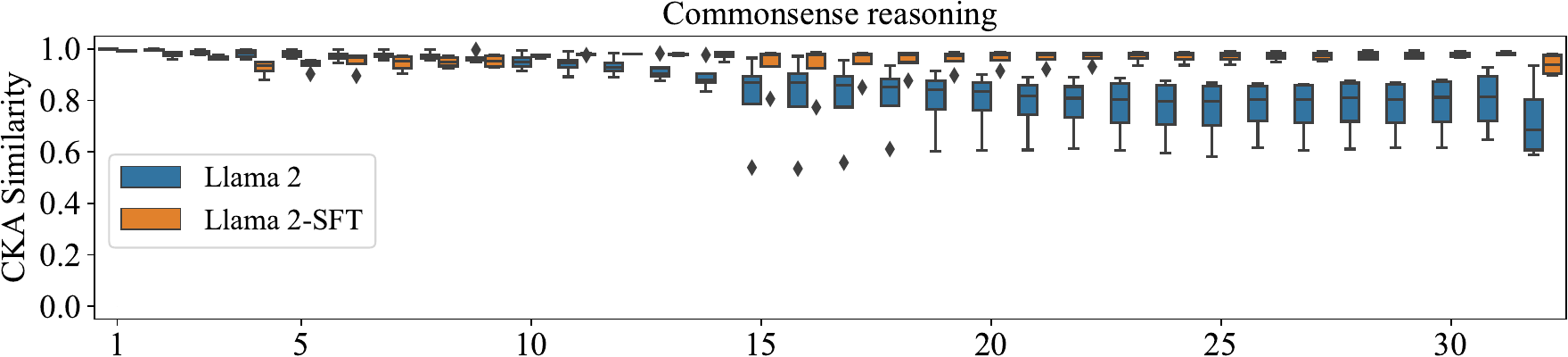}
        \vspace{-2ex}
    \end{subfigure}
    \begin{subfigure}{\textwidth}
        \includegraphics[width=\textwidth]{new_figures/llama2_vanilla_sft_cka_boxplot_wo_xlabel_coreference.pdf}
        \vspace{-2ex}
    \end{subfigure}
    \begin{subfigure}{\textwidth}
        \includegraphics[width=\textwidth]{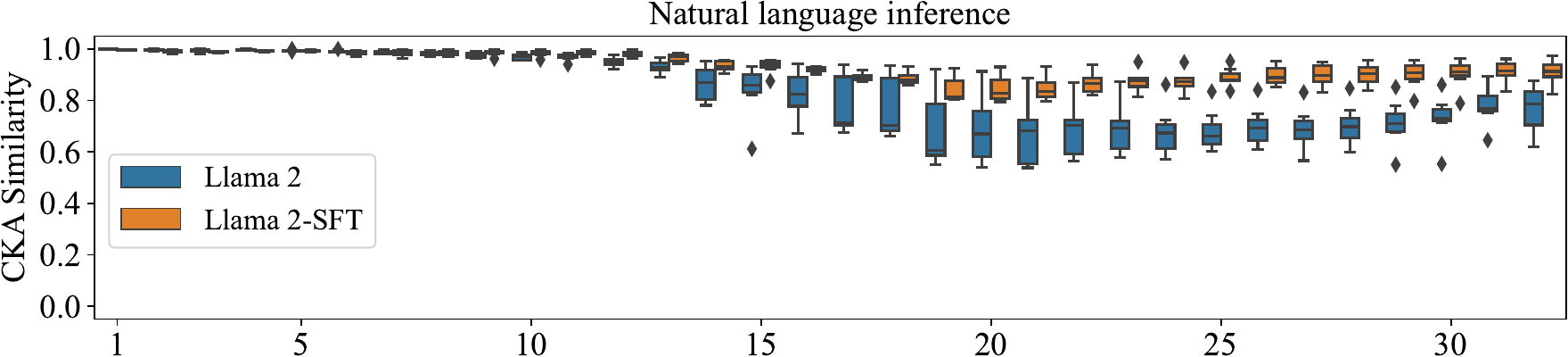}
        \vspace{-2ex}
    \end{subfigure}
    \begin{subfigure}{\textwidth}
        \includegraphics[width=\textwidth]{new_figures/llama2_vanilla_sft_cka_boxplot_wo_xlabel_paraphrase.pdf}
        \vspace{-2ex}
    \end{subfigure}
    \begin{subfigure}{\textwidth}
        \includegraphics[width=\textwidth]{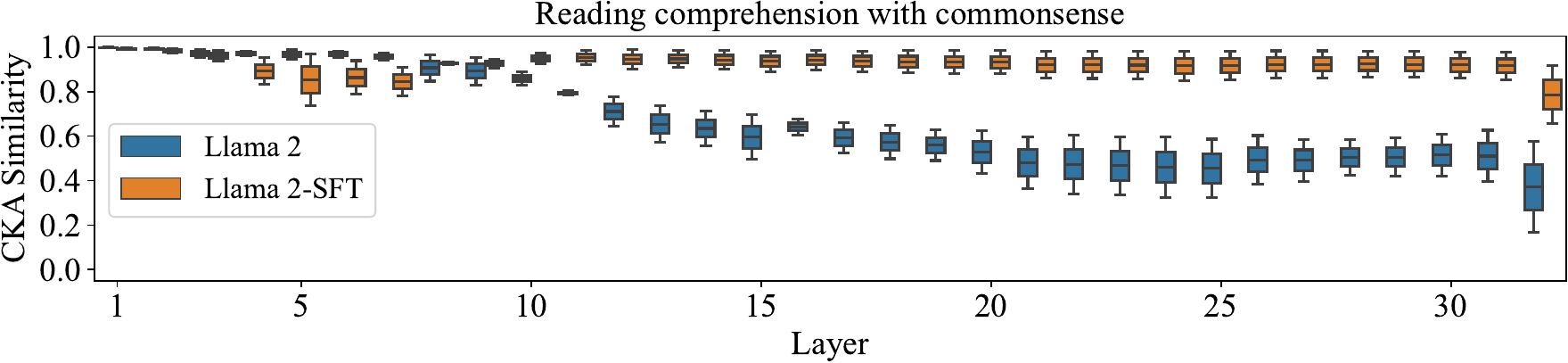}
        \vspace{-2ex}
    \end{subfigure}
\caption{Distribution of CKA similarities across all layers for the pre-trained Llama 2 model and the instruction-tuned Llama 2-SFT model, grouped by different task clusters.}
\label{fig:cka_all_layers_all_clusters_1}
\end{figure*}

\begin{figure*}[ht]
    \centering     
    \begin{subfigure}{\textwidth}
        \includegraphics[width=\textwidth]{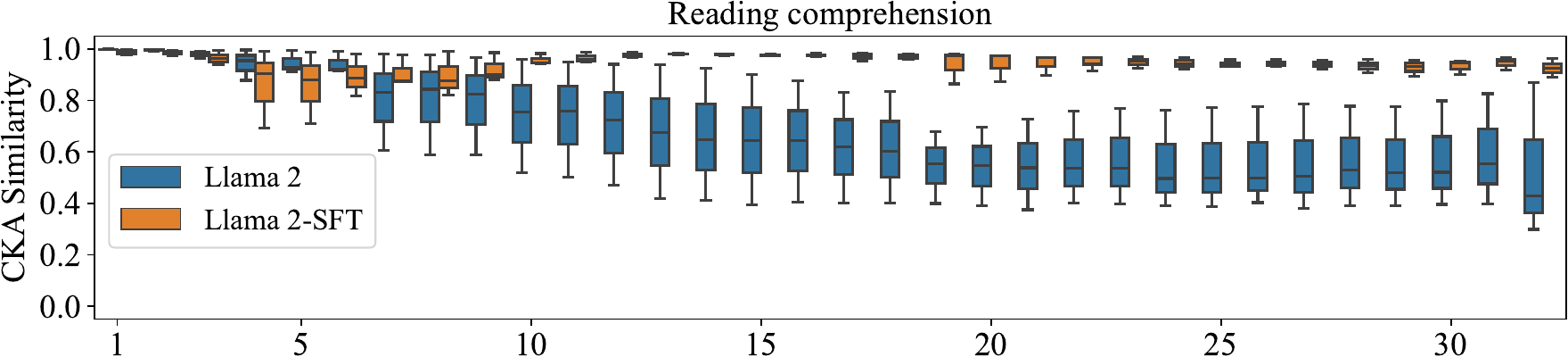}
        \vspace{-2ex}
    \end{subfigure}
    \begin{subfigure}{\textwidth}
        \includegraphics[width=\textwidth]{new_figures/llama2_vanilla_sft_cka_boxplot_wo_xlabel_sentiment.pdf}
        \vspace{-2ex}
    \end{subfigure}
    \begin{subfigure}{\textwidth}
        \includegraphics[width=\textwidth]{new_figures/llama2_vanilla_sft_cka_boxplot_wo_xlabel_struct_to_Text.pdf}
        \vspace{-2ex}
    \end{subfigure}
    \begin{subfigure}{\textwidth}
        \includegraphics[width=\textwidth]{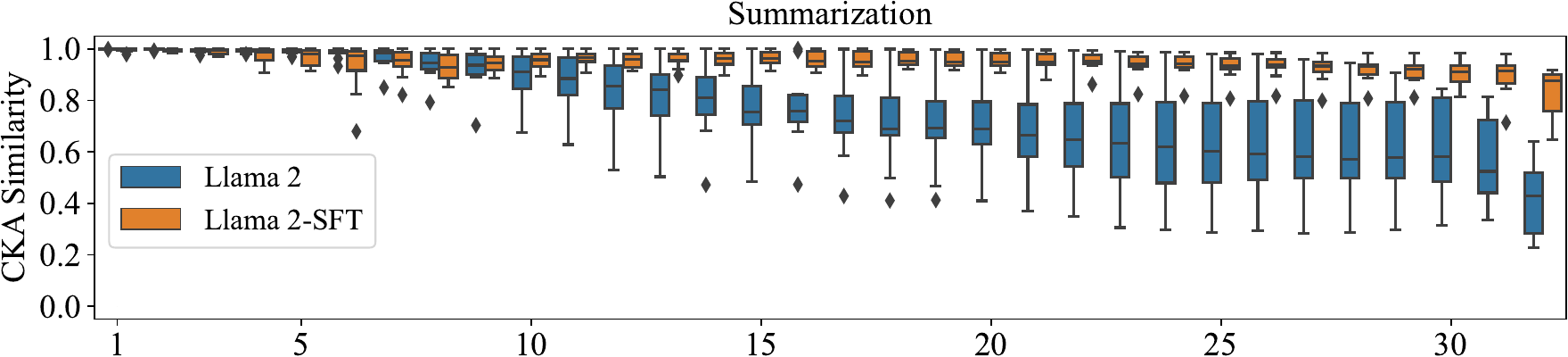}
        \vspace{-2ex}
    \end{subfigure}
    \begin{subfigure}{\textwidth}
        \includegraphics[width=\textwidth]{new_figures/llama2_vanilla_sft_cka_boxplot_w_xlabel_translation.pdf}
        \vspace{-2ex}
    \end{subfigure}
\caption{Distribution of CKA similarities across all layers for the pre-trained Llama 2 model and the instruction-tuned Llama 2-SFT model, grouped by different task clusters.}
\label{fig:cka_all_layers_all_clusters_2}
\end{figure*}

\begin{figure*}[!ht]
     \centering
     \begin{subfigure}[b]{\textwidth}
         \centering
         \includegraphics[width=\textwidth]{tsne_figures/legend.png}
         \vspace{-2ex}
     \end{subfigure}
     \begin{subfigure}[b]{0.19\textwidth}
         \centering
         \includegraphics[width=\textwidth]{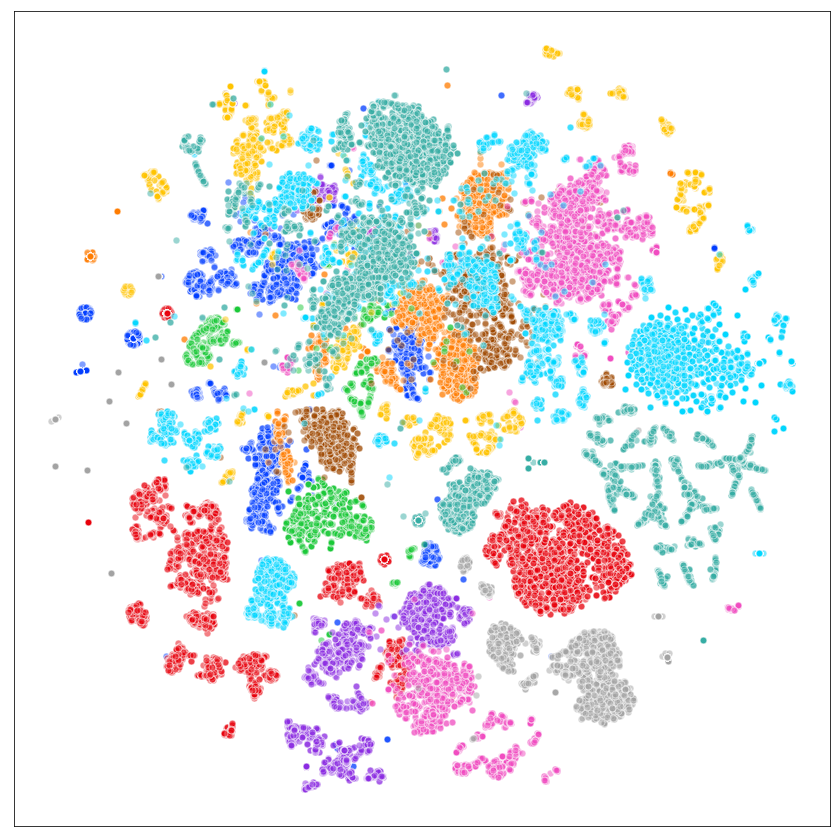}
         \caption*{Layer 1}
     \end{subfigure}
    \hfill
    \begin{subfigure}[b]{0.19\textwidth}
         \centering
         \includegraphics[width=\textwidth]{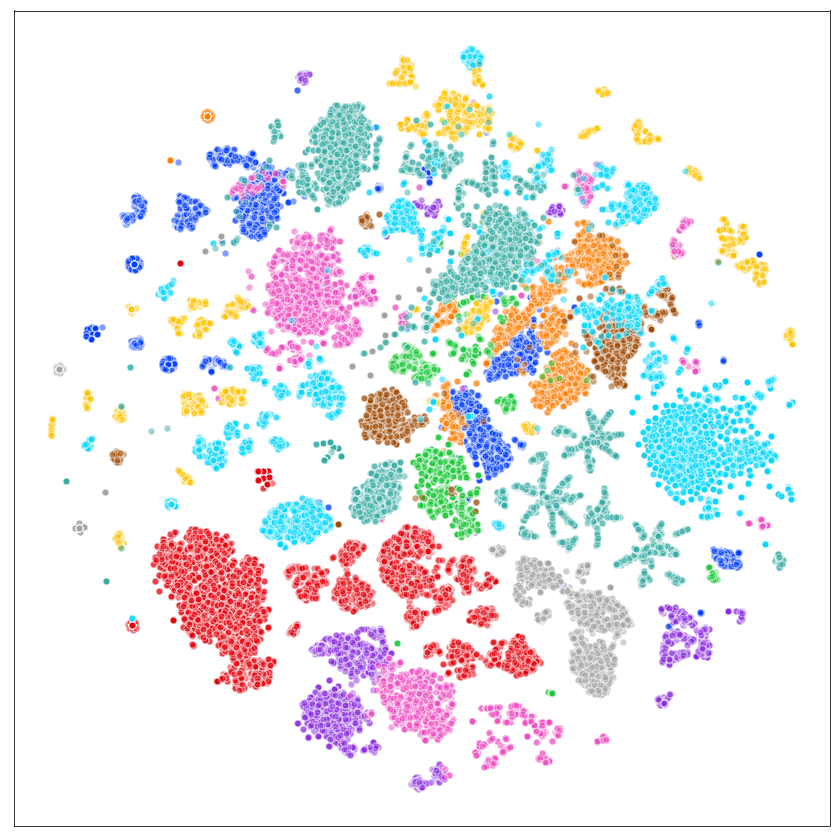}
         \caption*{Layer 2}
     \end{subfigure}
     \hfill
    \begin{subfigure}[b]{0.19\textwidth}
         \centering
         \includegraphics[width=\textwidth]{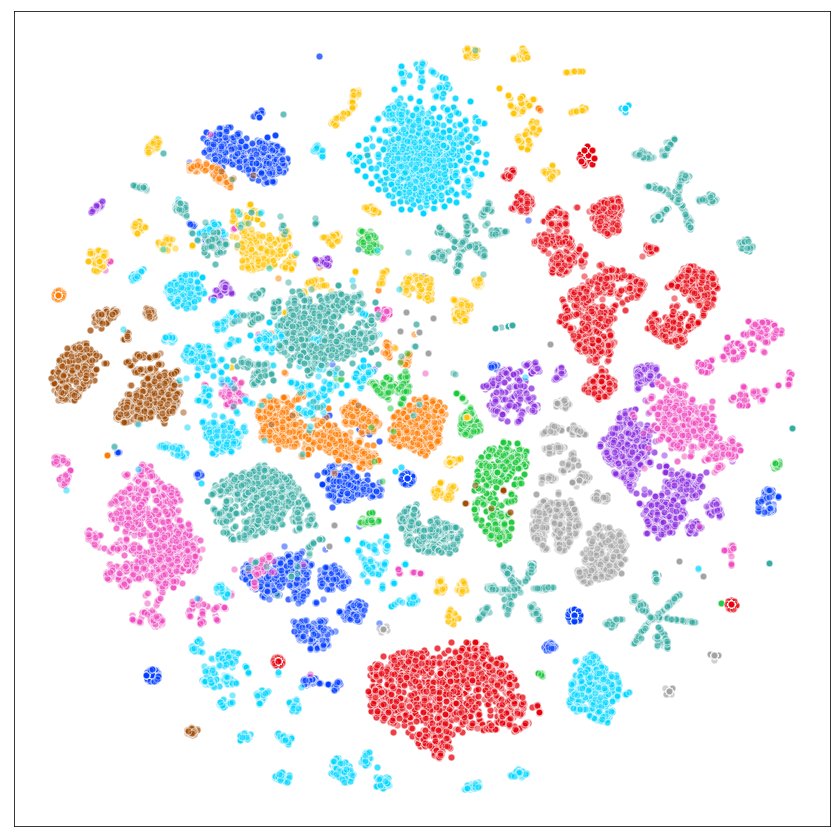}
         \caption*{Layer 3}
     \end{subfigure}
     \hfill
    \begin{subfigure}[b]{0.19\textwidth}
         \centering
         \includegraphics[width=\textwidth]{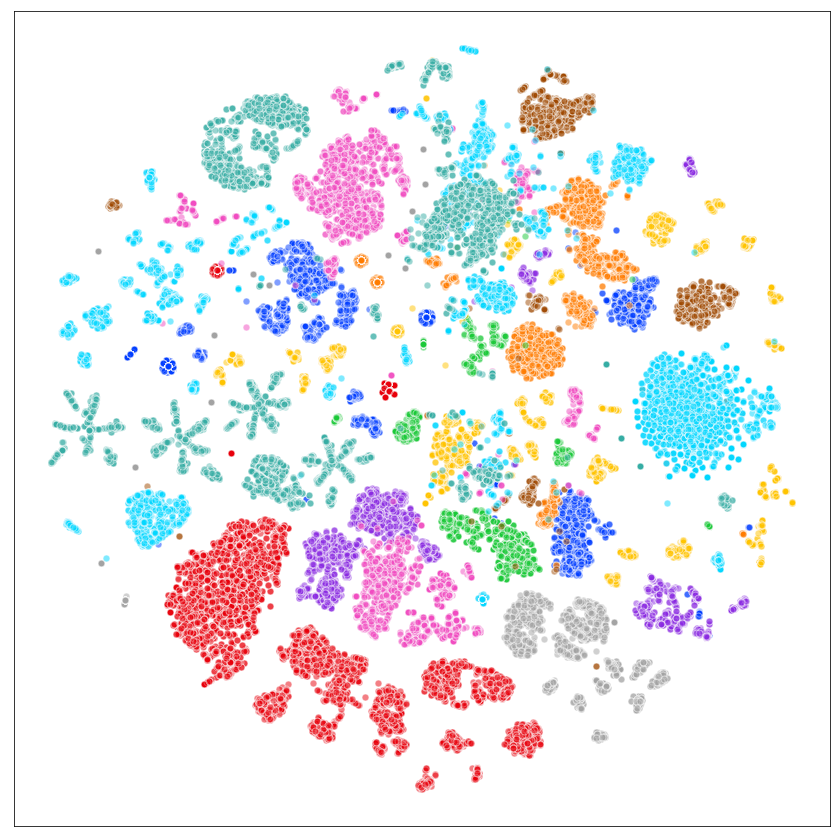}
         \caption*{Layer 4}
     \end{subfigure}
     \hfill
     \begin{subfigure}[b]{0.19\textwidth}
         \centering
         \includegraphics[width=\textwidth]{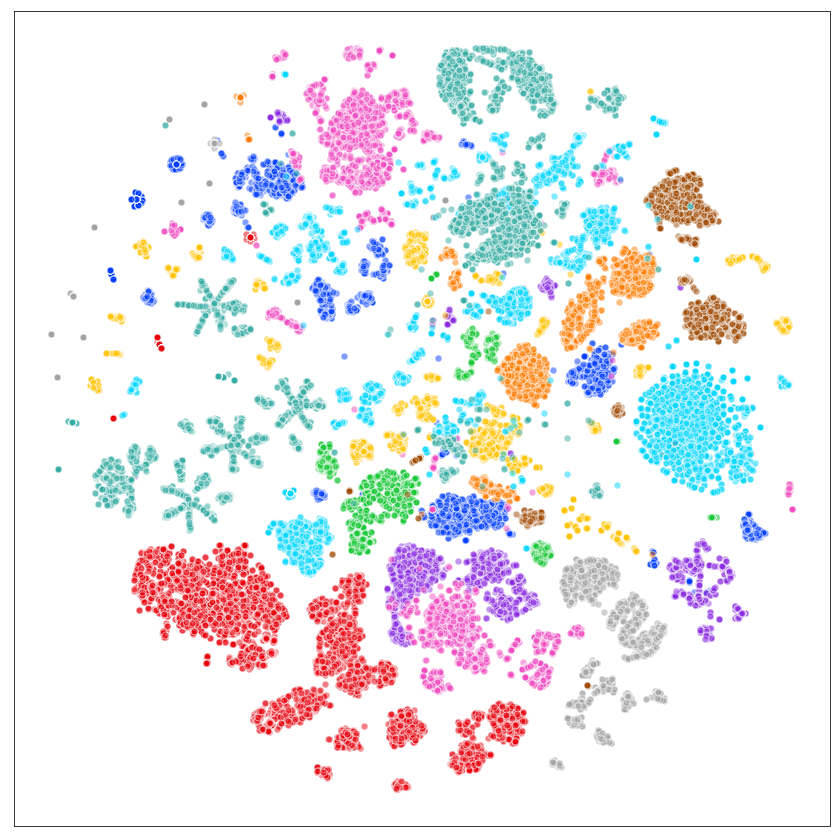}
         \caption*{Layer 5}
     \end{subfigure}
     \begin{subfigure}[b]{0.19\textwidth}
         \centering
         \includegraphics[width=\textwidth]{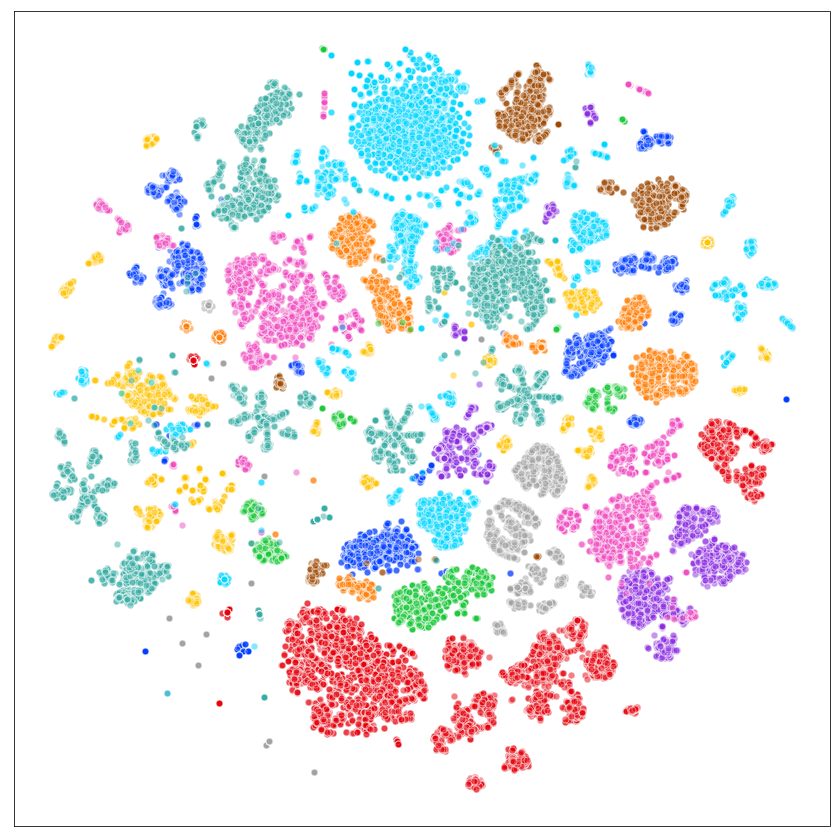}
         \caption*{Layer 6}
     \end{subfigure}
    \hfill
    \begin{subfigure}[b]{0.19\textwidth}
         \centering
         \includegraphics[width=\textwidth]{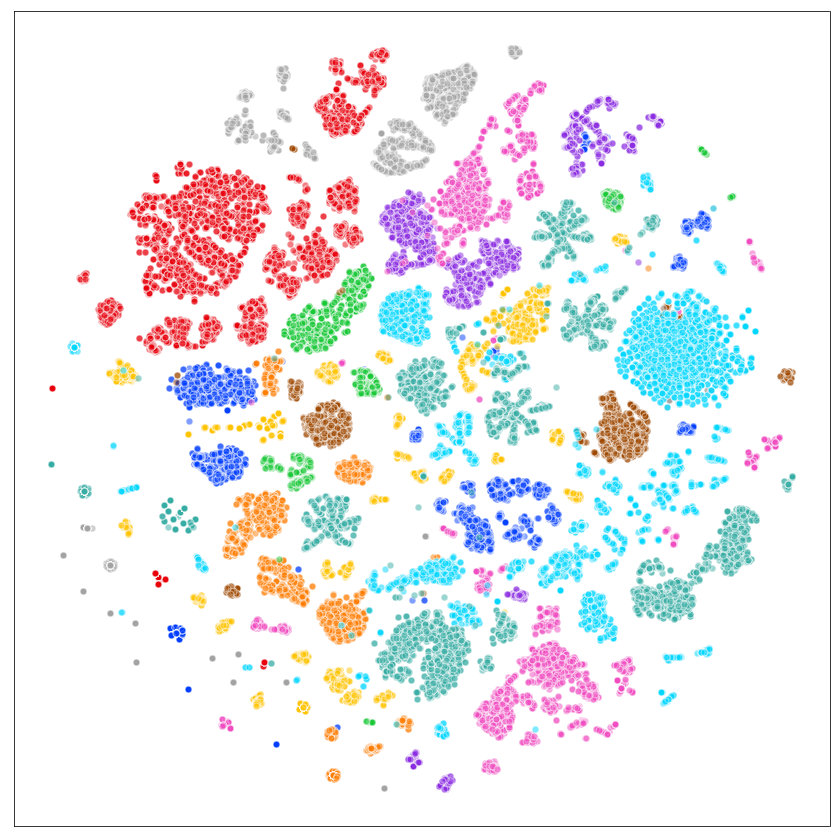}
         \caption*{Layer 7}
     \end{subfigure}
    \hfill
    \begin{subfigure}[b]{0.19\textwidth}
         \centering
         \includegraphics[width=\textwidth]{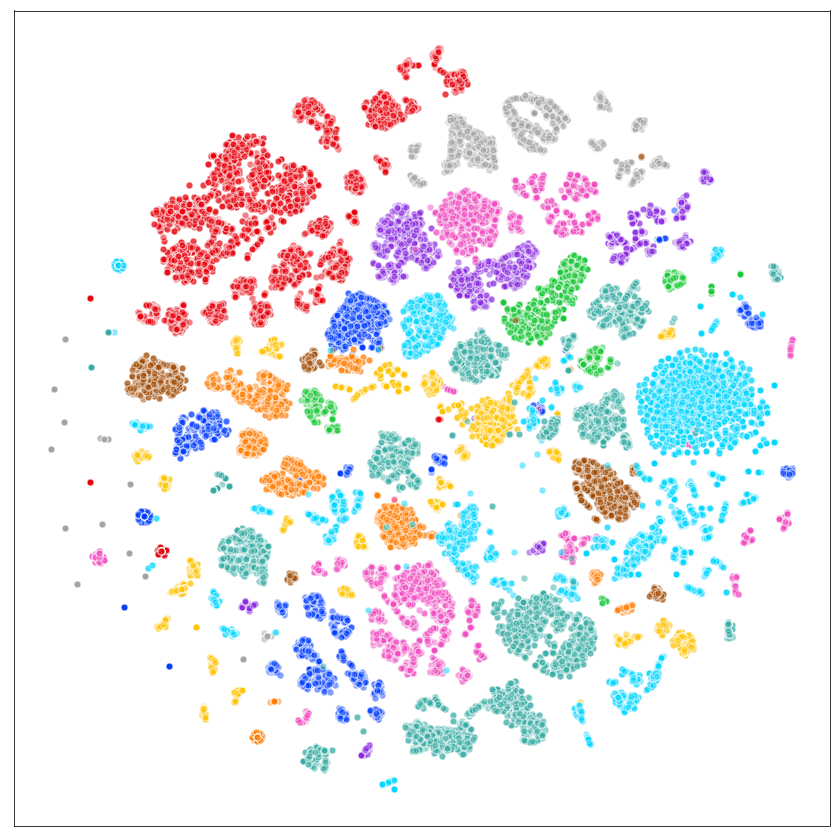}
         \caption*{Layer 8}
     \end{subfigure}
    \hfill
     \begin{subfigure}[b]{0.19\textwidth}
         \centering
         \includegraphics[width=\textwidth]{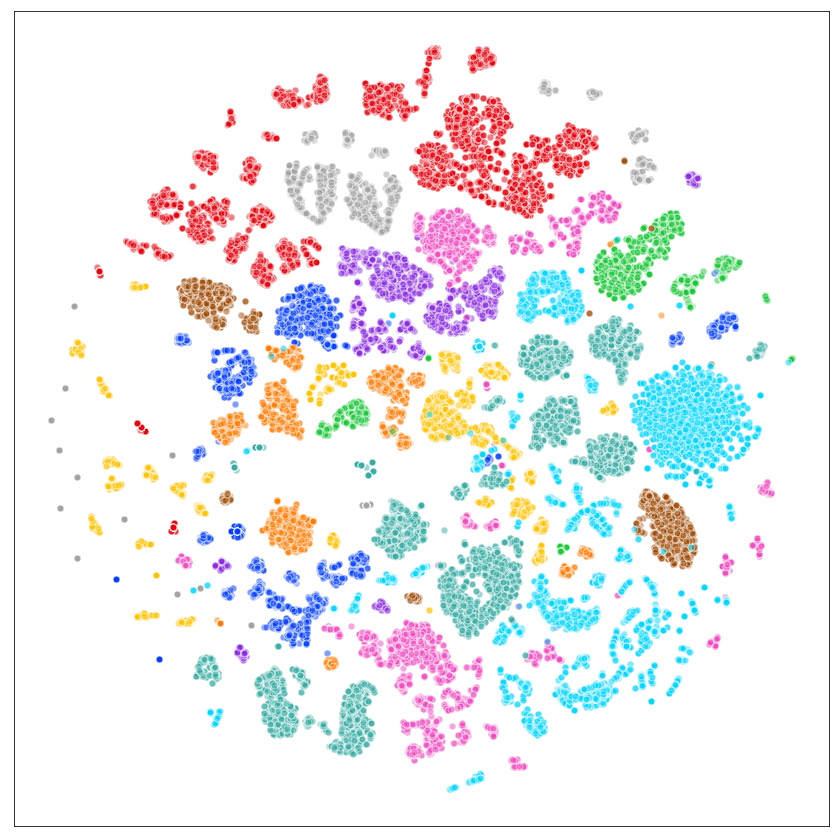}
         \caption*{Layer 9}
     \end{subfigure}
    \hfill
     \begin{subfigure}[b]{0.19\textwidth}
         \centering
         \includegraphics[width=\textwidth]{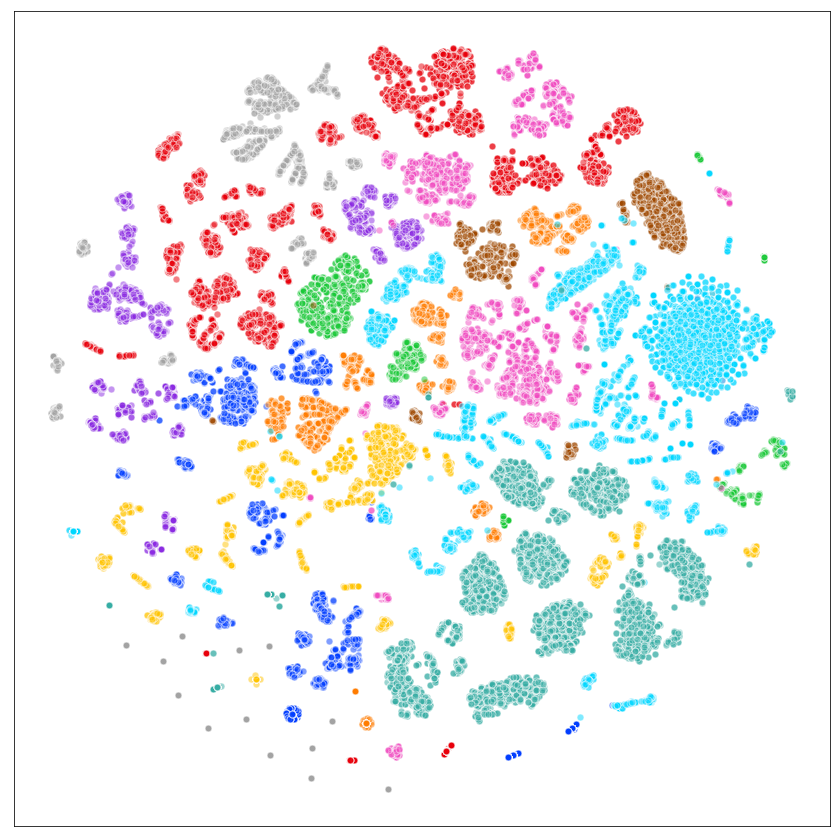}
         \caption*{Layer 11}
     \end{subfigure}
     \begin{subfigure}[b]{0.19\textwidth}
         \centering
         \includegraphics[width=\textwidth]{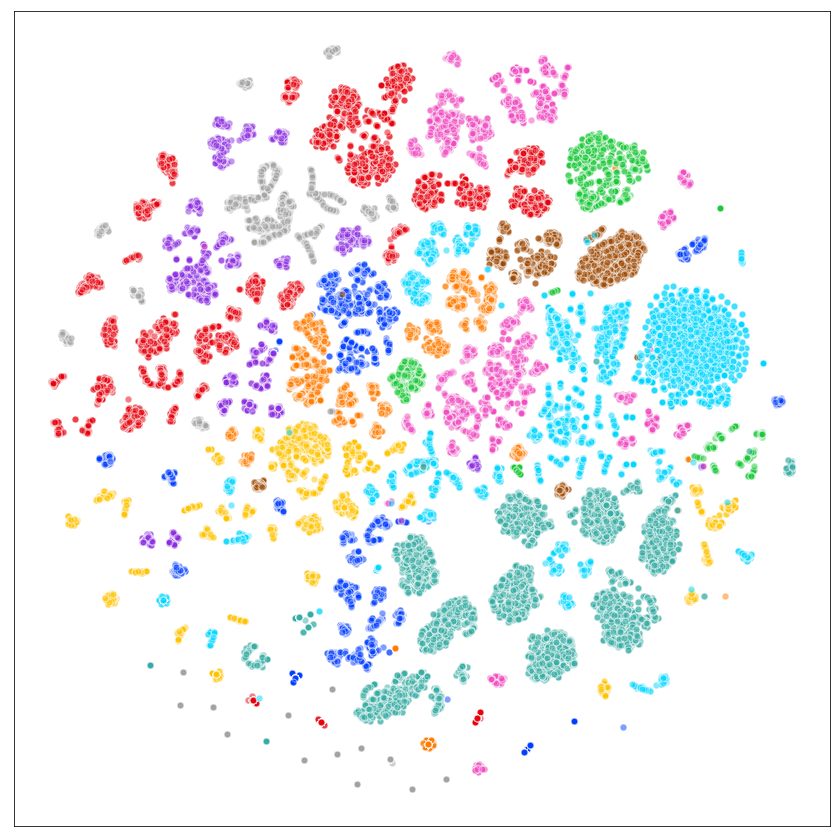}
         \caption*{Layer 12}
     \end{subfigure}
    \hfill
    \begin{subfigure}[b]{0.19\textwidth}
         \centering
         \includegraphics[width=\textwidth]{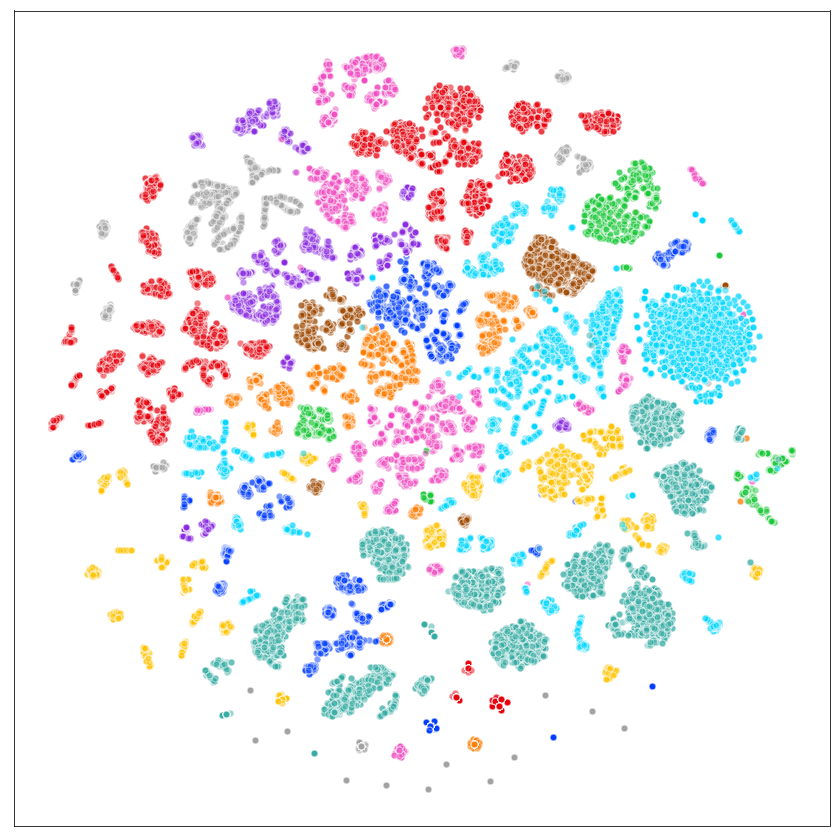}
         \caption*{Layer 13}
     \end{subfigure}
     \hfill
    \begin{subfigure}[b]{0.19\textwidth}
         \centering
         \includegraphics[width=\textwidth]{tsne_figures_low_res/tsne_llama2_last_token_by_cluster_l4.png}
         \caption*{Layer 14}
     \end{subfigure}
     \hfill
    \begin{subfigure}[b]{0.19\textwidth}
         \centering
         \includegraphics[width=\textwidth]{tsne_figures_low_res/tsne_llama2_last_token_by_cluster_l6.png}
         \caption*{Layer 16}
     \end{subfigure}
     \hfill
     \begin{subfigure}[b]{0.19\textwidth}
         \centering
         \includegraphics[width=\textwidth]{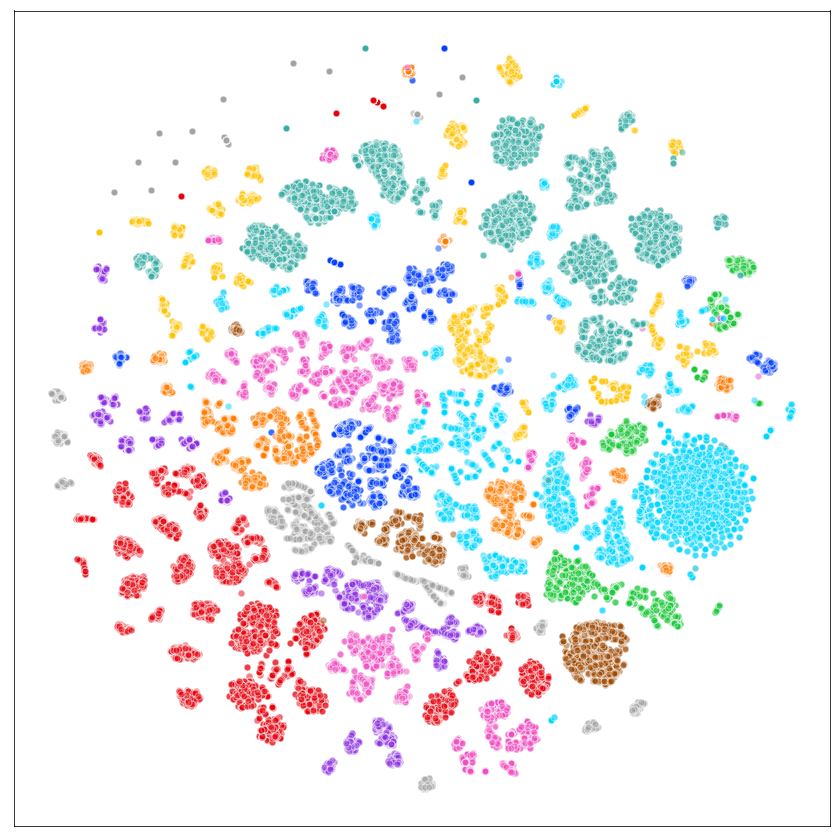}
         \caption*{Layer 17}
     \end{subfigure}
     \begin{subfigure}[b]{0.19\textwidth}
         \centering
         \includegraphics[width=\textwidth]{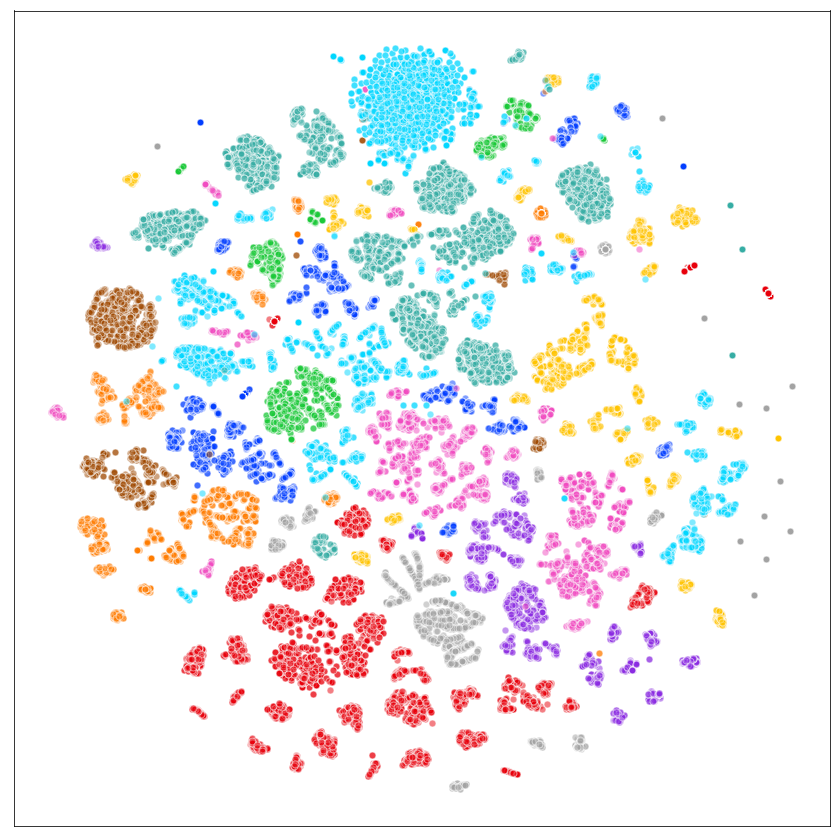}
         \caption*{Layer 18}
     \end{subfigure}
    \hfill
    \begin{subfigure}[b]{0.19\textwidth}
         \centering
         \includegraphics[width=\textwidth]{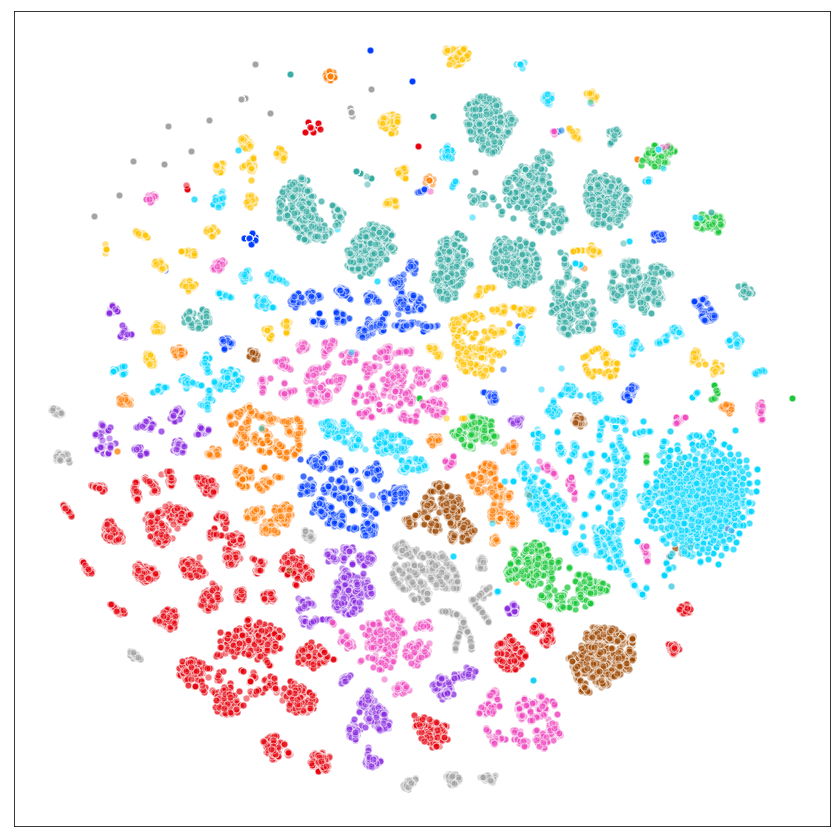}
         \caption*{Layer 19}
     \end{subfigure}
    \hfill
    \begin{subfigure}[b]{0.19\textwidth}
         \centering
         \includegraphics[width=\textwidth]{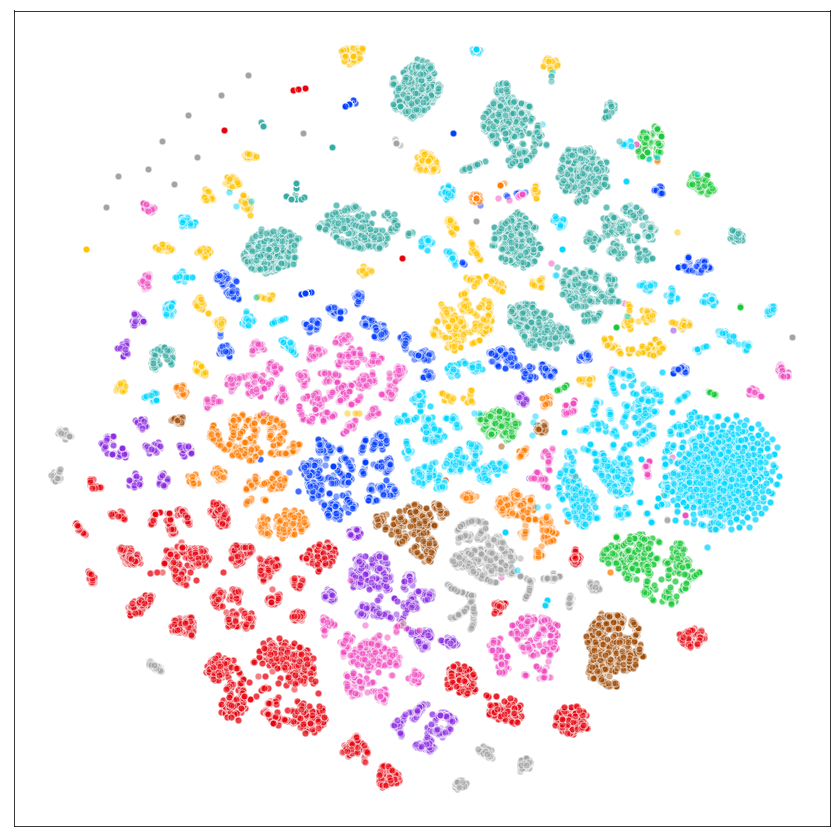}
         \caption*{Layer 20}
     \end{subfigure}
    \hfill
     \begin{subfigure}[b]{0.19\textwidth}
         \centering
         \includegraphics[width=\textwidth]{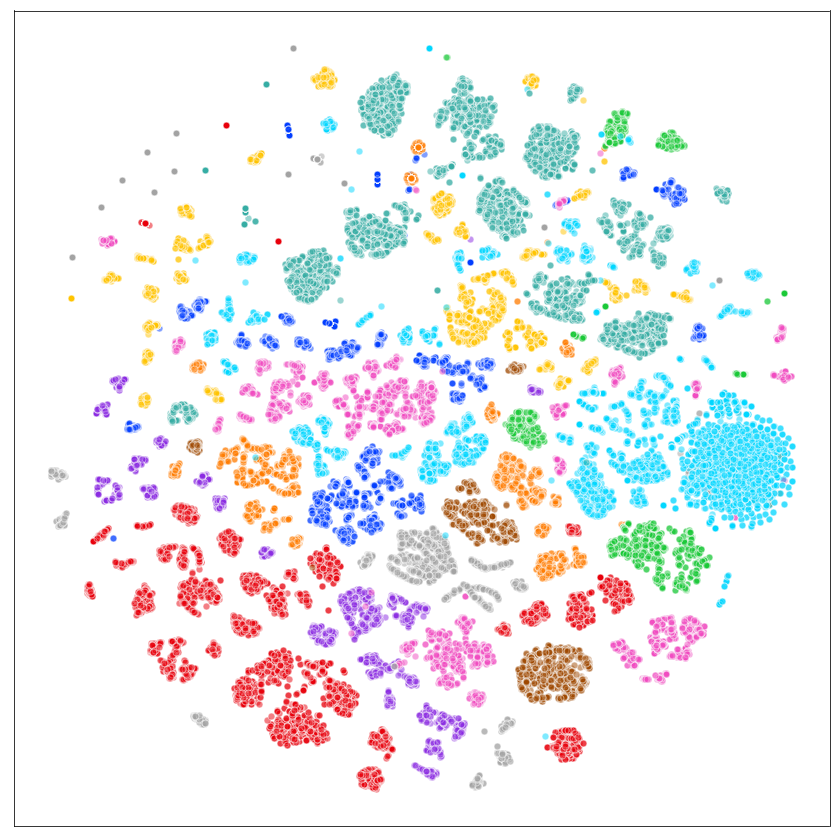}
         \caption*{Layer 21}
     \end{subfigure}
    \hfill
     \begin{subfigure}[b]{0.19\textwidth}
         \centering
         \includegraphics[width=\textwidth]{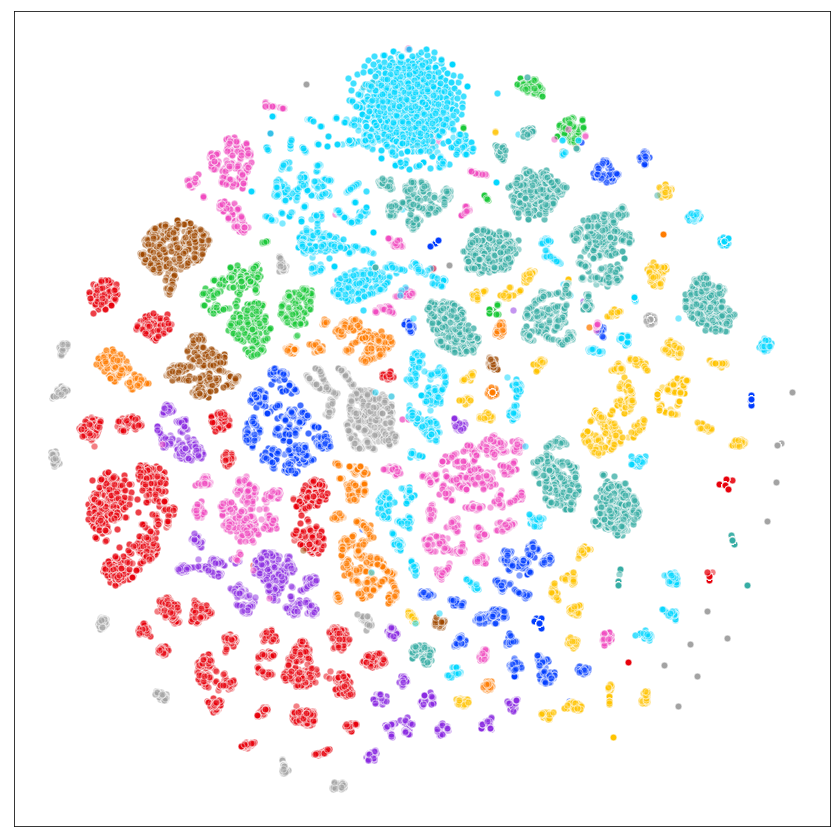}
         \caption*{Layer 22}
     \end{subfigure}
     \begin{subfigure}[b]{0.19\textwidth}
         \centering
         \includegraphics[width=\textwidth]{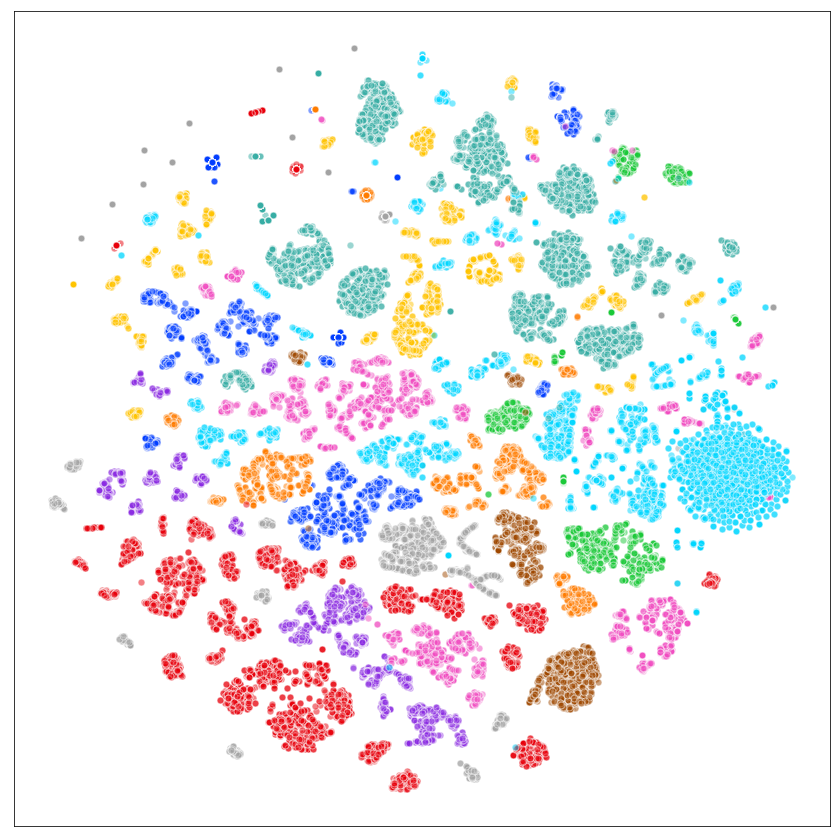}
         \caption*{Layer 23}
     \end{subfigure}
    \hfill
    \begin{subfigure}[b]{0.19\textwidth}
         \centering
         \includegraphics[width=\textwidth]{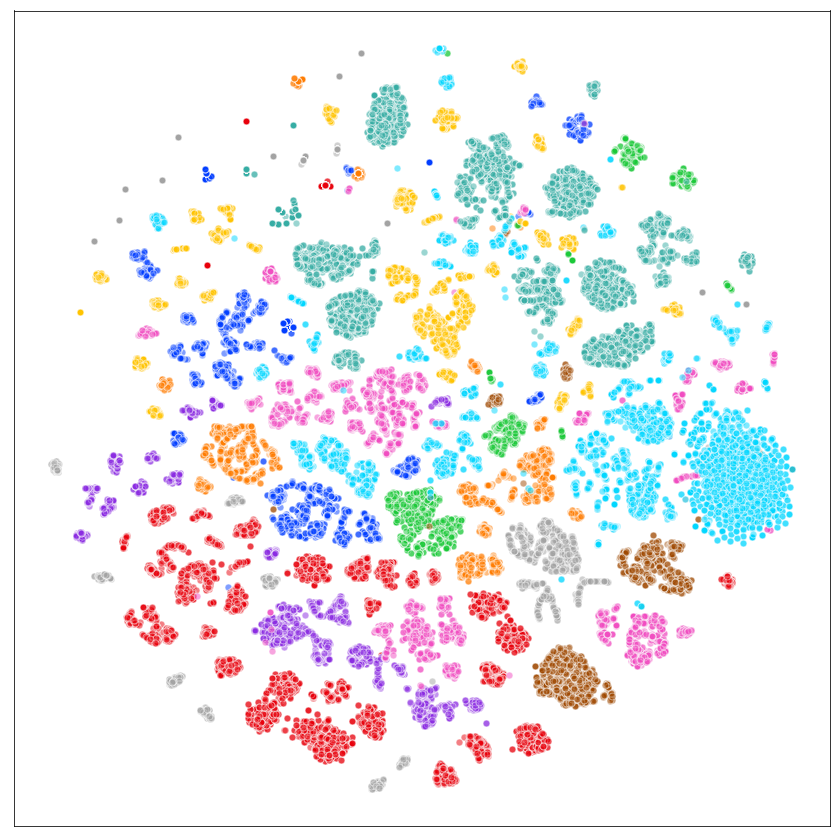}
         \caption*{Layer 24}
     \end{subfigure}
     \hfill
    \begin{subfigure}[b]{0.19\textwidth}
         \centering
         \includegraphics[width=\textwidth]{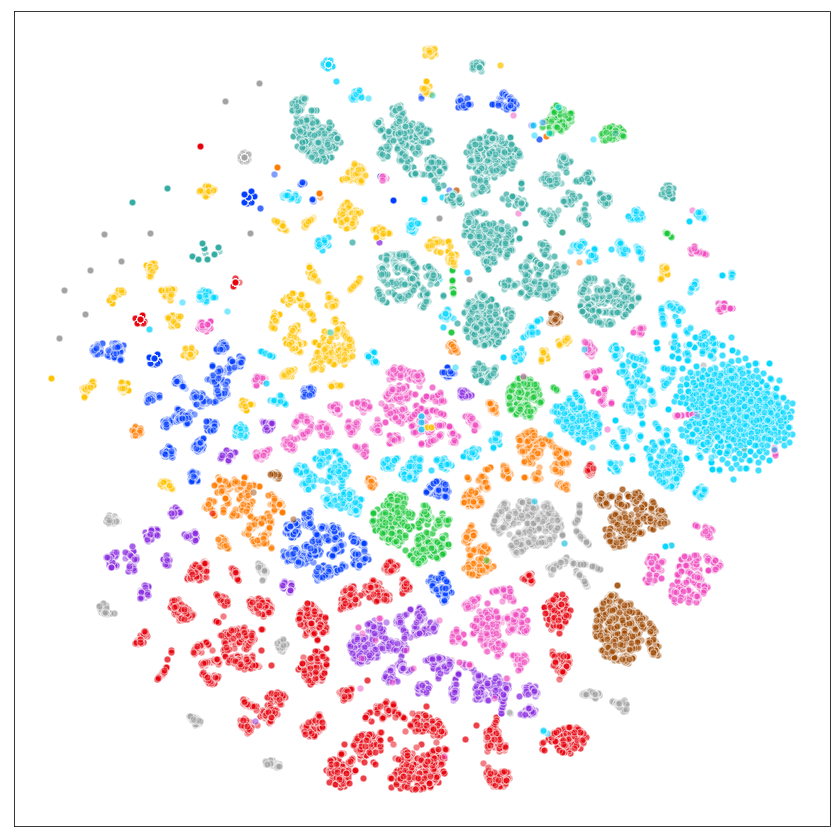}
         \caption*{Layer 25}
     \end{subfigure}
     \hfill
    \begin{subfigure}[b]{0.19\textwidth}
         \centering
         \includegraphics[width=\textwidth]{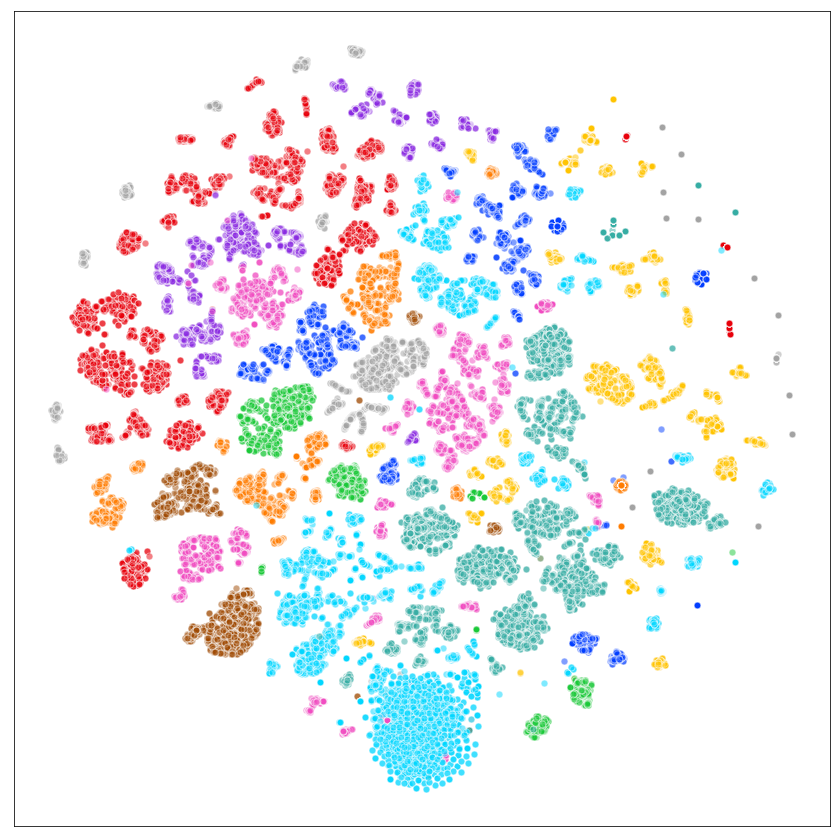}
         \caption*{Layer 26}
     \end{subfigure}
     \hfill
     \begin{subfigure}[b]{0.19\textwidth}
         \centering
         \includegraphics[width=\textwidth]{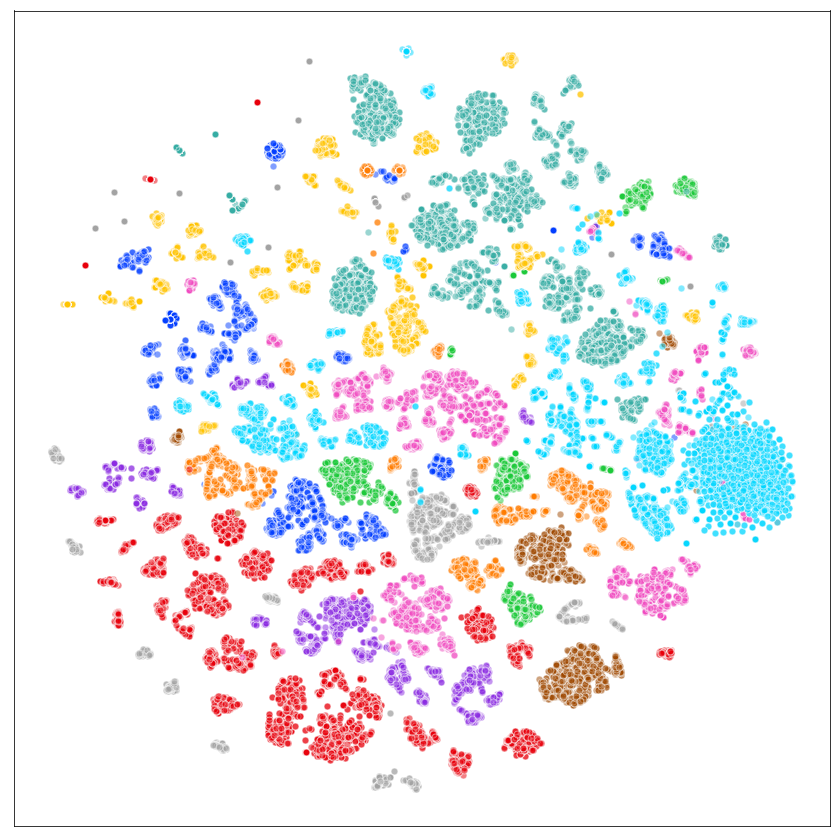}
         \caption*{Layer 27}
     \end{subfigure}
     \begin{subfigure}[b]{0.19\textwidth}
         \centering
         \includegraphics[width=\textwidth]{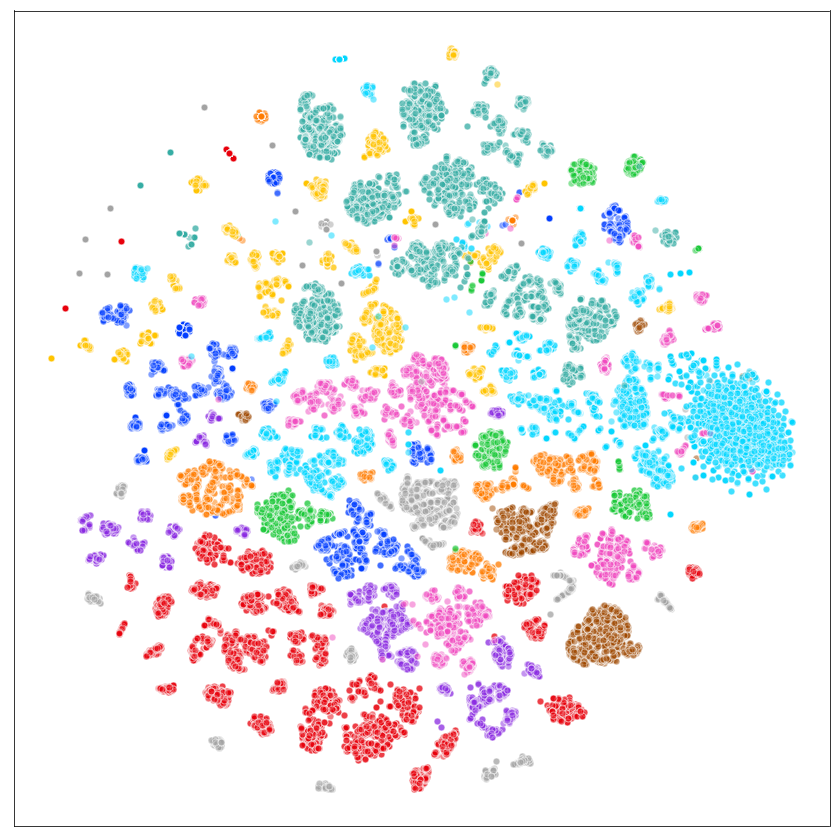}
         \caption*{Layer 28}
     \end{subfigure}
    \hfill
    \begin{subfigure}[b]{0.19\textwidth}
         \centering
         \includegraphics[width=\textwidth]{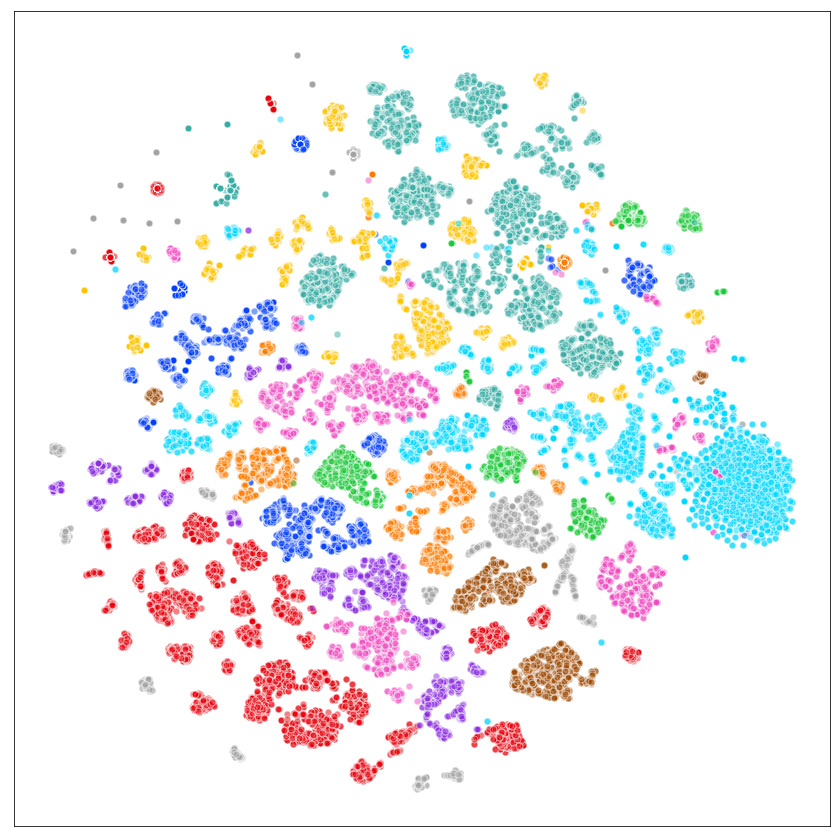}
         \caption*{Layer 29}
     \end{subfigure}
    \hfill
    \begin{subfigure}[b]{0.19\textwidth}
         \centering
         \includegraphics[width=\textwidth]{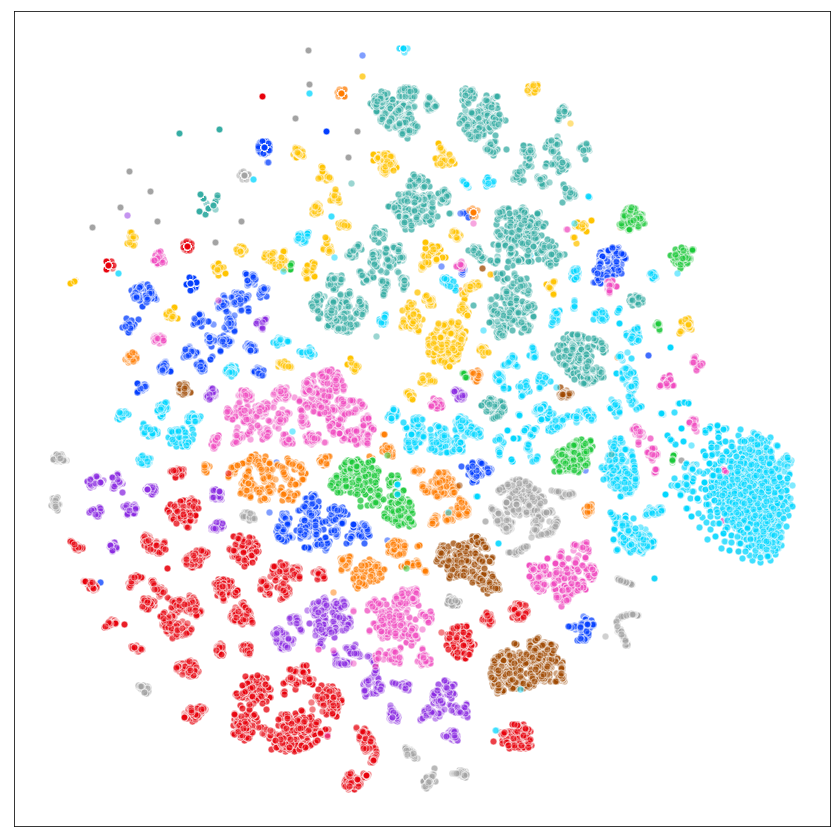}
         \caption*{Layer 30}
     \end{subfigure}
    \hfill
     \begin{subfigure}[b]{0.19\textwidth}
         \centering
         \includegraphics[width=\textwidth]{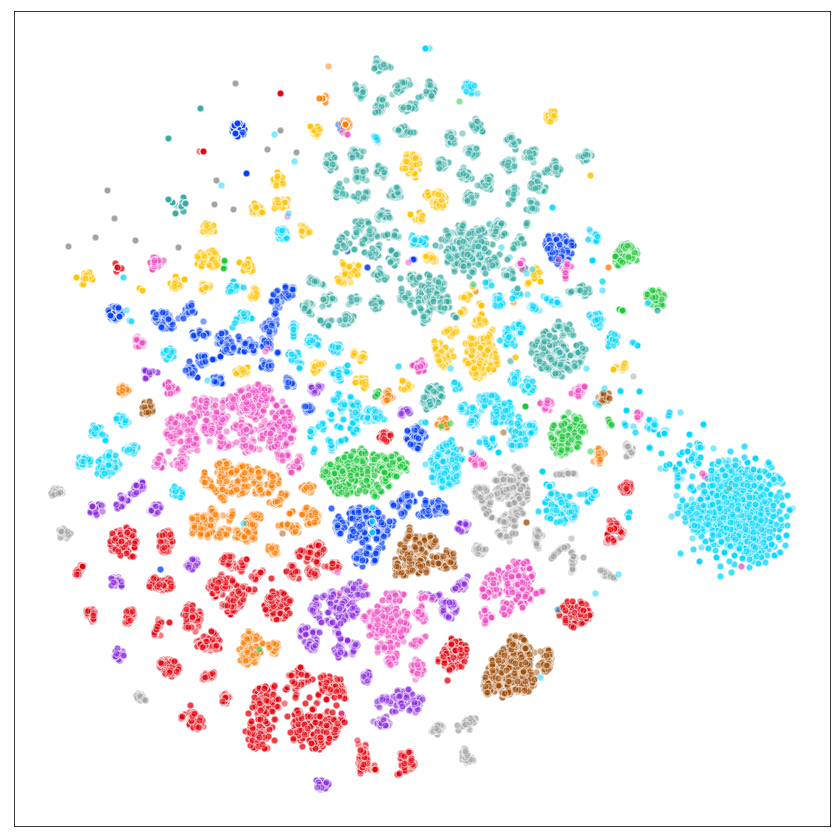}
         \caption*{Layer 31}
     \end{subfigure}
    \hfill
     \begin{subfigure}[b]{0.19\textwidth}
         \centering
         \includegraphics[width=\textwidth]{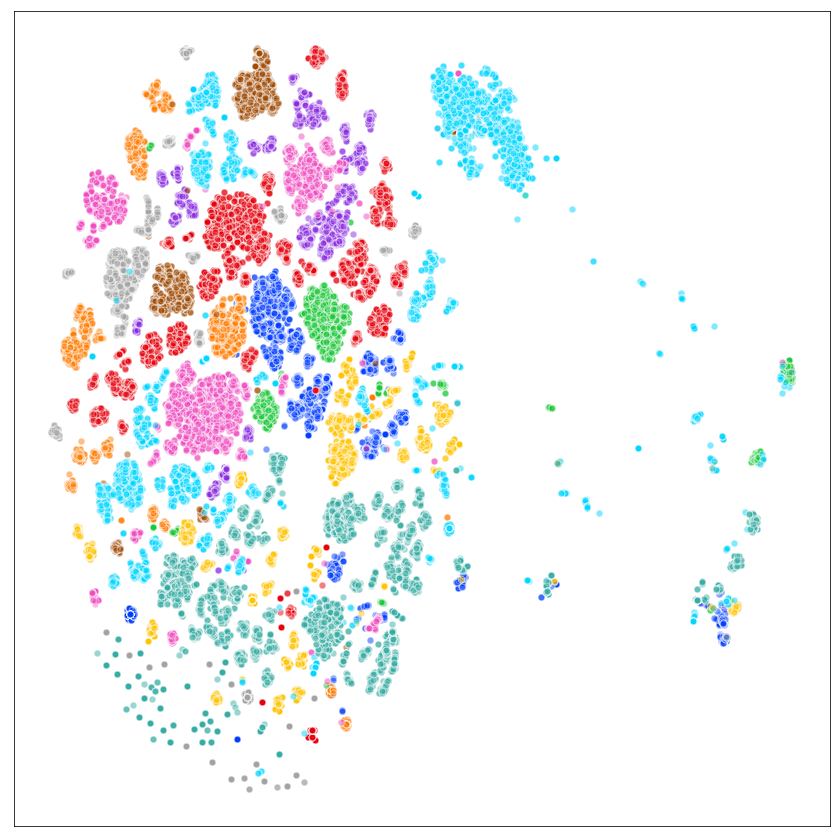}
         \caption*{Layer 32}
     \end{subfigure}
    \caption{t-SNE visualizations of the representations for each task cluster in different layers of the pre-trained Llama 2 model. Each subplot presents the t-SNE projection of the representations, color-coded by task cluster, for a specific layer of the respective model. ``Reading comp.'' denotes reading comprehension tasks, and ``reading comp. w/ c.s.'' denotes reading comprehension tasks with commonsense reasoning. We omit layer 10 and 15 to fit in one page and as we have provided them earlier.}
    \label{fig:tsne_all_llama2}
\end{figure*}

\begin{figure*}[!ht]
     \centering
     \begin{subfigure}[b]{\textwidth}
         \centering
         \includegraphics[width=\textwidth]{tsne_figures/legend.png}
         \vspace{-2ex}
     \end{subfigure}
     \begin{subfigure}[b]{0.19\textwidth}
         \centering
         \includegraphics[width=\textwidth]{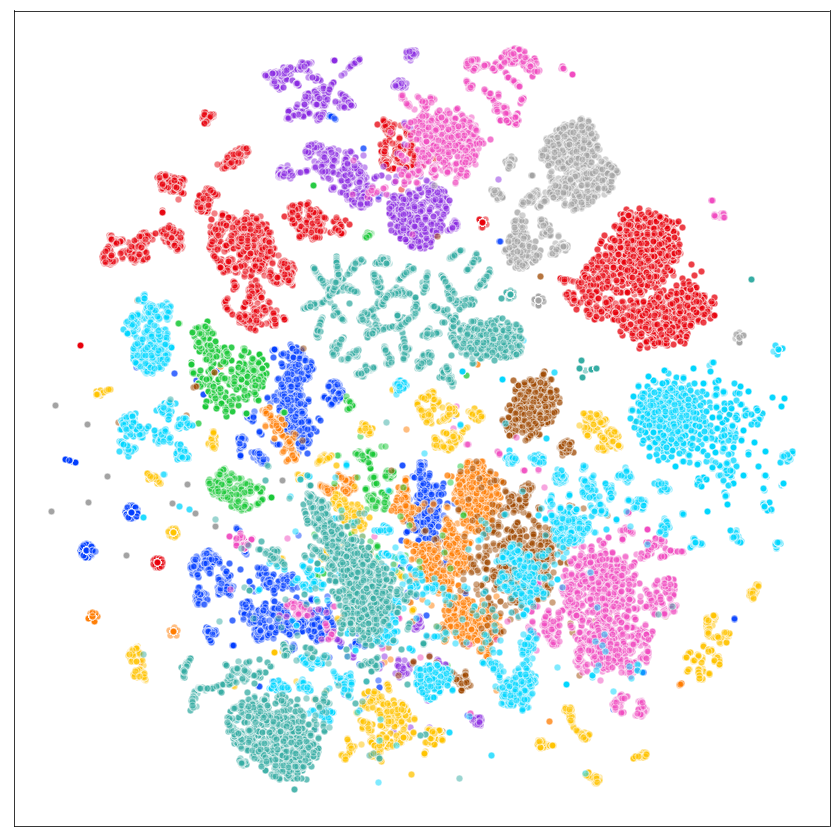}
         \caption*{Layer 1}
     \end{subfigure}
    \hfill
    \begin{subfigure}[b]{0.19\textwidth}
         \centering
         \includegraphics[width=\textwidth]{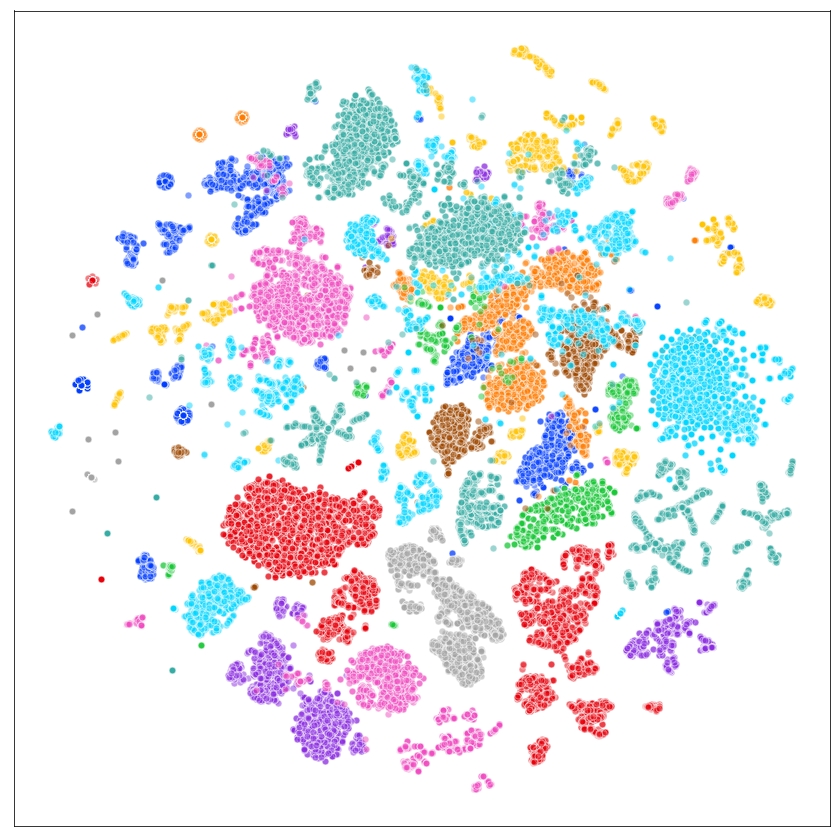}
         \caption*{Layer 2}
     \end{subfigure}
     \hfill
    \begin{subfigure}[b]{0.19\textwidth}
         \centering
         \includegraphics[width=\textwidth]{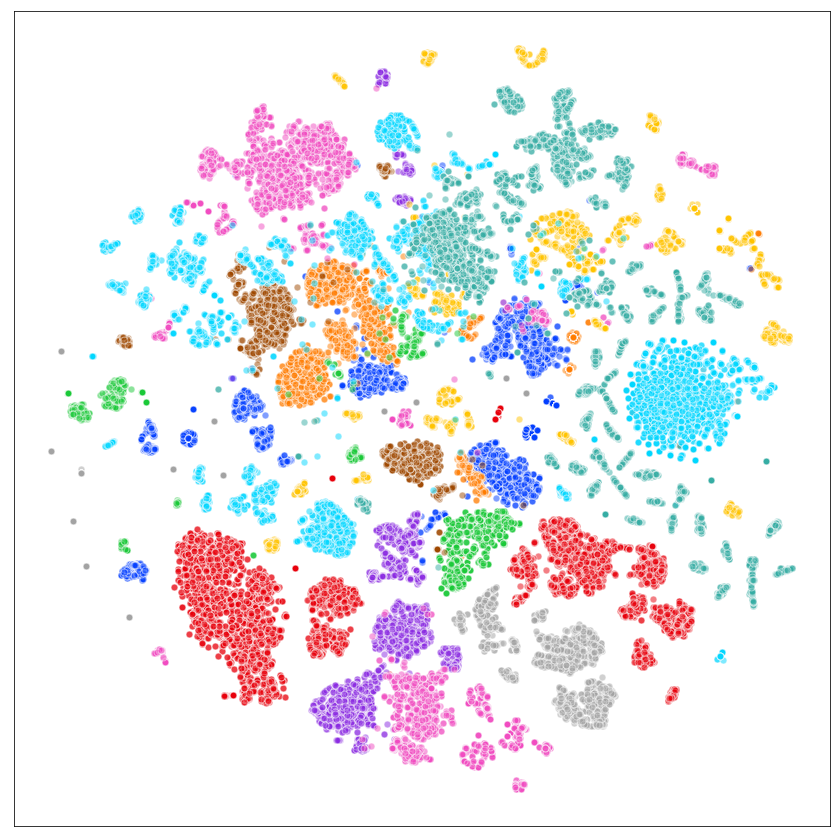}
         \caption*{Layer 3}
     \end{subfigure}
     \hfill
    \begin{subfigure}[b]{0.19\textwidth}
         \centering
         \includegraphics[width=\textwidth]{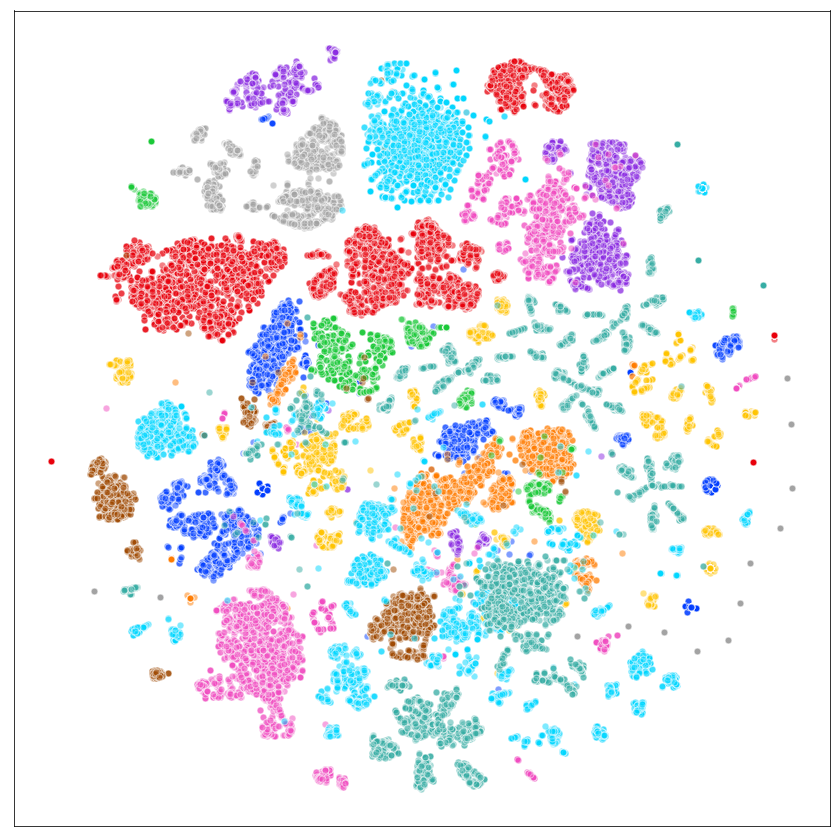}
         \caption*{Layer 4}
     \end{subfigure}
     \hfill
     \begin{subfigure}[b]{0.19\textwidth}
         \centering
         \includegraphics[width=\textwidth]{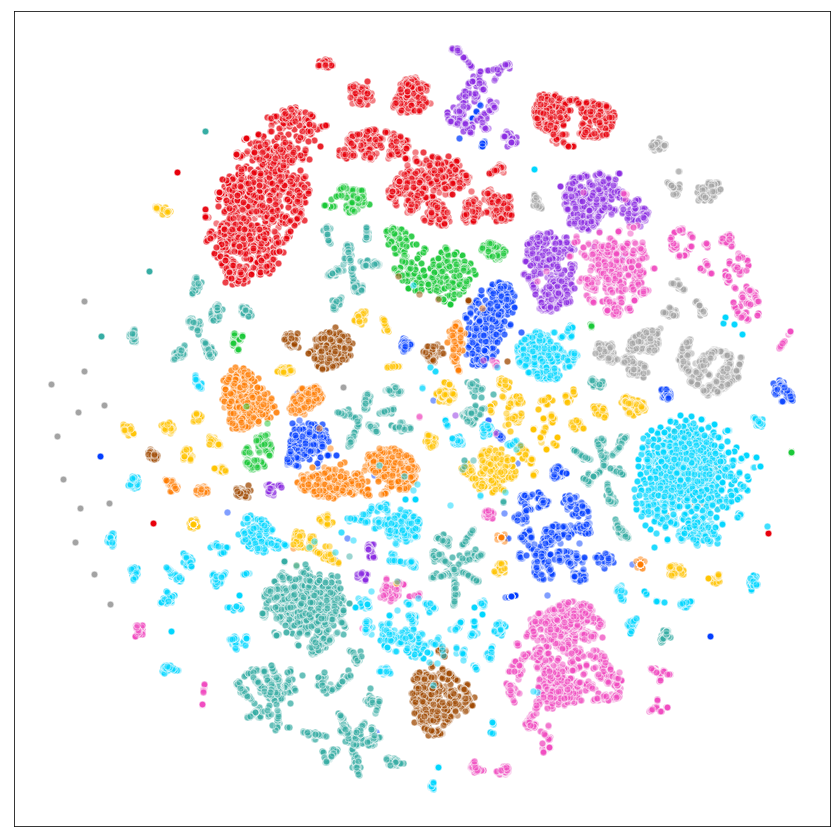}
         \caption*{Layer 5}
     \end{subfigure}
     \begin{subfigure}[b]{0.19\textwidth}
         \centering
         \includegraphics[width=\textwidth]{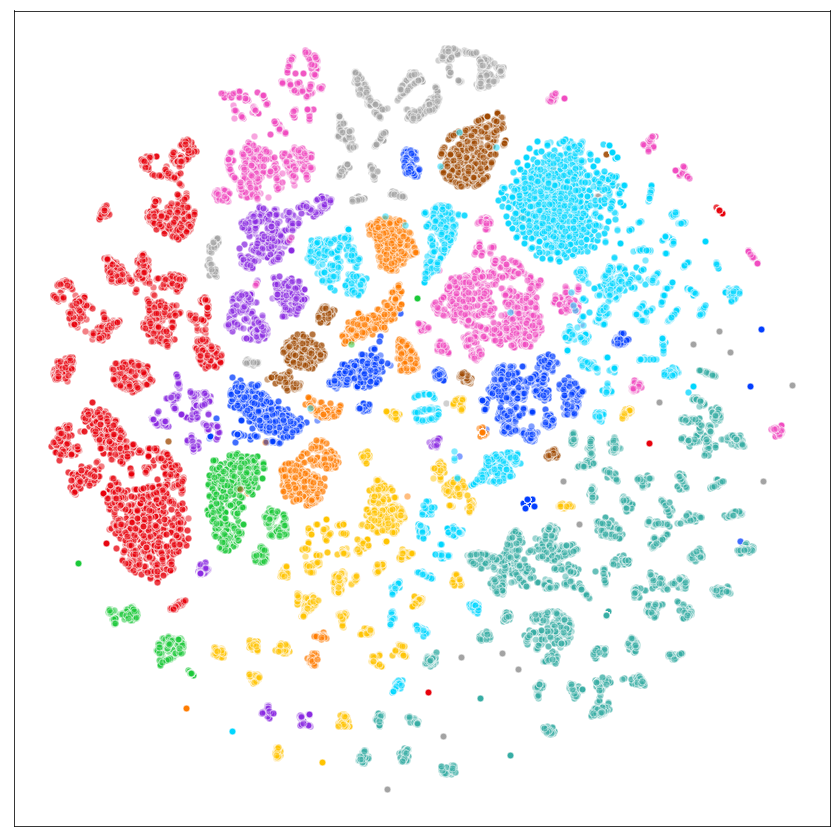}
         \caption*{Layer 6}
     \end{subfigure}
    \hfill
    \begin{subfigure}[b]{0.19\textwidth}
         \centering
         \includegraphics[width=\textwidth]{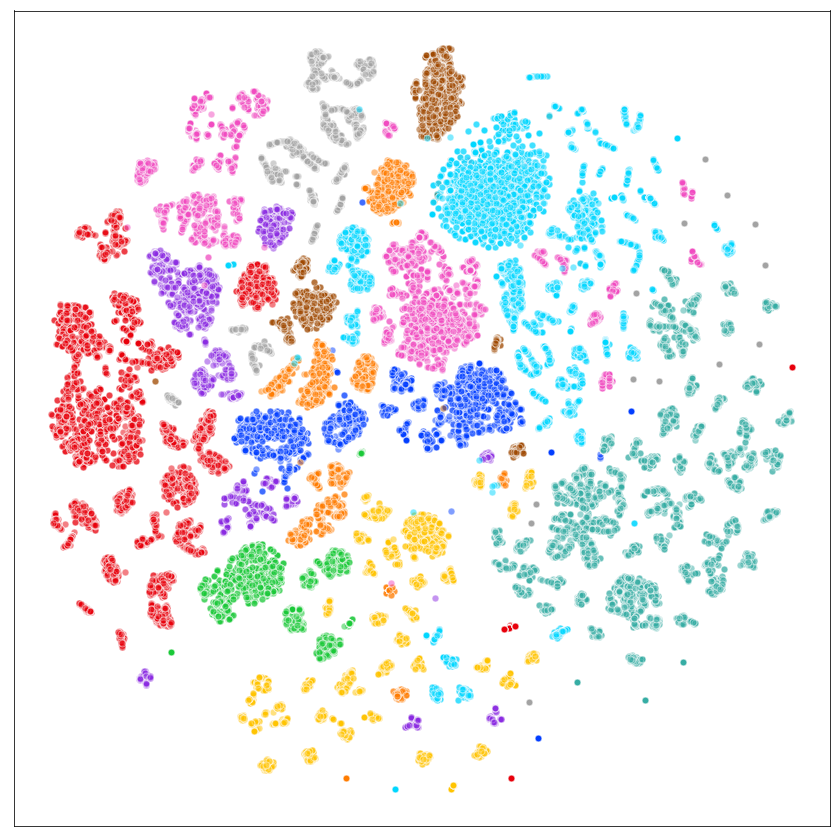}
         \caption*{Layer 7}
     \end{subfigure}
    \hfill
    \begin{subfigure}[b]{0.19\textwidth}
         \centering
         \includegraphics[width=\textwidth]{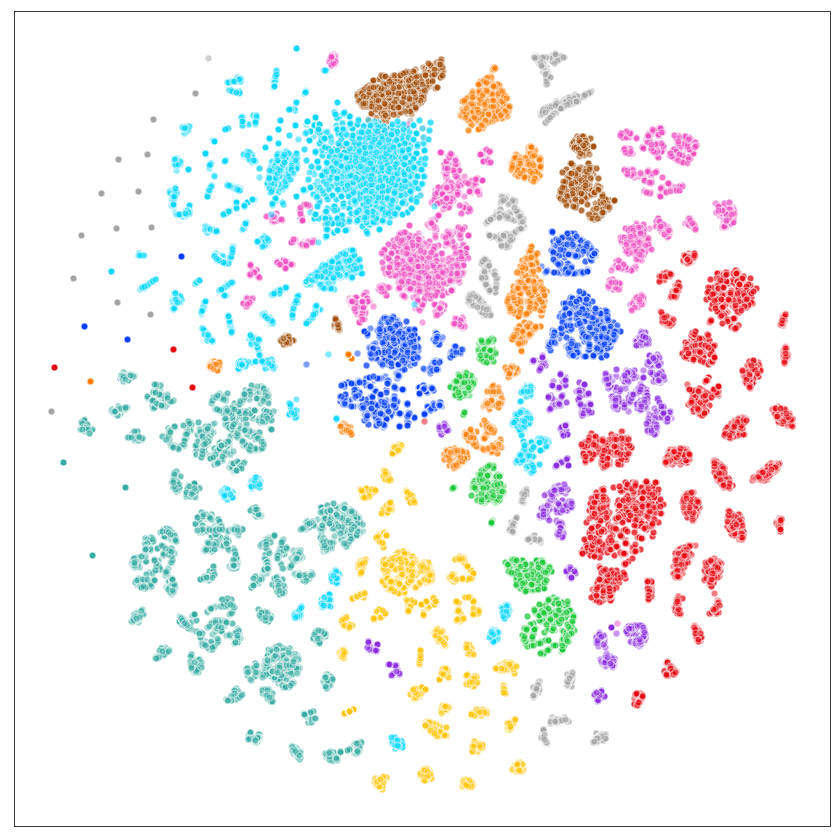}
         \caption*{Layer 8}
     \end{subfigure}
    \hfill
     \begin{subfigure}[b]{0.19\textwidth}
         \centering
         \includegraphics[width=\textwidth]{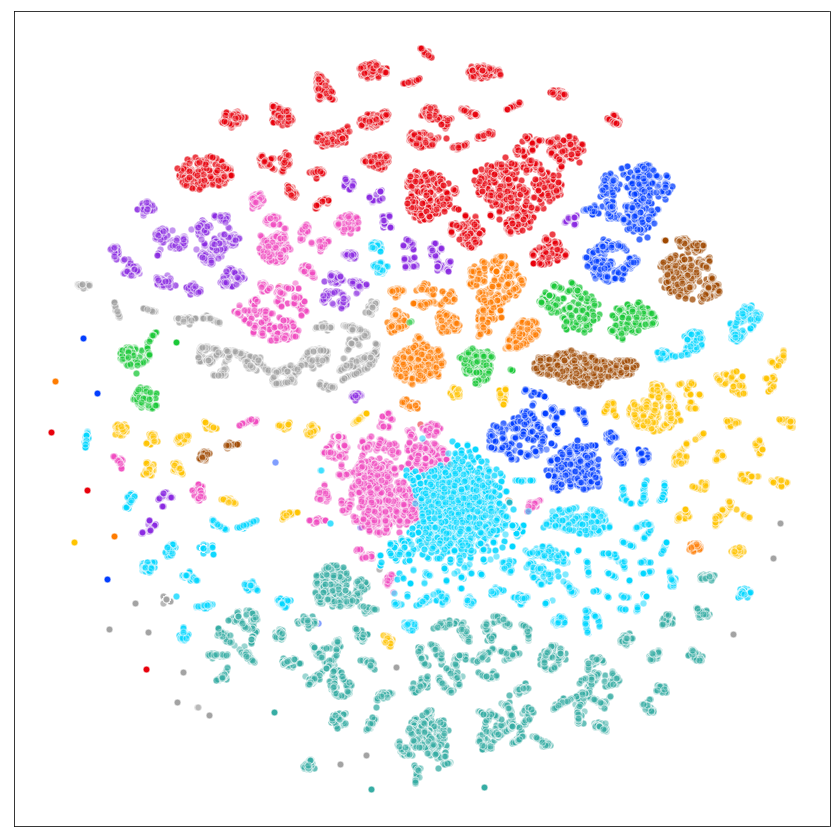}
         \caption*{Layer 9}
     \end{subfigure}
    \hfill
     \begin{subfigure}[b]{0.19\textwidth}
         \centering
         \includegraphics[width=\textwidth]{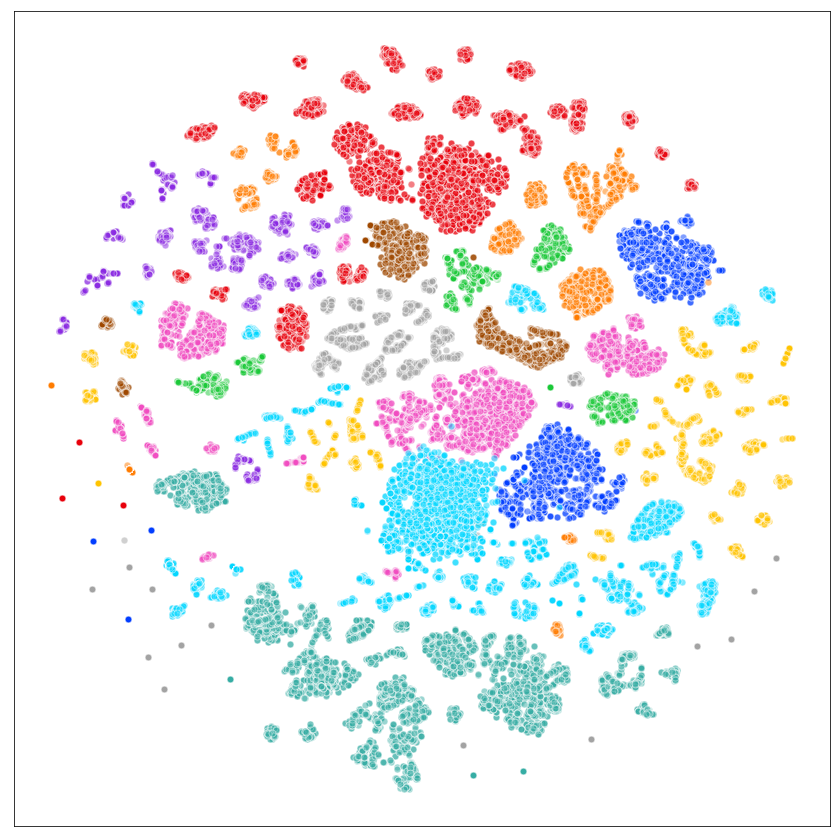}
         \caption*{Layer 11}
     \end{subfigure}
     \begin{subfigure}[b]{0.19\textwidth}
         \centering
         \includegraphics[width=\textwidth]{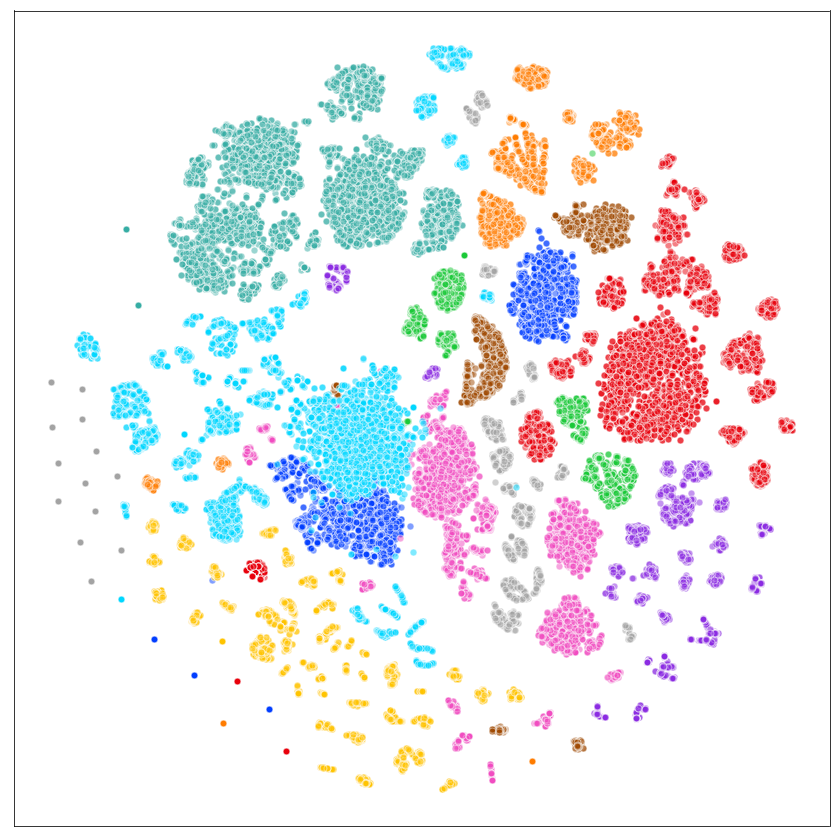}
         \caption*{Layer 12}
     \end{subfigure}
    \hfill
    \begin{subfigure}[b]{0.19\textwidth}
         \centering
         \includegraphics[width=\textwidth]{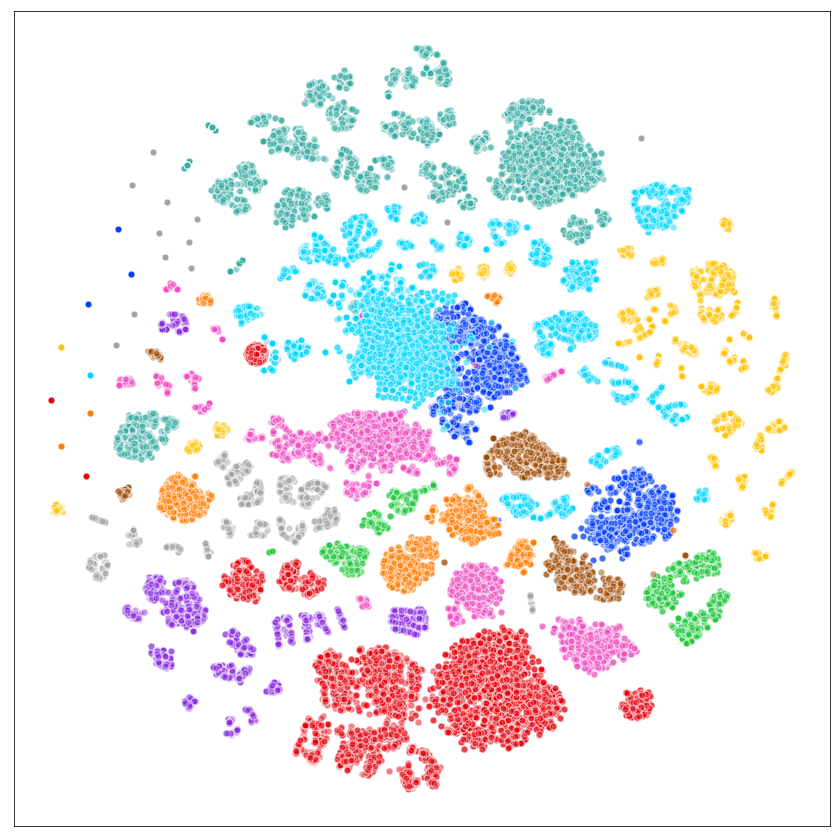}
         \caption*{Layer 13}
     \end{subfigure}
     \hfill
    \begin{subfigure}[b]{0.19\textwidth}
         \centering
         \includegraphics[width=\textwidth]{tsne_figures_low_res/tsne_llama2_sft_last_token_by_cluster_l4.png}
         \caption*{Layer 14}
     \end{subfigure}
     \hfill
    \begin{subfigure}[b]{0.19\textwidth}
         \centering
         \includegraphics[width=\textwidth]{tsne_figures_low_res/tsne_llama2_sft_last_token_by_cluster_l6.png}
         \caption*{Layer 16}
     \end{subfigure}
     \hfill
     \begin{subfigure}[b]{0.19\textwidth}
         \centering
         \includegraphics[width=\textwidth]{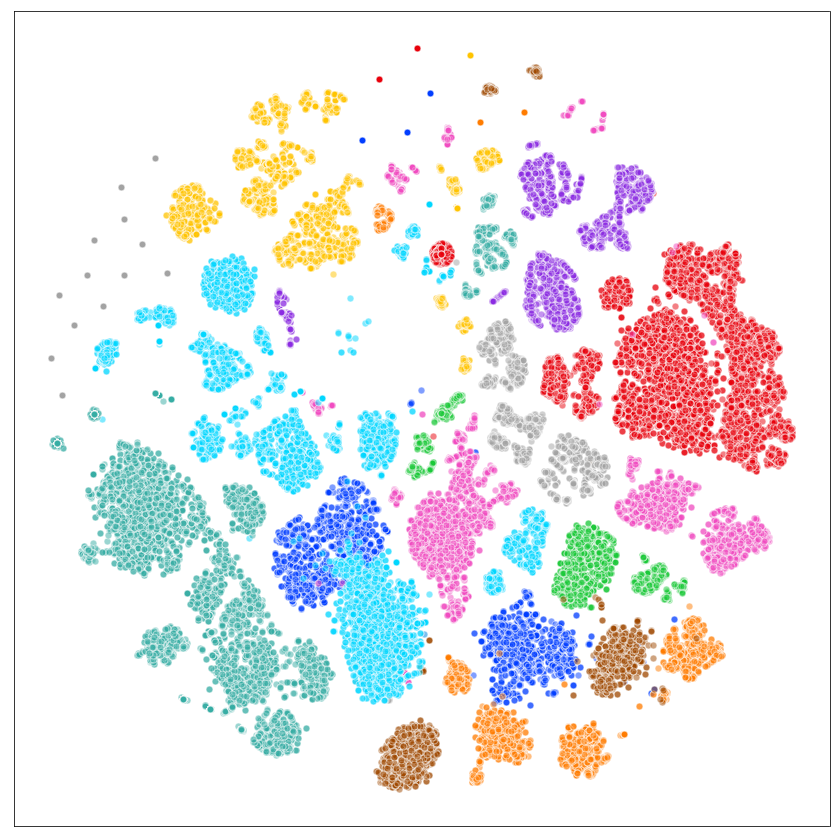}
         \caption*{Layer 17}
     \end{subfigure}
     \begin{subfigure}[b]{0.19\textwidth}
         \centering
         \includegraphics[width=\textwidth]{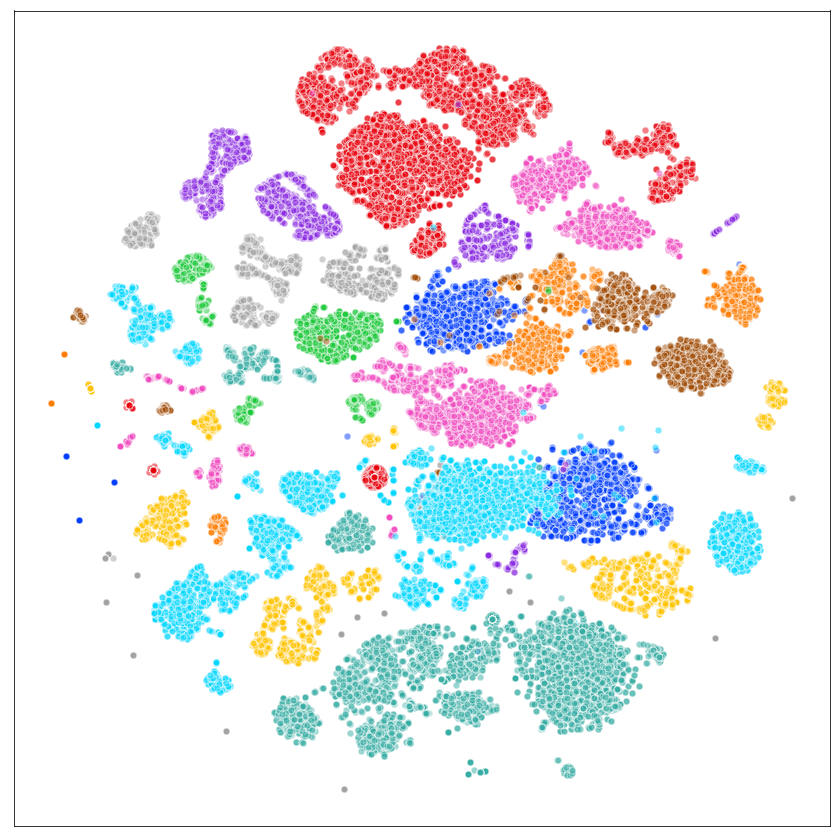}
         \caption*{Layer 18}
     \end{subfigure}
    \hfill
    \begin{subfigure}[b]{0.19\textwidth}
         \centering
         \includegraphics[width=\textwidth]{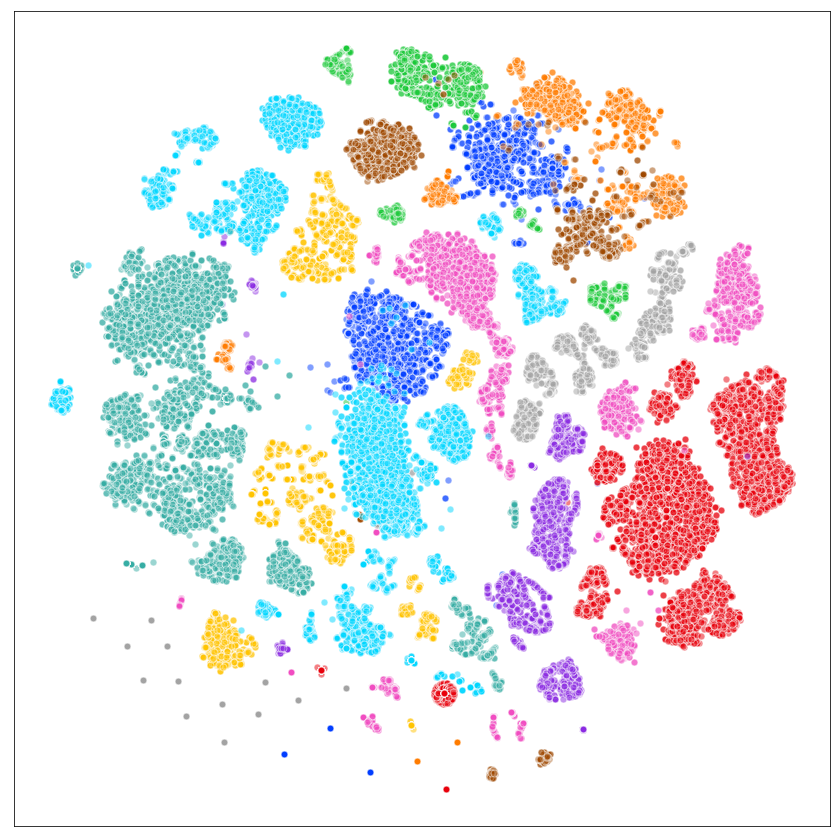}
         \caption*{Layer 19}
     \end{subfigure}
    \hfill
    \begin{subfigure}[b]{0.19\textwidth}
         \centering
         \includegraphics[width=\textwidth]{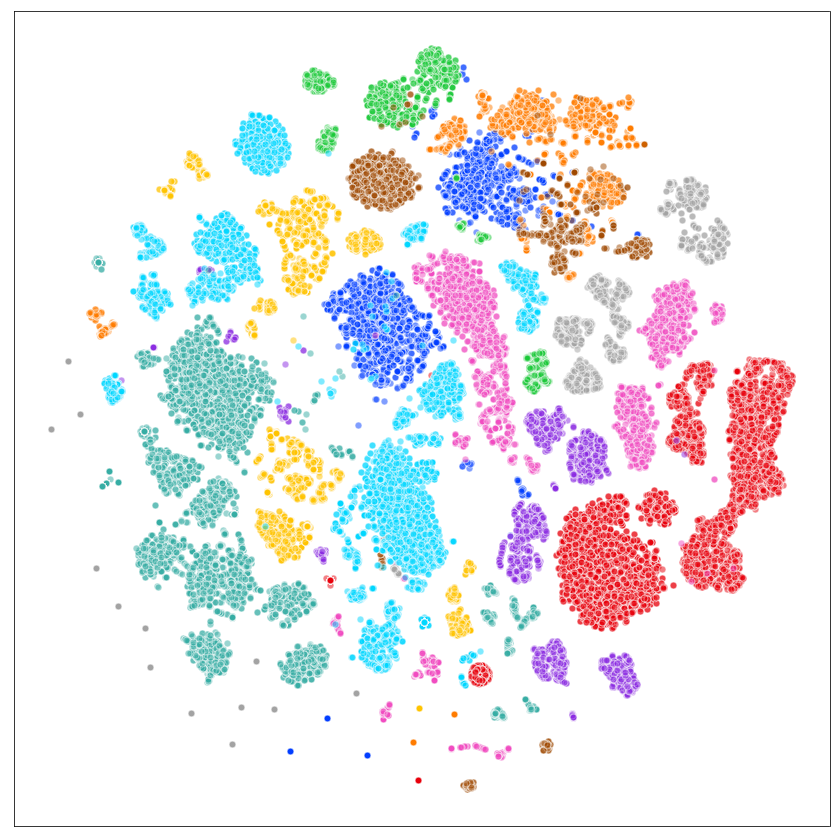}
         \caption*{Layer 20}
     \end{subfigure}
    \hfill
     \begin{subfigure}[b]{0.19\textwidth}
         \centering
         \includegraphics[width=\textwidth]{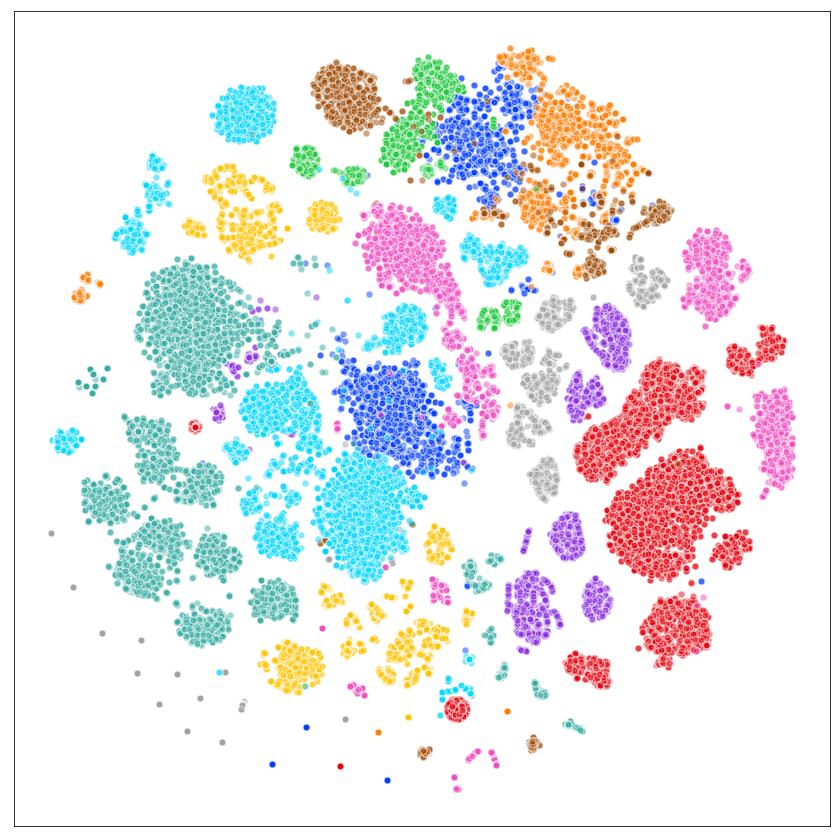}
         \caption*{Layer 21}
     \end{subfigure}
    \hfill
     \begin{subfigure}[b]{0.19\textwidth}
         \centering
         \includegraphics[width=\textwidth]{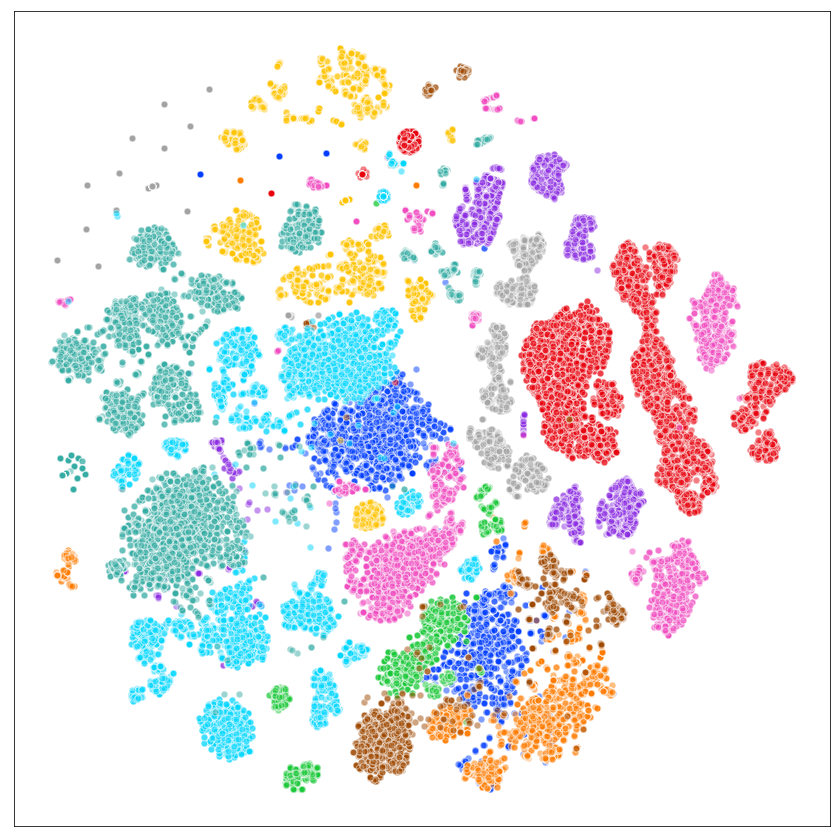}
         \caption*{Layer 22}
     \end{subfigure}
     \begin{subfigure}[b]{0.19\textwidth}
         \centering
         \includegraphics[width=\textwidth]{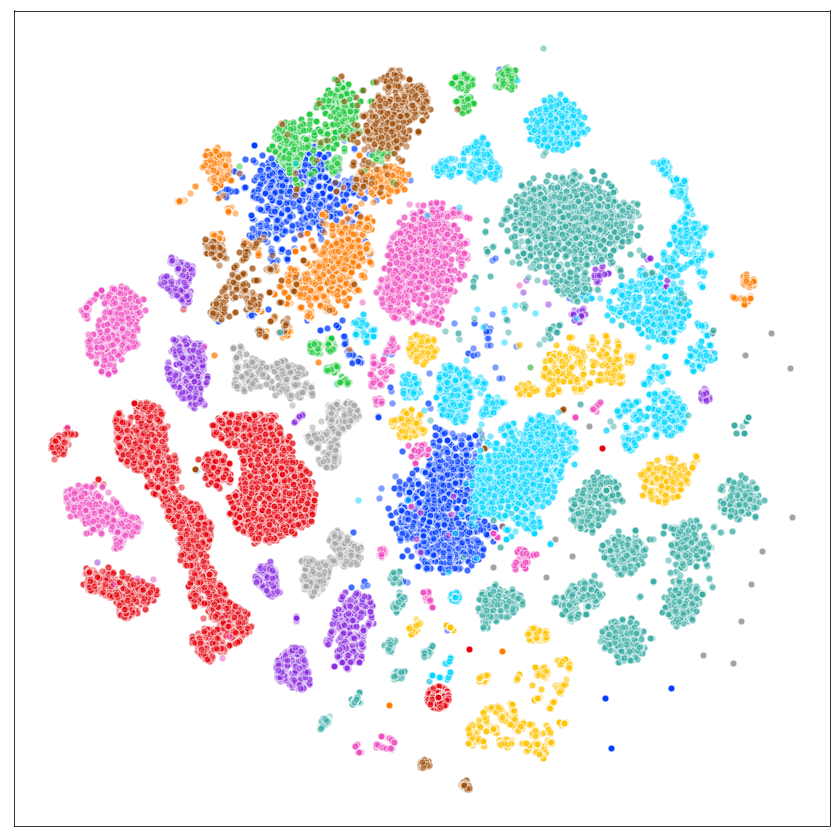}
         \caption*{Layer 23}
     \end{subfigure}
    \hfill
    \begin{subfigure}[b]{0.19\textwidth}
         \centering
         \includegraphics[width=\textwidth]{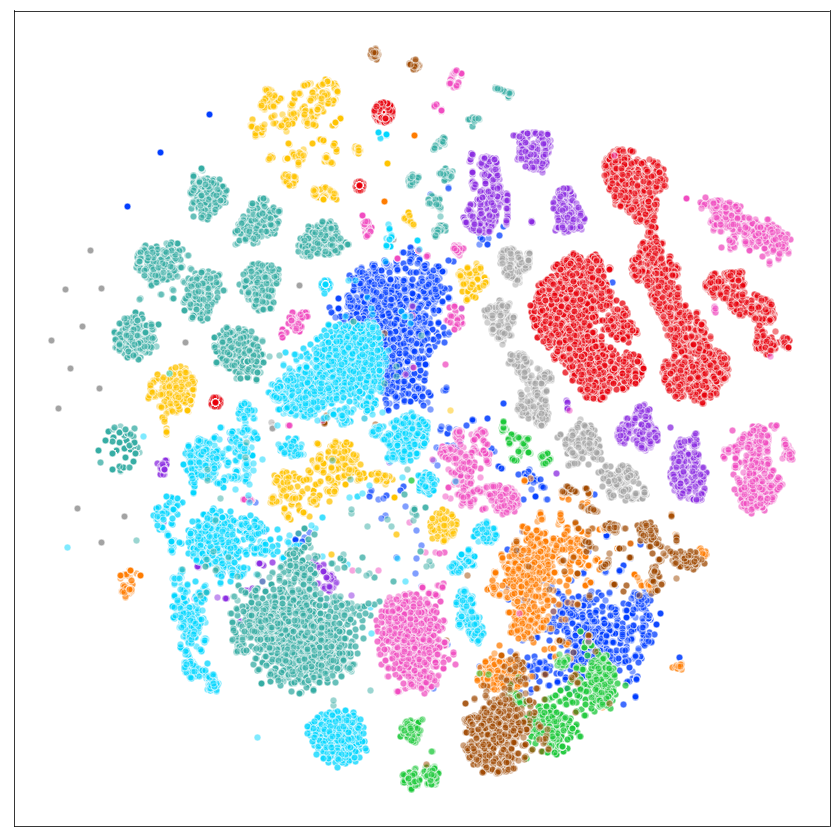}
         \caption*{Layer 24}
     \end{subfigure}
     \hfill
    \begin{subfigure}[b]{0.19\textwidth}
         \centering
         \includegraphics[width=\textwidth]{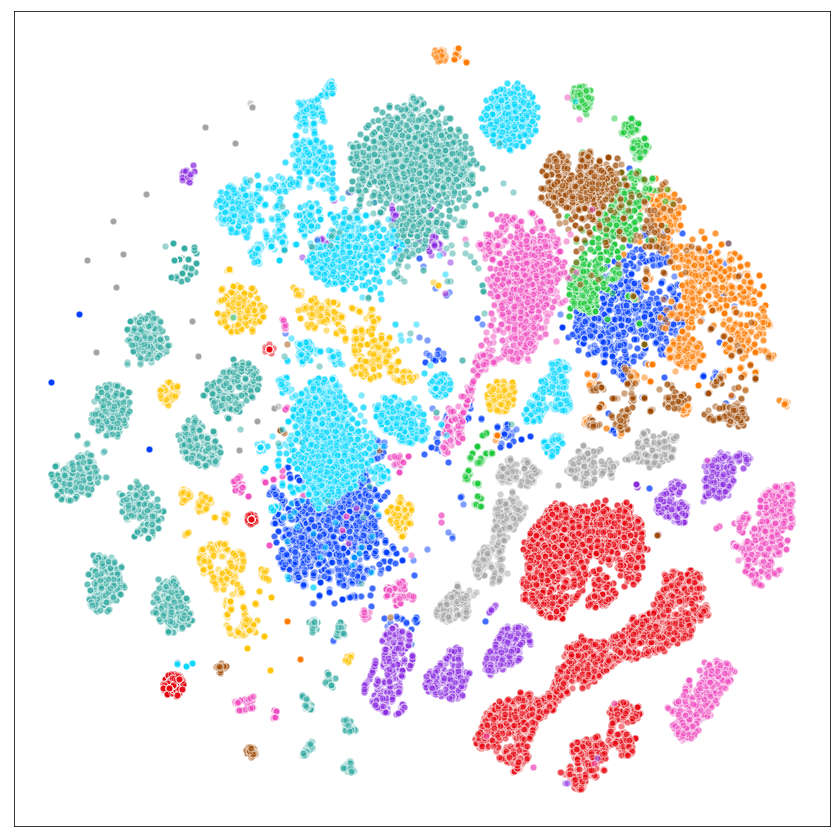}
         \caption*{Layer 25}
     \end{subfigure}
     \hfill
    \begin{subfigure}[b]{0.19\textwidth}
         \centering
         \includegraphics[width=\textwidth]{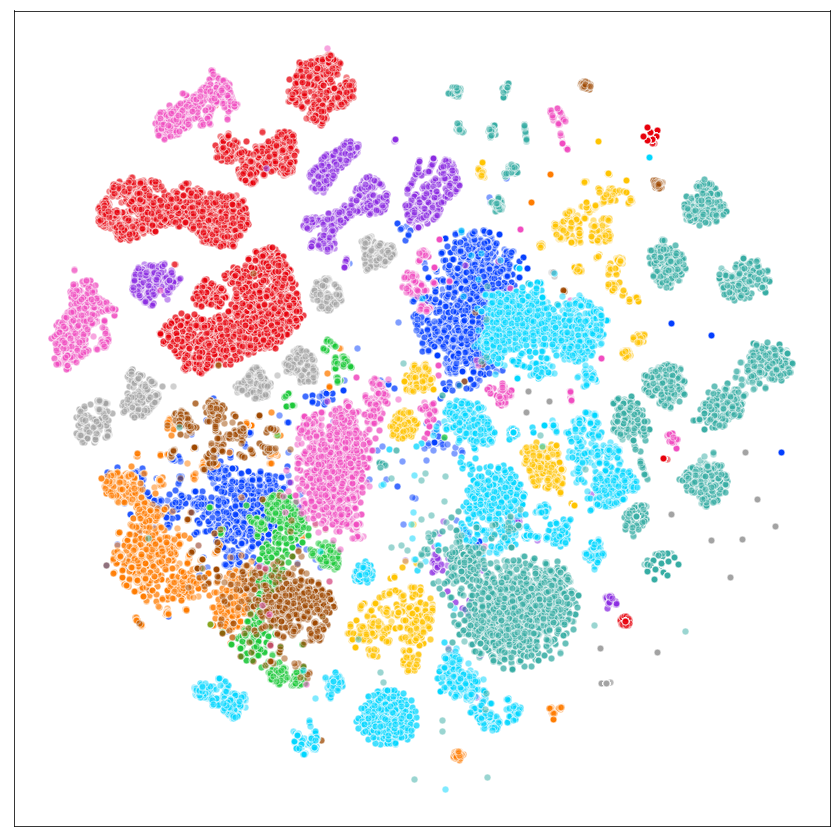}
         \caption*{Layer 26}
     \end{subfigure}
     \hfill
     \begin{subfigure}[b]{0.19\textwidth}
         \centering
         \includegraphics[width=\textwidth]{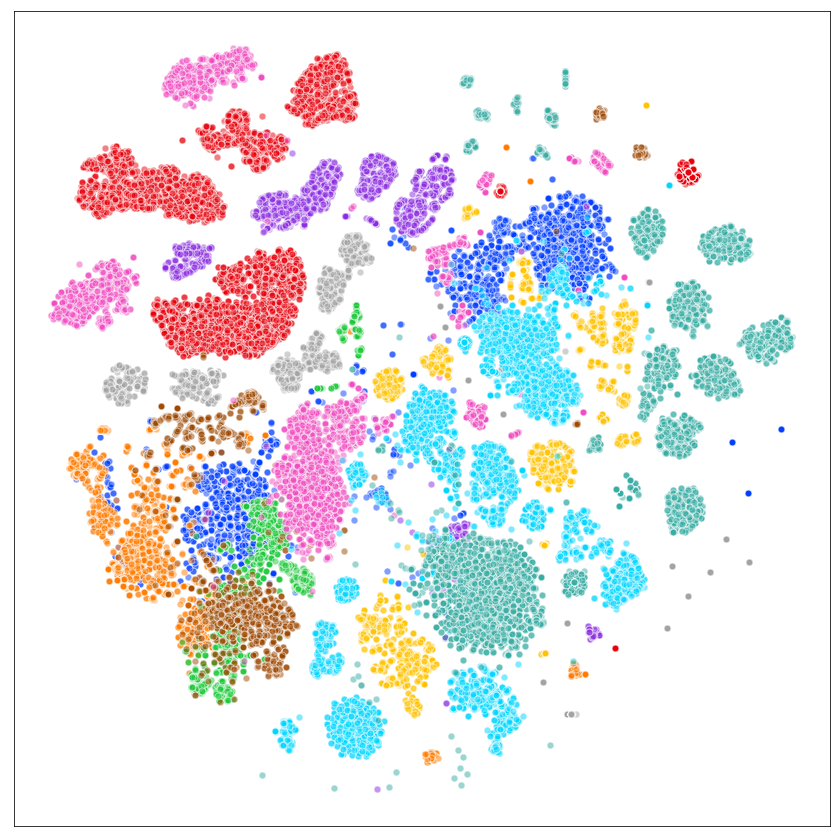}
         \caption*{Layer 27}
     \end{subfigure}
     \begin{subfigure}[b]{0.19\textwidth}
         \centering
         \includegraphics[width=\textwidth]{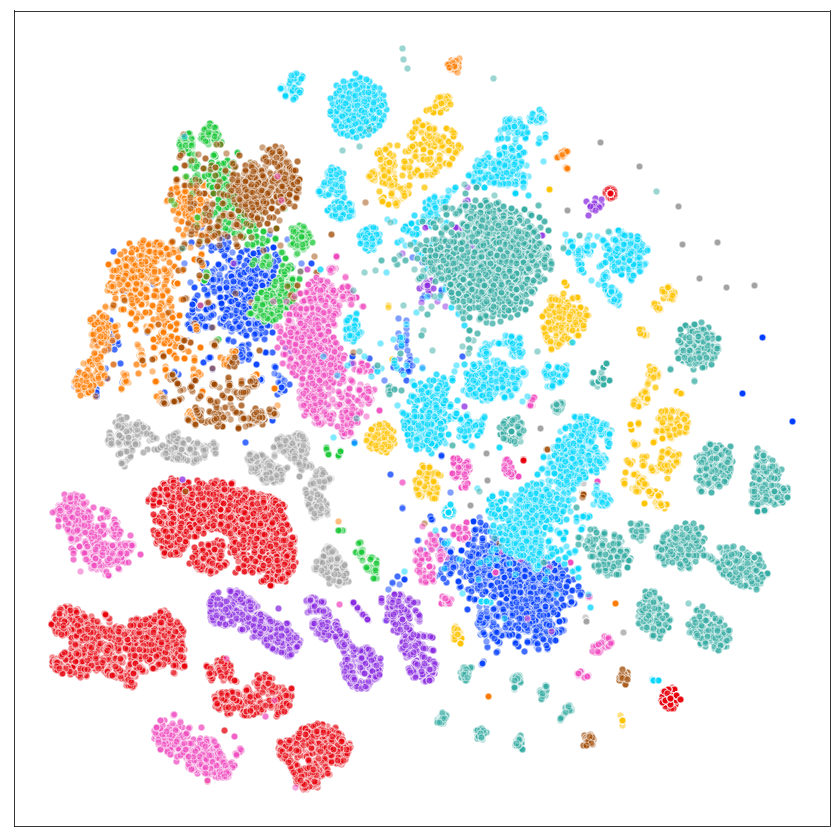}
         \caption*{Layer 28}
     \end{subfigure}
    \hfill
    \begin{subfigure}[b]{0.19\textwidth}
         \centering
         \includegraphics[width=\textwidth]{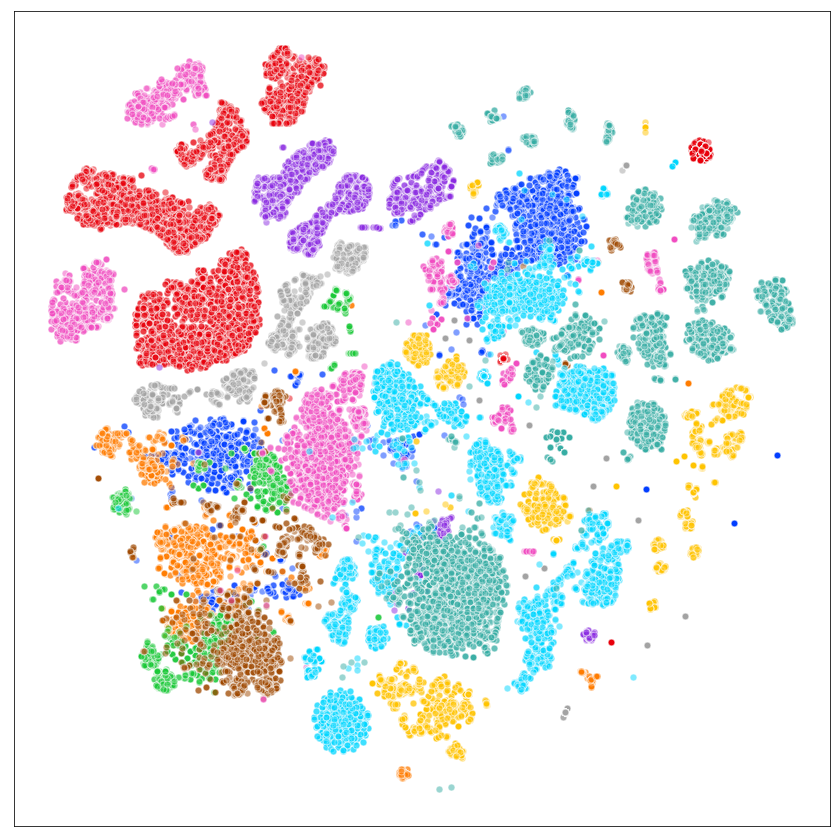}
         \caption*{Layer 29}
     \end{subfigure}
    \hfill
    \begin{subfigure}[b]{0.19\textwidth}
         \centering
         \includegraphics[width=\textwidth]{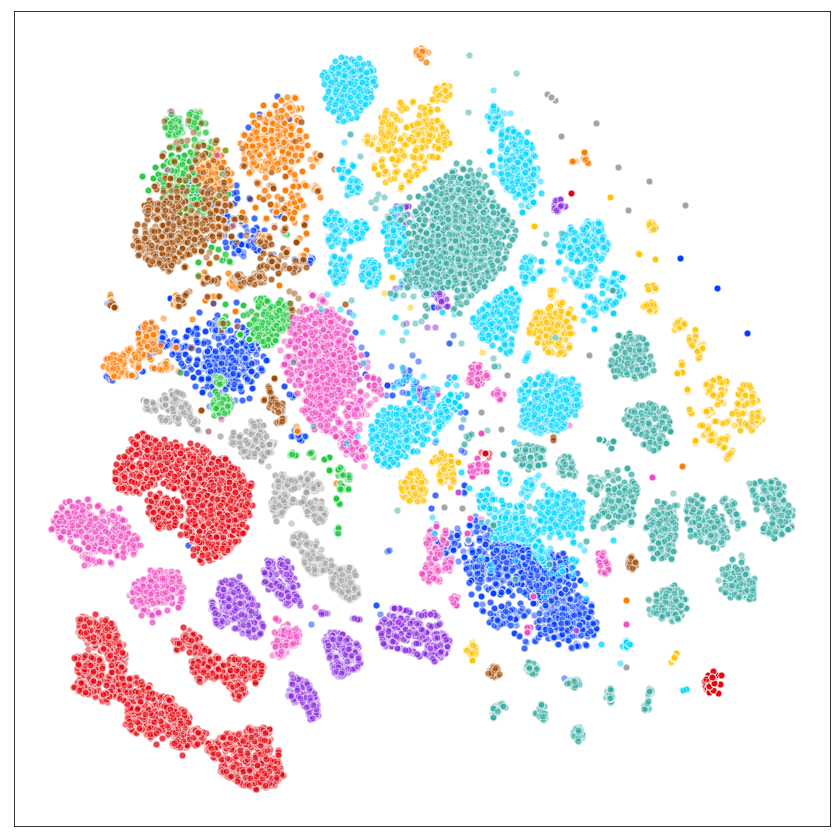}
         \caption*{Layer 30}
     \end{subfigure}
    \hfill
     \begin{subfigure}[b]{0.19\textwidth}
         \centering
         \includegraphics[width=\textwidth]{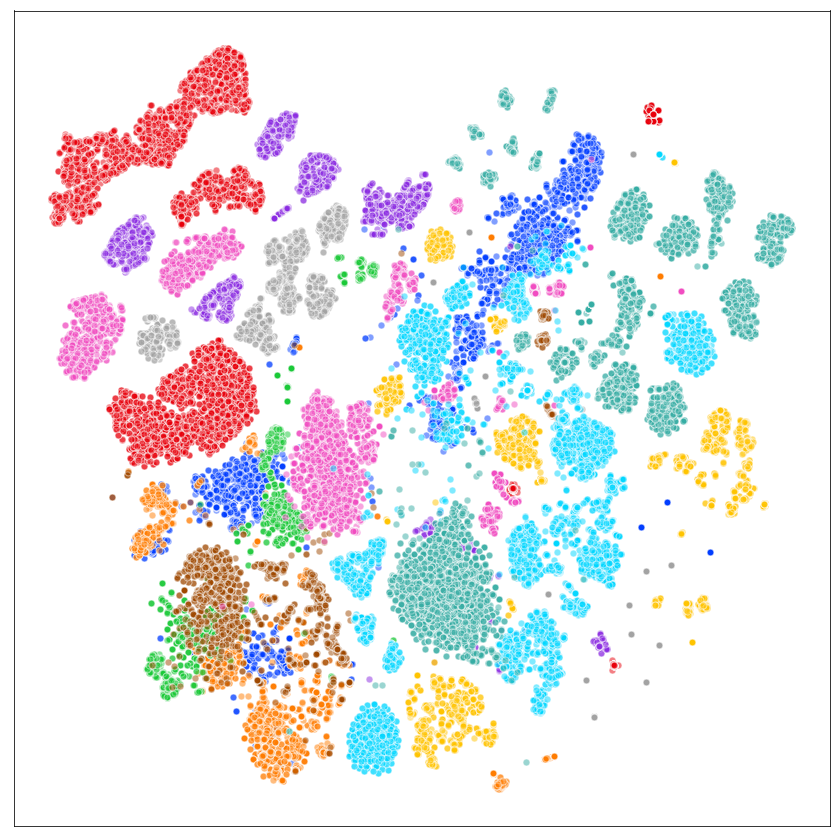}
         \caption*{Layer 31}
     \end{subfigure}
    \hfill
     \begin{subfigure}[b]{0.19\textwidth}
         \centering
         \includegraphics[width=\textwidth]{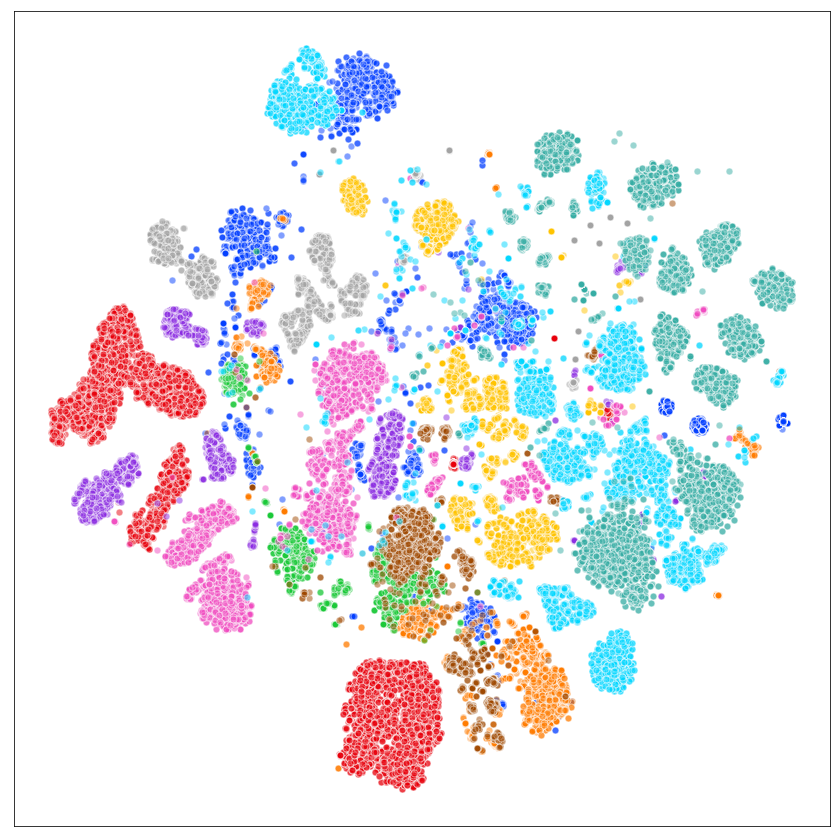}
         \caption*{Layer 32}
     \end{subfigure}
    \caption{t-SNE visualizations of the representations for each task cluster in different layers of the instruction-tuned Llama 2-SFT model. Each subplot presents the t-SNE projection of the representations, color-coded by task cluster, for a specific layer of the respective model. ``Reading comp.'' denotes reading comprehension tasks, and ``reading comp. w/ c.s.'' denotes reading comprehension tasks with commonsense reasoning. We omit layer 10 and 15 to fit in one page and as we have provided them earlier.}
    \label{fig:tsne_all_llama2_sft}
\end{figure*}

Here we provide additional results on our analysis. We provide the distribution of CKA similarities for all layers by tasks clusters in Figure~\ref{fig:cka_all_layers_all_clusters_1} and \ref{fig:cka_all_layers_all_clusters_2}. 
We also provide the t-SNE visualizations of representations in different layers of Llama 2 in Figure~\ref{fig:tsne_all_llama2}. Lastly, we provide the same visualizations for Llama 2-SFT in Figure~\ref{fig:tsne_all_llama2_sft}.

\end{document}